\documentclass[mnsc,nonblindrev]{informs3a} 

\OneAndAHalfSpacedXI

\usepackage{graphicx}
\graphicspath{{../}}
\usepackage{subfigure, epsfig}
\usepackage{natbib}

 \bibpunct[, ]{(}{)}{,}{a}{}{,}\def\bibfont{\small}
 \usepackage{pifont}
\usepackage{makecell}
\usepackage{verbatim}

\renewcommand{\bar}{\widebar}

\newcommand{\I}{ \mathbb{I} }

\newcommand{\R}{ \mathbb{R} }
\renewcommand{\P}{ \mathbb{P}}

\newcommand{\E}{ \mathbb{E} }

\usepackage{parskip} 

\definecolor{RevisionRed}{RGB}{180,30,90}
\definecolor{RevisionBlue}{RGB}{30,90,180}

\usepackage{pifont}
\usepackage{braket}
\usepackage{algorithm}
\usepackage{algorithmicx}
\usepackage[noend]{algpseudocode}
\def\1{(\mathrm{\uppercase\expandafter{\romannumeral1}})}
\def\2{(\mathrm{\uppercase\expandafter{\romannumeral2}})}
\def\O{\mathcal O}

\RequirePackage[colorlinks,citecolor=blue,linkcolor=blue,urlcolor=blue]{hyperref}
\usepackage{enumitem}
\DeclareMathAlphabet{\mathscr}{OT1}{pzc}{m}{it}

\def\O{\mathcal O}

\usepackage{bm}
\usepackage{hyperref}
\usepackage{tablefootnote}

\usepackage{subcaption}

\usepackage{xurl}

\TheoremsNumberedThrough     \ECRepeatTheorems

\EquationsNumberedThrough    

\usepackage{xcolor}
\definecolor{DSgray}{cmyk}{0,1,0,0}

\usepackage{amssymb}

\usepackage{mathabx}

\renewcommand{\hat}{\widehat}
\renewcommand{\tilde}{\widetilde}

\begin{document}

\RUNTITLE{Regret Optimality of SAA for Data-Driven Newsvendor Problems}

\TITLE{Regret Optimality of Sample Average Approximation for Data-Driven Newsvendor Problems: A General Optimization Perspective}

\ARTICLEAUTHORS{
 \AUTHOR{Jiameng Lyu\footnotemark[1]}
 \AFF{Department of Management Science, School of Management, Fudan  University, Shanghai, 200433, China,  \EMAIL{jiamenglyu@fudan.edu.cn}}
 \AUTHOR{Shilin Yuan\footnotemark[1]}
\AFF{School of Management, Huazhong University of Science and Technology, Wuhan, 430074, China, \EMAIL{shilinyuan-research@outlook.com}}
 \AUTHOR{Bingkun Zhou\footnotemark[1]}
 \AFF{Qiuzhen College, Tsinghua University, Beijing 100084, China, \EMAIL{zbk23@mails.tsinghua.edu.cn}}
 \AUTHOR{Yuan Zhou\footnotemark[1]}
 \AFF{Yau Mathematical Sciences Center \& Department of Mathematical  Sciences, Tsinghua University, Beijing 100084, China;Beijing Institute of Mathematical Sciences and Applications, Beijing 101408, China \EMAIL{yuan-zhou@tsinghua.edu.cn}}
} \renewcommand{\thefootnote}{\fnsymbol{footnote}}
\footnotetext[1]{Author names listed in alphabetical order.}

\renewcommand{\thefootnote}{\arabic{footnote}}

\ABSTRACT{Numerous existing studies have examined the performance of Sample Average Approximation (SAA) in the fundamental newsvendor problem. Despite these advances, critical gaps remain in two aspects. First, existing works focus on the linear-cost newsvendor problem and heavily rely on the quantile expression of the optimal solution.  As a result, their analytical methods limit generalizability to more general inventory problems, where the optimal solution is not a quantile of the demand distribution. Second, even within the linear-cost setting, notable gaps exist between the state-of-the-art regret lower bound and upper bound for SAA under various conditions. 

In this paper, we generalize the structure of the newsvendor problem to generic convexity conditions and provide a unified regret analysis of SAA for general sequential stochastic optimization problems. Our approach provides further insights to a broader range of data-driven inventory problems, improves both the upper and lower regret bounds, and establishes the regret rate optimality of SAA. Our lower bound identifies the performance limit achievable by any policy for sequential stochastic optimization and inventory management problems, offering important guidance for future policy design in this area. Moreover, in empirical studies, SAA's performance is frequently used as a benchmark for evaluating new algorithms. The regret rate optimality result provides strong support for its role in assessing other data-driven methods, benefiting both practitioners and researchers. Our new analysis techniques enrich the analysis tools for regret upper and lower bounds for data-driven decision-making problems and other general stochastic optimization problems.
}


\maketitle

\section{Introduction}
\label{sec:Introduction}
The newsvendor problem, as one of the simplest and most fundamental problems in inventory management and capacity management \citep{zipkin2000foundations}, has been extensively applied across various industries, including retailing, manufacturing, and healthcare. In a newsvendor problem, the decision-maker needs to carefully decide the stock or capacity level to minimize costs, including underage costs due to unsatisfied demand and overage costs due to excess inventory or unused capacity.

In retail supermarkets such as Walmart, Whole Foods, perishable items (vegetables, fresh meat, prepared foods) are ordered daily and unsold units quickly lose value. Understocking leads to lost sales, customer dissatisfaction, and potential long-term brand damage, while overstocking causes disposal costs.
In hospital emergency departments, nurse staffing, which is an important problem in capacity management, faces the same trade-off: too few nurses mean long waits, boarding, worsened patient outcomes, and expensive overtime or agency hires, while too many nurses result in idle staff and payroll waste.

In recent years, the \emph{data-driven inventory management} has received great attention \citep{gao2022inventory}, where the demand distribution is unknown in advance, and the decision-maker collects historical data to inform decisions. 
Sample Average Approximation (SAA) is one of the most popular methods \citep{shapiro2021lectures} to address the data-driven inventory problem, due to its simplicity and competitive performance. 
Specifically, the SAA policy constructs an empirical approximation of the cost function using all available historical data and solves the resulting approximate problem to make a decision.

\begin{figure}[h!]
    \centering
\includegraphics[width=0.8\textwidth]{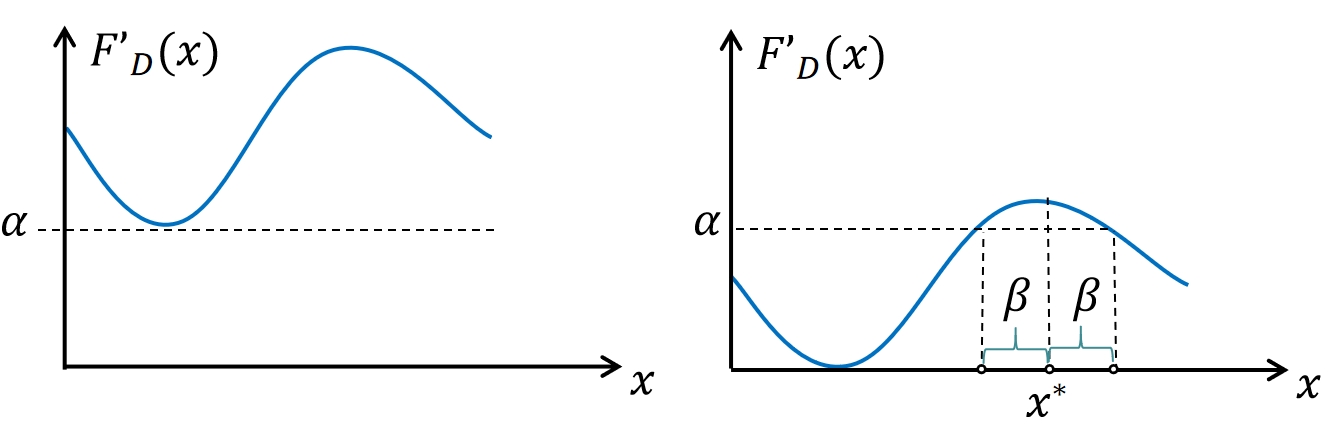}
        \caption{Illustration of Distribution PDFs for Global and Local Minimal Separation Conditions}\label{fig:illustration}
\end{figure}

The stylized nature of the newsvendor problem, together with the power of the SAA method, can significantly enhance our understanding of data-driven inventory management in several key aspects. First, the insights derived from the newsvendor setting are likely to extend to more complicated inventory systems. Second, the performance achieved by the SAA method is often regarded as an ``upper bound'', thus offering a benchmark and potential target for future algorithm design and analysis. Specifically, for general demand distributions, the design and analysis of algorithms evolved from the basic newsvendor model \citep{levi2007nearoptimal} to more complex settings incorporating fixed ordering costs \citep{yuan2021marrying} or positive lead times \citep{zhang2020closing,agrawal2022learning}. Furthermore, a line of research focused on establishing improved algorithm convergence results under well-behaved demand distributions satisfying the  \emph{\(\alpha\)-global minimal separation condition} \citep{huh2009nonparametric}, where the PDF of the demand distribution is at least \(\alpha\) globally within the optimization domain (see the left plot of Figure~\ref{fig:illustration}). These studies also typically began with a deep understanding of the newsvendor model \citep{besbes2013implications} and gradually extended to systems with additional features, such as multi-product systems \citep{shi2016nonparametric,lyu2023minibatch}, perishable systems \citep{zhang2018perishable}, and multi-echelon systems \citep{zhang2021sampling}.

Although data-driven inventory management under the \(\alpha\)-global minimal separation condition is widely studied in the literature, the requirement that the demand density is uniformly bounded from below appears to be rather restrictive. Therefore, a recent work of \cite{lin22datadriven} introduced a \textit{\((\alpha,\beta)\)-local minimal separation condition}, which relaxes the global minimal separation condition. Specifically, it only requires that the density of the demand distribution is bounded below by a positive constant \(\alpha\) within the \(\beta\)-neighborhood of the optimal quantity (see the right plot of Figure~\ref{fig:illustration}). However, contrary to the full development of literature related to \(\alpha\)-global minimal separation condition, the study of data-driven inventory management under \((\alpha,\beta)\)-local minimal separation condition is still lacking. To the best of our knowledge, no paper further studies the performance of algorithms for more general inventory management problems beyond the linear-cost newsvendor problems under the \((\alpha,\beta)\)-local minimal separation condition.

Further investigation of inventory management problems under the aforementioned more general demand distributions may, to some extent, be shaped and constrained by existing research approaches. Existing works mainly focus on the linear-cost newsvendor problem \citep{levi2007nearoptimal,besbes2013implications,lin22datadriven} and heavily rely on the special structure of the optimal solution. Specifically, the optimal solution can be explicitly expressed as a quantile of the demand distribution. Therefore, the analytical methods in these studies are also developed based on the quantile structure of the optimal solution, limiting their generalizability to more general inventory problems, where the optimal solution is generally not a quantile of the demand distribution. 
For example,  for the generalized forms of the newsvendor problems (see Section~\ref{subsec:example-nvp}), the optimal solution would be complicated. 
For these problems, the existing analytical methods are no longer effective and fail to provide further insights, which may be the reason that data-driven investigations for these problems remain underdeveloped.
Additionally, even for the linear-cost newsvendor problem, there is a notable gap between the state-of-the-art regret lower bound for any policies and the regret upper bound for SAA, under both global and local minimal separation conditions. Besides, there is no lower bound result that matches the regret upper bound  $\mathcal O(\log T/\alpha)$ of SAA and SGD\footnote{\cite{huh2009nonparametric} established the regret upper bound $\mathcal O(\log T/\alpha)$ for their SGD-based algorithm under the $\alpha$-global minimal separation condition.} simultaneously for the parameter $\alpha$ and the time horizon $T$ under the $\alpha$-global minimal separation condition. Moreover, even without considering \(\alpha\), we find that the proof of the widely cited \(\Omega (\log T)\) lower bound result in~\cite{besbes2013implications} may be flawed due to a subtle but critical technical oversight.\footnote{In Appendix~\ref{sec:ub-uniform}, we design an algorithm achieving $\O(1)$ regret under the hard instances constructed in their proof, therefore contradicting their $\Omega(\log T)$ lower bound. We also present a detailed discussion about the subtle technical oversight in their proof both theoretically and numerically.}

Our objective is to investigate data-driven newsvendor problems from a more general perspective on structural properties of cost functions, and take a step toward a deeper and more precise understanding of SAA's performance in data-driven inventory management problems.
Our approach is motivated by the following observation: for linear-cost newsvendor problems, the $\alpha$-global minimal separation condition 
implies the $\alpha$-global strong convexity of the cost function up to a scaling constant. It is the convexity structure (influenced by demand distributions) behind the cost function that fundamentally shapes the performance of algorithms. This observation motivates us to investigate the performance of SAA under broader convexity conditions, moving beyond reliance on problem-specific closed-form solutions.
 From the above research perspective, we propose new analytical methods that go beyond the linear-cost newsvendor model, making them applicable to a wider range of sequential stochastic optimization problems. 
Moreover, we address the regret gaps in the existing literature and establish the regret rate optimality of SAA. 

\subsection{Our Contributions}
In the following, we first provide a summary of our main contributions, and then we present a more detailed discussion of each point. 

\begin{table}[t]
\caption{Comparison between our results and the results in literature.}
\label{tab1}
\renewcommand{\arraystretch}{1.3}
\begin{center}
\scalebox{0.8}{\begin{tabular}{l|l|l|l|l}
\hline & \makecell[l]{Lower Bound\tablefootnote{The lower bound results are all established with the hard instance of linear cost newsvendor problems.}} & \makecell[l]{Upper Bound} & \makecell[l]{Overage/Underage\\ Cost}  & \makecell[l]{Setting} \\ \hline

\cite{besbes2013implications}      & $\Omega \left(\log T\right)$    & ---  &  Linear & $\alpha$-Global Minimal Separation            \\ \hline

       \cite{lin22datadriven}      & $\Omega \left(\frac{1}{\alpha}\right)$     & $\mathcal O\left(\frac{1}{\alpha}\log T\right)$ &  Linear &   $\alpha$-Global Minimal Separation        \\ \hline    
\textbf{Our results} & \bm{$\Omega \left(\frac{1}{\alpha}\log T\right)$}     &  {$\bm{\O\left(\frac{1}{\alpha}\log T\right)}$}   & \textbf{Convex} &   $\alpha$-Global Strong Convexity       \\ \hline
\cite{lin22datadriven}      & $\Omega \left(\frac{1}{\alpha\beta}\right)$     & $\mathcal O\left(\frac{1}{\alpha\beta}\log T\right)$  &  Linear&   $(\alpha,\beta)$-Local Minimal Separation     \\ \hline
\textbf{Our results} & \bm{$\Omega \left(\frac{1}{\alpha}\log T+\frac{1}{\alpha\beta}\right)$}     & \bm{$\mathcal O\left(\frac{1}{\alpha}\log T+\frac{1}{\alpha\beta}\right)$}  & \textbf{Convex} &   $(\alpha,\beta)$-Local Strong Convexity       \\ \hline
\end{tabular}}
\end{center}
\end{table}
\noindent\textbf{Summary of main contributions.} 
In this paper, we generalize the structure of newsvendor problem to generic convexity conditions and provide a unified regret analysis of SAA for general sequential stochastic optimization problems under different convexity conditions including {general convexity}, {$\alpha$-global strong convexity} and $(\alpha,\beta)$-local strong convexity\footnote{The $(\alpha,\beta)$-local strong convexity captures that $C(x)$ is strongly convex on a small neighborhood around the optimal solution,  abstracting the local minimal separation condition on demand to a general local strong convexity condition on objective functions. }, thereby providing further insight into other data-driven inventory problems. We advance both the upper and lower bounds for SAA (see Table~\ref{tab1} for a summary of our results and comparisons) and resolve the above key gaps in the literature by proposing new techniques that differ significantly from existing analyses relying on the closed-form expression of the optimal solution.
Our new analysis techniques enrich the analysis tools of both regret upper and lower bounds in data-driven inventory management literature and have independent interests for more general stochastic optimization problems.

\noindent {\bf  Contribution I: general optimization perspective.} We explore the structural properties of the linear-cost newsvendor model under different conditions on demand distributions and abstract them into corresponding convexity assumptions, thereby formulating a general sequential stochastic optimization problem. This framework enables our subsequent analysis to move beyond the specific newsvendor problems. Consequently, our analysis can capture how general structural properties of the objective function influence the problem, and also offer insights into further research on data-driven inventory management. This framework models a general sequential stochastic optimization problem, which exhibits both global and local strong convexity in the objective function. 
Reducing to the linear-cost newsvendor problem, the framework covers the problem under local and global minimal separation conditions \citep{lin22datadriven}, as well as a variety of other classical overage/underage cost formulations (see Section~\ref{subsec:example-nvp} for more discussions). 

We establish the regret rate optimality of SAA for the general sequential stochastic optimization problem under both global and local strong convexity of the cost objective function as follows.

\noindent {\bf  Contribution II: regret analysis techniques and insights.} For the aforementioned general optimization problem, we prove that SAA enjoys \(\mathcal{O}(\log T/\alpha + 1/ (\alpha\beta))\) regret under the \((\alpha,\beta)\)-local strong convexity condition.  Compared with the upper bound $\O(\log T/(\alpha\beta))$ established by \cite{lin22datadriven}, our result demonstrates that the regret of SAA is only influenced by $\alpha$ in the long run, and the width $\beta$ only affects the regret in the early periods.  This insight enhances our understanding of how the local convexity properties of objective functions affect the long-term regret performance of decision-making strategies. 

Regarding the analysis techniques, our approach is different from the existing methods that either study general SAA theory by approximating objective function $C(x)$ (e.g., \cite{shapiro2021lectures}) or utilize the quantile properties of newsvendor problems (e.g., \cite{levi2007nearoptimal,lin22datadriven}).  In particular, we introduce a new proof technique of \textit{gradient approximation}.  Specifically, we analyze the behavior of the true gradient of $C(x)$ at the SAA solution, by relating it to the gradient of the empirical cost function (i.e., the gradient approximation). Together with the local strong convexity condition, we obtain that the SAA decisions will fall into $[x^*-\beta,x^*+\beta]$ with a high probability for large $t$, which enables us to bound the regret leveraging the strong convexity. Based on this observation, we divide the regret into two parts according to whether the SAA decisions fall within or outside the specified interval, and bound them individually by relating the regret to the norm of the gradient.

\noindent {\bf  Contribution III: hard instances and lower bound.} 
We prove that the worst-case regret of any data-driven method for the sequential stochastic optimization problems is lower bounded by \(\Omega(\log T/\alpha)\) and \(\Omega(\log T/\alpha +1/(\alpha\beta))\) under global and local strong convexity, respectively.
This result is the first to match the existing upper bound with respect to both the parameter \(\alpha\) and the time horizon \(T\) simultaneously under the \(\alpha\)-global strong convexity condition, and thus advances the theoretical understanding of the limit of data-driven methods in sequential stochastic optimization problems and the newsvendor problems.\footnote{Given the possible flaw in the proof of the \(\Omega (\log T)\) lower bound in~\cite{besbes2013implications}, our result may also be the first sound \(\Omega (\log T)\) lower bound without considering the parameter \(\alpha\).} 

As in many existing lower bound proofs, the most critical part of our proof is the design of hard instances, which in our case is a family of parameterized demand distributions.  To achieve the desired lower bound \(\Omega(\log T/\alpha)\), the main challenge is to achieve two seemingly conflicting objectives while meeting several basic requirements such as global minimal separation condition and distribution regularity.  First, to incur a large regret, the optimal solutions should be sensitive to changes in the unknown problem parameter. Second, to avoid easy learning of the problem parameters, the demand distributions should exhibit a stable response to parameter shifts. To achieve these two objectives, our hard instances feature a uniquely designed smooth inverted-hat shape, which is notably distinct from the existing hard instances in inventory management literature. We believe that our lower bound proof enriches the lower bound techniques in data-driven inventory management literature and may contribute to understanding the tight regret bounds of broader data-driven problems as a function of both the time horizon \(T\) and other problem parameters.

\subsection{Related Works}
In this section, we introduce literature related to our paper and discuss how our paper is appropriately placed into contemporary literature by giving comparisons with
closely related existing works.

\textbf{SAA for inventory management.}
Motivated by the widespread adoption of SAA in newsvendor problems, several studies have concentrated on enhancing the theoretical understanding of this method.
\cite{levi2007nearoptimal}  investigated the sample size required to ensure a specific accuracy level. Specifically, they show that when the number of demand samples is $\Omega\left(\left(1/\epsilon^2\right)\log\left(1/\delta\right)\right)$, SAA can guarantee that the expected cost of the solution is within $(1+\epsilon)$ times the clairvoyant optimal expected cost with probability at least $1-\delta$.
\cite{levi2015data} improved the sample complexity result in \cite{levi2007nearoptimal} and found a weighted mean spread effect of demand distribution on the accuracy of SAA.
\cite{lin22datadriven} established several state-of-the-art theoretical results for linear-cost newsvendor problems under the \((\alpha,\beta)\)-local minimal separation condition. However,  the regret analysis in \cite{lin22datadriven} is tailored for newsvendor problems with linear overage and underage costs. It does not work for the more general convex overage and underage costs. Besides,
    there is a notable gap between their regret lower bound $\Omega(1/(\alpha\beta))$ for any data-driven policy and their regret upper bound $\mathcal O(\log T/(\alpha\beta))$ for SAA, under the $(\alpha,\beta)$-local minimal separation condition.
Recently, \cite{chen2024survey} (released shortly after our initial released preprint) extends the $(\alpha,\beta)$-minimal separation condition to a more refined $(\alpha,\beta,\zeta)$-clustered class framework, where the parameter $\zeta$ captures the transition between general convexity ($\zeta = \infty$) and strong convexity ($\zeta = 0$). Their focus is on characterizing the optimal regret bounds with respect to $\zeta$ and the time horizon $T$ under linear inventory costs, emphasizing how $\zeta$ influences the convergence rate as $T$ increases.
In contrast, our work primarily investigates the optimal regret bounds with respect to $\alpha$, $\beta$, and $T$. In this version, we extend our analysis to study the $(\alpha, \beta, \zeta)$-clustered setting using our methodology, establishing the optimal regret bounds with respect to $\zeta$ and $T$ under general inventory cost (see Section \ref{sec:clustered} for details).

\cite{besbes2023big} derived exact regret bounds to characterize the number of offline samples needed for a given accuracy. For the classical linear-cost newsvendor problem, they show that SAA, which selects the $\lceil qn\rceil$-th order statistic, can be suboptimal in finite samples: the minimax-optimal policy randomizes between two neighboring order statistics to balance the two worst-case Bernoulli regimes.
This finite-sample gap is tied to their distributional class. Their worst-case Bernoulli demand distributions may have widely separated neighboring order statistics, and the expected cost is not locally strongly convex around the optimum. In contrast, our $(\alpha,\beta)$-local strong convexity condition rules out such degenerate local geometry. Our work and \cite{lin22datadriven, chen2024survey} instead establish the regret rate optimality of SAA by matching upper and lower regret bounds for various distribution classes.

\textbf{Sampling-based methods for inventory management.}  There are also various papers studying different data-driven inventory problems with sampling-based methods. For example, \cite{cheung2019sampling} and \cite{zhang2021sampling} studied the sample complexity of SAA for capacitated and multi-echelon serial inventory systems, respectively. \cite{qin2023sailing} established 
the optimal sample complexity for the basic multi-period inventory control problems. \cite{qin2022data} studied the sample complexity of joint pricing and inventory control problems. \cite{wang2023hybrid} proposed an algorithm that integrates SAA and gradient descent and applied it to two complex inventory systems successfully. The seminal work of \cite{keskin2023nonstationary} was the first to provide regret guarantees for data-driven learning in the nonstationary newsvendor problem and proposed several sampling-based two-stage estimation policies.
More recently, \cite{chen2025learning} developed a distributional-detection-and-restart framework for nonstationary newsvendor problems under both uncensored and censored demand.
Their fully adaptive algorithms handle both abrupt and gradual distributional changes without prior knowledge of the degree or type of nonstationarity.

\textbf{Other related works on data-driven inventory management.}
For lost-sales systems with zero lead time, the seminal work of \cite{huh2009nonparametric} introduced a stochastic gradient descent (SGD)-based algorithm for single-product settings. This approach was extended to multi-product systems by \cite{shi2016nonparametric}. Adopting a novel technical perspective, \cite{lyu2023minibatch} proposed a minibatch-SGD-based meta-policy applicable to various inventory systems with myopic optimal policies. Other notable contributions include: \cite{yuan2021marrying} for systems with demand learning and fixed ordering costs; \cite{zhang2018perishable} for perishable inventory systems with fixed lifetimes; \cite{chen2020optimal} for backlogging systems with random production capacity; \cite{gong2023bandits} for online models with unknown cyclic demand distributions; and \cite{gao2022efficient} for multi-product systems with customer choice considerations. Additionally, based on the Operational Data Analytics (ODA) framework introduced in \cite{feng2023framework}, \cite{feng2025contextual} studied the contextual newsvendor problem using data-integrated methods, and \cite{chu2023solving} studied the pricing-setting newsvendor problems with parametric models.
Besides, several papers have investigated lost-sales inventory systems with positive lead times under unknown demand distributions. Since computing the optimal policy is computationally intractable, existing literature has focused on asymptotic heuristics as benchmarks. A series of works by \cite{huh2009adaptive,zhang2020closing,agrawal2022learning} adopted the best base-stock policy as their benchmark. \cite{lyuucb} developed Upper Confidence Bound (UCB)-type learning algorithms to learn the optimal capped base-stock and base-stock policies.  \cite{chen2022learning,chen2023learning} used the optimal constant-order policy as their benchmark. Specifically, \cite{chen2022learning} examined systems with both positive lead times and random supply, while \cite{chen2023learning} studied systems with stochastic lead times and unknown random supply functions. \cite{chen2023learning} designed bandit optimization algorithms and established the optimal regret by utilizing the hidden convexity of the cost function. 
For the lower bound results of data-driven newsvendor problems, \cite{zhang2018perishable} established a lower bound $\Omega (\sqrt{T})$ of any learning algorithm under the general continuous demand distribution. Their hard instance construction is inspired by the lower bound result in \cite{besbes2013implications} for discrete demand distribution.  \cite{besbes2013implications} also established a lower bound $\Omega(\log T)$ for continuous demand distribution under the \(\alpha\)-global minimal separation condition (see their Theorem 1). Their lower bound does not consider the dependence on parameter \(\alpha\). Please refer to Appendix \ref{sec:ub-uniform} for a detailed discussion of the $\Omega(\log T)$ lower bound result in \cite{besbes2013implications}.

\subsection{Organization}
The remainder of this paper is organized as follows.
In Section~\ref{sec:problem formulation}, we introduce the problem formulation. We first present the SAA framework for the general sequential stochastic optimization problem, and then illustrate its application to the newsvendor problems.
In Section~\ref{sec:Upper bound of SAA}, we establish upper bounds on the performance of the SAA method with a new analytic method. We present the main results, provide the intuitions and insights behind the proof, and discuss potential extensions.
In Section~\ref{sec:lower bound}, we derive minimax regret lower bounds under both global convexity and local strong convexity. Our analysis focuses on the newsvendor problem with linear inventory costs, which serves as a representative and challenging case.
In Section~\ref{sec:extension}, we extend our gradient-approximation-based approach to a broader function class and explore relaxations of several technical assumptions.
In Section~\ref{sec:numerical}, we conduct numerical experiments to evaluate the empirical performance of the SAA method in various settings and to validate the predicted dependence of regret on parameters $\alpha$ and $T$.
Finally, Section~\ref{sec:Conclusion} concludes the paper and outlines directions for future research.
\section{Problem Formulations}\label{sec:problem formulation}

Motivated by the data-driven repeated newsvendor problem over $T$ periods, we consider the following general sequential stochastic optimization problem over a compact convex set $\mathcal{X}$
\begin{align}
\label{eq:sto-prog}
     \min_{x \in \mathcal{X}}C(x)= \E_D[c(x,D)], 
\end{align}
where $c(x,d)$ is the cost when the decision is $x$ and random variable $D$ is realized as $d$.

We assume in period $t \in [T]$, the decision-maker has access to the historical data set $\mathcal{H}_t=\{d_1,\dots,d_t\}$ with $t$ samples \emph{i.i.d.}~drawn from distribution $D$. The goal of the decision-maker is to design an admissible policy $\pi = (\pi_1,\pi_2,\dots) $ to minimize the total expected costs, where $\pi_t$
maps $\mathcal{H}_t$ to order quantity $x_t^{\pi}$ in period $t$. Let $\Pi$ be the set of admissible policies. For a policy $\pi\in\Pi$, we measure its performance by regret, defined as 
\begin{align}
        \nonumber \mathcal{R}^{\pi}\left(F_D,T\right)=\sum_{t=1}^T\mathbb{E}\left[C\left(x_t^\pi\right)-C\left(x^*\right)\right],
\end{align}
where $x^*= \argmin_{x\in \mathcal{X}}C(x)$ and the expectation is taken over the distribution of $D$ (with cumulative distribution function $F_D$) and the randomness of $\pi$.

\begin{remark}[Choice of benchmark]
This paper chooses the cumulative regret as the performance measure of algorithms merely to keep consistency with existing literature \citep{huh2009nonparametric,besbes2013implications,lin22datadriven}. For other performance metrics, our analysis techniques can also establish corresponding results. For sample complexity of the SAA method, we also establish the results as an extension; please refer to Section \ref{subsec:sample-compexity} for details. 
\end{remark}

As mentioned earlier, this paper studies the widely adopted SAA method for sequential stochastic optimization problems.  In this method, the expected cost function $C(x)$ is approximated by the sample average cost as 
\[
\hat{C}_t(x)=\frac{1}{t}\sum_{i=1}^tc(x,d_i),
\]
using historical data set $\mathcal{H}_t=\{d_1,\dots,d_t\}$,
and we solve $\min_{x \in \mathcal{X}} \hat{C}_t(x)$ to obtain an approximate solution. Therefore, under the SAA method, the decision in period $t$ is given by
\begin{equation}
    \label{eq:saa-xt-def}
    \hat{x}_t=\argmin_{x \in \mathcal{X}} \hat{C}_t(x).
\end{equation}

The above general formulation covers a wide range of applications where samples are observable in sequential stochastic optimization. For instance, in the literature on online statistical learning, including online regression, online estimation, and online classification, the sample is typically assumed to be observable. This paper primarily focus on its application to newsvendor problems, which are fundamental in operations management. Next, we provide more motivations and examples.

\subsection{Modeling Motivations and Representative Newsvendor examples}\label{subsec:example-nvp}

The general formulation in Eq.~\eqref{eq:sto-prog} is motivated by several practically important extensions of the classical newsvendor problem. In particular, it naturally accommodates both (i)~nonlinear convex cost structures and (ii)~service-level considerations. We discuss these two modeling motivations below.

\textbf{Motivation I: Convex cost structures.}
A key reason for adopting the general formulation is that, in many inventory and production systems, marginal costs are not constant. The classical linear overage and underage model assumes that each additional unit of excess inventory or unmet demand incurs the same cost. However, this assumption is often too restrictive in practice. For example, warehouse congestion or limited storage capacity may cause marginal holding costs to increase with leftover inventory; expediting fees, lost goodwill, or contractual penalties may make shortage costs escalate with the size of unmet demand; and production costs may become increasingly expensive because of overtime labor, capacity stress, or equipment wear. These features are naturally captured by convex cost functions, including quadratic and piecewise linear forms; see, e.g., \cite{porteus1990stochastic,zipkin2000foundations,rosling2002inventory}.

Under this motivation, Eq.~\eqref{eq:sto-prog} covers several representative newsvendor-type models.

\begin{itemize}
    \item \textbf{Convex inventory cost.}
    The cost function includes both overage and underage costs:
    \[
    c(x,d)= h_f((x-d)^+)+b_f((d-x)^+),
    \]
    where $h_f(\cdot)$ and $b_f(\cdot)$ are convex overage and underage cost functions, respectively.

    \item \textbf{Linear inventory cost.}
    As a special case of the above model, the classical newsvendor cost is
    \[
    c(x,d)= h(x-d)^+ + b(d-x)^+,
    \]
    where the overage and underage costs are linear with coefficients $h$ and $b$, respectively.

    \item  \textbf{Convex production cost.} $c(x,d) = c_p(x) - p\min\{x,d\}$, where $c_p(\cdot)$ is the convex production cost function (e.g. linear, quadratic costs) and $p\min\{x,d\}$ is the realized revenue.
\end{itemize}

\begin{remark}[Specification of model and parameters]
More specific nonlinear cost models, such as quadratic and piecewise linear inventory formulations, can be found in \citep{porteus1990stochastic,lu2014inventory}. From a practical perspective, the parameters in such models can be calibrated using standard operational and accounting data. For instance, linear holding and shortage coefficients can be estimated from warehousing, capital, expediting, penalty, or lost-margin data, while coefficients in quadratic or piecewise linear specifications can be fitted from historical cost records against surplus or shortage levels.    
\end{remark}

\textbf{Motivation II: Service-level considerations.}
A second motivation for Eq.~\eqref{eq:sto-prog} is that it also allows service-level objectives or requirements to be incorporated in a tractable way. In many applications, firms do not optimize cost alone, but also aim to maintain a target service level, such as fill rate or the probability of meeting demand. Let $h(x)$ denote a convex baseline inventory cost, and let $s(x)$ be a service measure. The hard constraint of service level is $s(x) \geq \eta$, where $\eta$ is a target service level. Since the hard service-level constraint is usually computationally hard, one can reformulate it as a soft service-level requirement \citep{hopp1997easily}, and a common formulation is
\[
\min_{x\in\mathcal{X}} \; h(x)-\lambda\bigl(s(x)-\eta\bigr), \qquad \lambda\ge 0,
\]
where $\lambda$ is a penalty parameter selected by the decision-maker. Since improvements in service often exhibit diminishing returns as inventory increases, $s(x)$ is frequently concave in $x$. Under standard conditions, the resulting objective remains convex and therefore still fits within our general framework.

Overall, these examples show that Eq.~\eqref{eq:sto-prog} is not merely an abstract generalization of the classical linear-cost newsvendor model. Rather, it provides a unified formulation for a broad class of practically relevant inventory problems, including both convex cost extensions and service-level-driven decision models.

\subsection{Global and Local Convexity Conditions}
\label{subsec:cvx-condition}
Motivated by the linear-cost newsvendor problem under global and local minimal separation conditions mentioned in the introduction section, we introduce two convexity conditions on $C(x)$ as follows.

\begin{definition}[Global strong convexity]\label{def:global}
We say that $C(x)$ satisfies the $\alpha$-global strong convexity condition 
if there exists a constant $\alpha > 0$ such that for all $x,y \in \mathcal{X}$,
\[
C(y) \ge C(x) + g(x)(y-x) + \frac{\alpha}{2}(y-x)^2,
\]
where $g(x)$ is a subgradient of $C$ at $x$.
\end{definition}

Recall the \textit{$\alpha$-global minimal separation condition} that the density of the demand distribution is bounded from below by a positive constant $\alpha$ for all $x\in[0,\bar D]$, i.e., $F'_D(x) \geq \alpha$, $x \in [0,\bar{D}]$ for some $\bar{D}$. For the newsvendor problem with linear inventory cost, $C''(x) = (h+b)F'(x)$, and the $\alpha$-global minimal separation condition is equivalent to the $\alpha'$-strong convexity for some $\alpha'$. Therefore, Definition \ref{def:global} abstracts the $\alpha$-global minimal separation condition on demand to general objective functions.

Although the $\alpha$-global minimal separation condition on $D$ has been widely investigated in the inventory management literature (see, e.g., \cite{huh2009nonparametric,shi2016nonparametric,zhang2018perishable,lyu2023minibatch}), it seems to be a little bit restrictive and can not be satisfied by a wide range of demand distributions. Therefore, \cite{lin22datadriven} proposed a weaker assumption known as \textit{$(\alpha,\beta)$-local minimal separation condition}, which specifies that the density of the demand distribution is bounded below by a positive constant \(\alpha\) within a \(\beta\)-neighborhood of the newsvendor solution $x^*$. Specifically, $F'_D(x) \geq \alpha$, $x \in [x^*-\beta,x^*+\beta]$. Under this weaker condition on demand and linear inventory cost, the objective function $C(x)$ is $\alpha'$-strongly convex on a small interval $[x^*-\beta,x^*+\beta]$.

Inspired by the $(\alpha,\beta)$-local minimal separation condition in \cite{lin22datadriven}, we abstract the local minimal separation condition on demand to a general local strong convexity condition on objective functions.
\begin{definition}[Local strong convexity]\label{def:local}
   We say $C(x)$ satisfies the $(\alpha,\beta)$-local strong convexity condition, if there exist constants $\alpha,\beta >0$, such that $C(x)$ is $\alpha$-strongly convex on $ \mathcal{X} \cap \mathcal{N}(x^*,\beta) $, where $x^*$ is the unique minimum of $C(x)$ over $\mathcal{X}$ and $\mathcal{N}(x,r) :=\{y \mid \Vert y-x\Vert_2 \leq r\}$.
\end{definition}

Local strong convexity is a broadly applicable condition that arises naturally in a wide variety of inventory cost structures. It can be shown that under the linear inventory cost, the objective function satisfies local strong convexity whenever the demand distribution satisfies the local minimal separation condition \cite{lin22datadriven}. More generally, local strong convexity might hold for cost functions that are locally convex inherently, a property shared by many classical and practically relevant loss functions. To further illustrate this universality, we consider the Huber-type inventory cost, which originates from robust statistics \cite{huber1964robust,huber1981robust} and interpolates between quadratic and linear penalties.  Such a quadratic-to-linear transition is naturally motivated in inventory and production contexts: small mismatches between the order quantity and realized demand can be absorbed through internal operational adjustments such as overtime, expedited handling, or transient storage, whose marginal cost rises with the mismatch size and yields a quadratic regime; once the mismatch exceeds the capacity of these adjustments, the firm falls back on external remedies with a fixed marginal per-unit cost such as expediting fees, backorder charges, or lost-sale costs, yielding a linear regime. The Huber-type inventory cost takes the form
\[
        c\left(x,d\right)=\left\{\begin{aligned}
		&b\delta \left(d-x-\delta/2\right),&x\in\left(-\infty,d-\delta\right],\\
                &b\left(x-d\right)^2/2,&x\in\left[d-\delta,d\right],\\
                &h\left(x-d\right)^2/2,&x\in\left[d,d+\delta\right],\\
                &h\delta \left(x-d-\delta/2\right),&x\in\left[d+\delta,+\infty\right).
                \end{aligned}
            \right.
\]
Here $b,h>0$ are positive coefficients governing the underage and overage penalties in the cost function, respectively, and $\delta>0$ is a fixed constant (rather than a function) that specifies the width of the quadratic region. As one concrete illustration, suppose $D\sim U[a,a+L]$, where $a$ is the lower endpoint of the support of the uniform distribution and $L>2\delta$ is its length. We show in Appendix~\ref{subsec:huber-lsc} that the resulting expected cost $C(x)=\mathbb{E}[c(x,D)]$ is locally strongly convex around its unique minimizer $x^\ast$ while failing to be globally strongly convex; similar arguments extend to other demand distributions. Intuitively, because the Huber cost is quadratic only within a $\delta$-neighborhood of the realized demand and linear outside it, taking the expectation preserves sufficient curvature near $x^\ast$ but not away from the optimum.

The $(\alpha,\beta)$-local strong convexity condition captures that $C(x)$ is strongly convex on a small neighborhood around the optimal solution. 
Intuitively, the performance of algorithms will be substantially influenced by the value of $(\alpha,\beta)$ that captures the local strong convexity. In the remainder of this paper, we investigate how these two parameters affect the performance of algorithms. Specifically, we establish matching upper and lower bounds on regret in terms of $T$, $\alpha$, and $\beta$.

\section{Upper Bound of SAA Method}\label{sec:Upper bound of SAA}

In this section, we establish upper bounds on the performance of the SAA method. We first present the main result and the intuition behind the proof. Since the main motivation of this paper stems from newsvendor problems, we focus our analysis on one-dimensional $\mathcal{X}$. This allows us to simplify the notation and highlight the core ideas of the analysis. It is worth noting that similar proof can also be generalized to the multi-dimensional case. We discuss the application of general theory to specific newsvendor problems in Section \ref{subsec:app-newsvendor}, and the detailed proof is presented in Section \ref{subsec:proof-ub}.

Besides the key global and local strong convexity assumptions (Definitions \ref{def:global} and \ref{def:local}), we also impose some regular assumptions on problem Eq.~\eqref{eq:sto-prog} for the regret analysis as follows. 

\begin{assumption} 
\label{ass:regular}
There exist non-negative constants $B$ and $C_1$ such that
\begin{enumerate}[label=\arabic*., ref=\ref{ass:regular}.\arabic*]
    \item \label{ass:regular-1} \textbf{Sub-differentiability and boundedness.} For any realization $d$ of $D$, $c(x,d)$ is sub-differentiable. Moreover, subgradient function $c'(\cdot,d)$ is bounded as  $|c'(\cdot,d)|\leq B$ for any realization $d$.
    \item \label{ass:regular-2} \textbf{Near-optimality of SAA solution.} Define $\hat{g}_t(x) = \frac 1t\sum_{i=1}^t c'(x,d_i)$ as the approximation of $g(x) :=C'(x)$, where we assume $C(x)$ is differentiable. The optimal solution to problem Eq.~\eqref{eq:sto-prog} can be obtained in the interior of $\mathcal{X}$ and SAA solution $\hat{x}_t$ given by Eq.~\eqref{eq:saa-xt-def} satisfies that $|\hat{g}_t(\hat{x}_t)| \leq C_1/\sqrt{t}$ w.p.$1$.
    \item \label{ass:regular-3} \textbf{Convexity of the inner-layer function.} Function $c(x,d)$ is convex in $x$ for any realization $d$.
\end{enumerate}
\end{assumption}

Note that the above assumptions are mild and easy to verify for specific problems, and we will discuss the verification of these assumptions in Section \ref{subsec:app-newsvendor}. We make some remarks on the assumptions.

\begin{remark}[Near-optimality of SAA solution]

For Assumption \ref{ass:regular-2}, if $c'(\cdot,d)$ is continuous and the optimal solution is obtained in the interior of $\mathcal{X}$, by the optimality condition, we can choose $C_1=0$, i.e., $\hat{g}_t(\hat{x}_t)=0$. 
For certain newsvendor problems, the cost function might be piecewise linear (e.g., linear cost function), whose derivative might be discontinuous. For these problems, the $\hat g_t(x)$ is piecewise continuous with $\O(1/t)$-jump. Therefore, we can find $C_1$ such that $|\hat{g}_t(\hat{x}_t)| \leq C_1/\sqrt{t}$; see Section \ref{subsec:app-newsvendor}.

\end{remark}

\begin{remark}[Convexity of inner-layer function]
Note that under the (global or local) strong convexity conditions, the convexity of the inner function $c(\cdot,d)$ seems to be a redundant assumption. However, it is remarkable that this assumption plays a role in two key aspects. First, the convexity of the inner function facilitates the derivation of uniform concentration bounds for the gradient function $g(x)$. Second, in the absence of strong convexity, the convexity of the inner function implies the convexity of $C(x)$, which allows us to establish regret bounds under general convexity conditions; see Remark \ref{rem:general-cvx}.
\end{remark}

The following theorem provides an upper bound on the regret of the SAA method.
\begin{theorem}\label{tm:tm_main_regret_ub}
Suppose Assumption \ref{ass:regular} holds, and $C(x)$ satisfies the $(\alpha,\beta)$-local strong convexity condition (Definition \ref{def:local}).
The regret of the SAA method is upper bounded as
\begin{align}
\sum_{t=1}^T\mathbb{E}\left[C(\hat{x}_t)-C\left(x^*\right)\right]\leq K_1 +K_2\cdot \frac{1}{\alpha\beta}+K_3 \cdot \frac{\ln T}{\alpha},\nonumber
\end{align}
where $C_0$ is the constant in Lemma~\ref{technical lemma concentration}, and we define $\mathcal{D}_{\mathcal{X}}:=\max_{x_1,x_2 \in \mathcal{X}} \Vert x_1-x_2\Vert$, and $K_4 =: (4C_0B+C_1)^2+2B^2+\sqrt{2\pi}B(4C_0B+C_1)$, then we can set $K_1= 2\mathcal{D}_{\mathcal{X}}\sqrt{K_4}$, $K_2 = 8\sqrt{2}\mathcal{D}_{\mathcal{X}}\sqrt{K_4}(4C_0 B +C_1)+ 8\mathcal{D}_{\mathcal{X}}\sqrt{K_4}/(4C_0B+C_1)$, $K_3 = 2K_4$.
\end{theorem}
\begin{remark}[Conclusion under the global strong convexity condition]
        Note that if $\beta \geq \mathcal{D}_{\mathcal{X}}$, we have $\mathcal{X} \subset\mathcal{N}(x^*,\mathcal{D}_{\mathcal{X}}) $. Therefore, the $(\alpha,\beta)$-local strong convexity condition covers the $\alpha$-global strong convexity condition when $\beta = \mathcal{D}_{\mathcal{X}}$. Under the $\alpha$-global strong convexity condition, we could apply Theorem~\ref{tm:tm_main_regret_ub} and set $\beta = \mathcal{D}_{\mathcal{X}}$ to obtain a regret upper bound of $\O(\log T/\alpha)$ for the SAA method. 
\end{remark}

\begin{remark}[combination with the general convex case]
\label{rem:general-cvx}
In  Section~\ref{sec:osqrtT}, we also establish the $\mathcal{O}(\sqrt{T})$ regret upper bound for the general convex $C(x)$, i.e., without any strong convexity assumption. Combining the $\mathcal{O}(\sqrt{T})$ result with the above Theorem~\ref{tm:tm_main_regret_ub}, we have $\sum_{t=1}^T\mathbb{E}\left[C(\hat{x}_t)-C\left(x^*\right)\right]\leq \O(\min\{\sqrt{T},1/\alpha\beta+\log T/\alpha\})$. Similarly, under the global strong convexity, we have $\sum_{t=1}^T\mathbb{E}\left[C(\hat{x}_t)-C\left(x^*\right)\right]\leq \O(\min\{\sqrt{T},\log T/\alpha\})$.
\end{remark}

Compared with the $\O(\log T/(\alpha\beta))$ upper bound established by \cite{lin22datadriven} (for linear-cost newsvendor problems), our result not only works for general sequential stochastic optimization problems, but also better reflects the effect of parameters $\alpha,\beta$. The term $\O(\log T/\alpha)$ shows that regret is only influenced by $\alpha$ in the long run (as $T \to \infty$), i.e., how strongly convex is $C(x)$ in the small neighborhood $\mathcal{N}(x^*,\beta)$. Moreover, the term $\O( (\alpha\beta)^{-1})$ shows the width $\beta$ only affects the regret in the early periods.  

Our proof techniques can better illustrate the above observations. Different from the existing methods that either study general SAA theory by approximating the objective function $C(x)$ \citep{shapiro2021lectures} or utilize the quantile properties of newsvendor problems \citep{levi2007nearoptimal,lin22datadriven}, our proof is based on a new idea of \textit{gradient approximation} as follows. The detailed proof can be found in Section~\ref{subsec:proof-ub}.

\subsection{Main Idea of Regret Analysis: Gradient Approximation} 

Since the optimal solution is obtained in the interior of $\mathcal{X}$, we know that $g(x^*)=0$. By the strong convexity within $\mathcal{X} \cap\mathcal{N}(x^*,\beta)$, we know that $|g(x)| \geq \alpha\beta$ for all $x \in \mathcal{X}$ but not in $\mathcal{N}(x^*,\beta)$.

By establishing the uniform concentration of $\hat{g}_t(\cdot)$ around $g(\cdot)$ and the definition of $\hat{x}_t$ given by SAA, we have $|g(\hat{x}_t)| \approx |\hat{g}_t(\hat{x}_t)|  = \O(1/\sqrt{t})$. 
 Therefore, as the local strong convexity implies that $|g(x)| \geq \alpha\beta$ for all $x \in \mathcal{X}$, we know that $\hat{x}_t$ will enter $\mathcal{N}(x^*,\beta)$ when $t$ grows large. The observation motivates us to utilize the strong convexity of $C(x)$ in $\mathcal{N}(x^*,\beta)\cap \mathcal{X}$ to improve the convergence rate of the SAA method.

Specifically, for each $t$, we define the good event that $\mathcal{B}_t=\left\{\left|g(\hat{x}_t)\right|\leq \alpha\beta\right\}$,
which is related to whether $\hat{x}_t$ falls into $\mathcal{N}(x^*,\beta)\cap \mathcal{X}$. Then we decompose regret in period $t$ into
\begin{equation}
\label{eq:regret-decompose}
I_{1,t}=\mathbb{E}\left[\left(C(\hat{x}_t)-C\left(x^*\right)\right)\mathbb{I}\left[\mathcal{B}_t\right]\right] \text{~~~and~~~}
I_{2,t}=\mathbb{E}\left[\left(C(\hat{x}_t)-C\left(x^*\right)\right)\mathbb{I}\left[\mathcal{B}_t^\complement\right]\right].        
\end{equation}
Intuitively, $\sum_{t=1}^T I_{1,t}$ is the total regret incurred when $\hat{x}_t$ enters $\mathcal{N}(x^*,\beta)\cap \mathcal{X}$ and we will prove it of order $\O(\log T/\alpha)$. Note the bound, which is not related to $\beta$, resembles the optimal regret of $\alpha$-strongly convex optimization problems as if $C(x)$ were globally $\alpha$-strongly convex.

The term $\sum_{t=1}^T I_{2,t}$ is the regret incurred when $\hat{x}_t$ is out of $\mathcal{N}(x^*,\beta)\cap \mathcal{X}$ and we will prove it of order $\O(1/\alpha\beta)$ by general convexity and bound for $g(\hat{x}_t)$. Note the bound would become larger as the width $\beta$ decreases. Intuitively, $\hat{x}_t$ may not be in $\mathcal{N}(x^*,\beta)\cap \mathcal{X}$ in the early stages and the stages would be longer as $\beta$ decreases, which contributes to the $\O(1/\alpha\beta)$ regret. The detailed proof is presented in Section \ref{subsec:proof-ub}.

\subsection{Application to Newsvendor Problems}\label{subsec:app-newsvendor}

Recall problem Eq.~\eqref{eq:sto-prog} covers a wide range of classical newsvendor problems, including the following:
\begin{itemize}
\item \textbf{Convex inventory cost:} $c(x,d)= h_f((x-d)^+)+b_f((d-x)^+)$.
\item \textbf{Linear inventory cost:} $c(x,d)= h\cdot(x-d)^++b\cdot(d-x)^+$.

\item  \textbf{Convex production cost:} $c(x,d) = c_p(x) - p\min\{x,d\}$.
\end{itemize}

For the above newsvendor problems, we make the following bounded assumption on demand.

\begin{assumption}\label{ass:demand-nvp}
Demand $D$ is bounded, i.e., $\bar D$ is a finite constant, where $\bar D:= \sup\{x: F_D(x) <1\}$. Moreover, the demand distribution $D$ has a differentiable CDF $F_D(x)$.
\end{assumption}

\textbf{Verification of Assumption \ref{ass:regular}.} By the convexity of $c(\cdot,d)$ and the above assumption on the distribution $D$, it is easy to see that Assumptions \ref{ass:regular-1} and \ref{ass:regular-3} hold. We explain why Assumption \ref{ass:regular-2} is satisfied. Since $D$ is bounded, the optimal solution $x^*$ to problem Eq.~\eqref{eq:sto-prog} satisfies $x^* \leq \bar{D}$. Therefore, if $[0,\bar{D}] \subset \mathcal{X}$, $x^*$ would be in the interior of $\mathcal{X}$. Similarly, the optimal solution of SAA problem can be obtained in the interior of $\mathcal{X}$. If $\hat{g}_t(x)$ is continuous, the optimality condition leads to $\hat{g}_t(\hat{x}_t)=0$. On the other hand, for problems with discontinuous $\hat{g}_t(x)$, for example, linear cost problem, $\hat g_t(x)$ is a stair function. Specifically,
\[
\hat{g}_t(x) = \frac 1t \sum_{i=1}^t c'(x,d_i) =  \frac 1t \sum_{i=1}^t h\I[x \geq d_i] -b\I[x < d_i].
\]
The optimal solution $\hat{x}_t$ is the $b/(h+b)$-quantile of the empirical distribution and $\hat{x}_t$ is nearly a zero point satisfying $\hat{g}_t(\hat{x}_t) \leq (h+b)/t$. Therefore, Assumption \ref{ass:regular-2} holds.

Note that the differentiability assumption on $F_D(x)$ is widely used for the analysis of the newsvendor problem with (local) strong convexity \citep{huh2009nonparametric,besbes2013implications,lin22datadriven}. For the boundedness assumption on $D$, we make the following discussion.

\begin{remark}[Boundedness Assumption] The boundedness assumption is quite mild and practical, since demand in real-world scenarios is always bounded. It is also remarkable that $\bar{D}$ is only used in verification of Assumption \ref{ass:regular} and it is not required in the implementation of the SAA method. Meanwhile, similar bounded assumptions on the optimization domain are widely adopted in the general SAA analysis. Our boundedness assumption plays a role similar to these assumptions.
There are also studies about linear-cost newsvendor problems \citep{levi2007nearoptimal,lin22datadriven} that do not require the boundedness of $D$. Their analysis is tailored for the linear inventory costs, utilizing the quantile solution structure, and does not work for the general cost function considered in this paper. In Sections \ref{subsec:linear_SC_to_ub}, we modify our general analysis techniques for linear-cost newsvendor problems without the boundedness assumption on demand. We also establish the optimal regret bound $\mathcal O\!\left(\frac{1}{\alpha}\log T+\frac{1}{\alpha\beta}\log\bigl((\alpha\beta)^{-1}\bigr)\right)$.
\end{remark}

\subsection{Proof of Theorem \ref{tm:tm_main_regret_ub}}\label{subsec:proof-ub}
This section presents the proof of Theorem \ref{tm:tm_main_regret_ub}, which proceeds in three steps.
\begin{enumerate}
    \item We first show, via a uniform gradient approximation argument, that the expected gradient evaluated at the SAA solution, i.e., $g(\hat{x}_t)$, is small with high probability.
    \item We then use the local strong convexity condition to convert this gradient control into a localization result: whenever $g(\hat{x}_t)$ is sufficiently small, $\hat{x}_t$ must lie in a neighborhood of the optimal solution $x^*$.
    \item Finally, we decompose the regret according to whether $\hat{x}_t$ lies inside or outside the strong convexity neighborhood. Within the neighborhood, the regret is controlled by $|g(\hat{x}_t)|^2$ through local strong convexity; otherwise, we bound the regret by the small probability of this bad event.
\end{enumerate}

We begin with the following lemma, which plays the central role in the regret analysis. This lemma shows that as $t$ increases, $g(\hat{x}_t)$ tends to concentrate on $0$ exponentially. We first bound the pseudo dimension of the function class $\mathcal{F}=\{c'(x,\cdot): x \in \mathcal{X}\}$ based on the convexity of $c(\cdot,d)$ (Assumption \ref{ass:regular-3}) and apply a classic uniform concentration result (see Appendix \ref{sec:omitted-ub} for its proof) to prove the lemma. 
\begin{lemma}\label{le:le_uniform_convergence_g}
There exist a universal constant $C_0>0$ such that for any $t$ and any $\lambda>0$, 
$$ \mathbb{P}\left[\left|g(\hat{x}_t)\right|\geq \lambda+\frac{4C_0B+C_1}{\sqrt t}\right]\leq \exp \left(-\frac{t\lambda^2}{2B^2}\right).
$$
\end{lemma}

Now we are ready for the detailed proof of Theorem \ref{tm:tm_main_regret_ub}.

\proof{Proof of Theorem \ref{tm:tm_main_regret_ub}.}

As discussed earlier, we decompose the regret into two parts as
\[
\mathcal{R}^{SAA}(T) = \sum_{t=1}^T I_{1,t} + \sum_{t=1}^T I_{2,t}.
\]

\noindent{\bf Upper bound of $\sum_{t=1}^T I_{1,t}$}. Since the optimal solution $x^*$ is obtained in the interior of $\mathcal{X}$, we know $g(x^*) = 0$.
Under the local strong convexity, $g'(x) \geq \alpha$ for $\mathcal{N}(x^*,\beta)\cap\mathcal{X}$ and $g(x^*)=0$, we know that $|g(x)| \geq \alpha\beta$, for $x$ in $\mathcal{X}$ but not in $\mathcal{N}(x^*,\beta)$. 
Therefore, event $\mathcal{B}_t$ that $g(\hat{x}_t) \leq \alpha\beta$ implies that $|\hat{x}_t-x^*| \leq \beta$.
Conditioned on event $\mathcal{B}_t$, by that $C(x)$ is $\alpha$-strongly convex on $\mathcal{N}(x^*,\beta)\cap\mathcal{X}$ and properties of strongly convex function, we obtain that 
\begin{align*}
I_{1,t}=\mathbb{E}\left[\left(C(\hat{x}_t)-C\left(x^*\right)\right)\mathbb{I}\left[\mathcal{B}_t\right]\right] 
\leq \mathbb{E}\left[\frac{1}{2\alpha}\left|g(\hat{x}_t)\right|^2\right].  
\end{align*}

By Lemma \ref{le:le_uniform_convergence_g} and some computation, it holds that 
\begin{align}
\mathbb{E}\left[\left|g(\hat{x}_t)\right|^2\right]=\int_{0}^{\infty} \mathbb{P}\left[\left|g(\hat{x}_t)\right|^2\geq y\right]\mathrm{d}y \leq \int_{0}^{r^2} 1\mathrm{d}y+\int_{r^2}^{\infty} \exp \left(-t\left(\sqrt{y}-r\right)^2/2B^2\right)\mathrm{d}y\leq \frac{K_4}{t},\label{eq:g-squre}
\end{align}
where we define $r = (4C_0B+C_1)/\sqrt t$ and $K_4 = (4C_0B+C_1)^2+2B^2+\sqrt{2\pi}B(4C_0B+C_1)$.

By the above two equations, we obtain that for $T \geq 2$,
\begin{align}
        \sum_{t=1}^TI_{1,t}\leq \sum_{t=1}^T \frac{K_4}{t\alpha} \leq 2K_4\cdot \frac{\ln T}{\alpha} .\label{eq:ub_I_1_total_result}
    \end{align}

\noindent{\bf Upper bound of $\sum_{t=1}^T I_{2,t}$.}
By the convexity of $C(x)$, we know
\begin{align*}
I_{2,t}&=\mathbb{E}\left[\left(C(\hat{x}_t)-C\left(x^*\right)\right)\mathbb{I}\left[\mathcal{B}_t^\complement\right]\right]\leq
\mathbb{E}\left[g(\hat{x}_t)|\hat{x}_t-x^*|\cdot \mathbb{I}\left[\mathcal{B}_t^\complement\right]\right]\leq \mathcal{D}_{\mathcal{X}}\left(\E[|g(\hat{x}_t)|^2]\mathbb{P}\left[\mathcal{B}_t^\complement\right]\right)^{1/2},
\end{align*}
where the second inequality is by the definition of $\mathcal{D}_{\mathcal{X}}$ (see the definition in Theorem~\ref{tm:tm_main_regret_ub}) and Cauchy-Schwartz inequality.
    
By Lemma \ref{le:le_uniform_convergence_g}, the probability of $\mathbb{P}[\mathcal{B}_t^\complement]$ can be bounded by 
\begin{align} \mathbb{P}[\mathcal{B}_t^\complement] \leq 
\begin{cases}
    1, &\text{ if } t \leq 4(4C_0B+C_1)^2/(\alpha\beta)^2,\\
    \exp \left(-\alpha^2\beta^2t/8\right), &\text{ if } t \geq 4(4C_0B+C_1)^2/(\alpha\beta)^2.
\end{cases}
\label{eq:ub_I2_lower_tail}
\end{align}    
    
Let $K_5=4(4C_0B+C_1)^2/(\alpha\beta)^2$. By Eq.~\eqref{eq:g-squre} and Eq.~\eqref{eq:ub_I2_lower_tail}, we obtain 
\begin{align}
\nonumber    \sum_{t=1}^T I_{2,t} &\leq \mathcal{D}_{\mathcal{X}}\sum_{t=1}^{\lceil K_5 \rceil}\frac{\sqrt{K_4}}{\sqrt{t}} +\mathcal{D}_{\mathcal{X}} \sum_{t=\lceil K_5 \rceil+1}^T \frac{\sqrt{K_4}}{\sqrt{t}} \exp \left(-\frac{\alpha^2\beta^2 t}{16}\right) \\
   &\leq 2\mathcal{D}_{\mathcal{X}}\sqrt{K_4\lceil K_5 \rceil} + \frac{16\mathcal{D}_{\mathcal{X}}\sqrt{K_4}}{\alpha^2\beta^2\sqrt{K_5}}\leq 2\mathcal{D}_{\mathcal{X}}\sqrt{K_4}+\frac{K_2}{\alpha\beta},\label{eq:I2-bound}
\end{align}
where $K_2 = 8\sqrt{2}\mathcal{D}_{\mathcal{X}}\sqrt{K_4}(4C_0 B +C_1)+ 8\mathcal{D}_{\mathcal{X}}\sqrt{K_4}/(4C_0B+C_1)$ as define before.

Let $K_1 = 2\bar D\sqrt{K_4}$ Combining Eq.~\eqref{eq:ub_I_1_total_result} and Eq.~\eqref{eq:I2-bound}, the total regret is upper bounded by \begin{align}
\nonumber\sum_{t=1}^T\mathbb{E}\left[C(\hat{x}_t)-C\left(x^*\right)\right]&=\sum_{t=1}^TI_{1,t}+\sum_{t=1}^TI_{2,t} \leq K_1 +K_2\cdot\frac{1}{\alpha\beta}+K_3\cdot\frac{\ln T}{\alpha},\nonumber
\end{align}
where we complete the proof.
\Halmos\endproof

\section{Lower Bound}\label{sec:lower bound}
In this section, we establish minimax regret lower bounds for the general optimization problem Eq.~\eqref{eq:sto-prog} under global convexity (Definition \ref{def:global}) and local strong convexity (Definition \ref{def:local}). As the general problem covers newsvendor problems as a special case and is more challenging, it suffices to establish minimax lower bounds for the newsvendor problem with linear inventory costs (i.e., $c(x,d)=h\cdot(x-d)^++ b\cdot(d-x)^+$).

As we discussed in Section~\ref{subsec:cvx-condition},  
for the linear-cost newsvendor problem, the $\alpha$-global strong convexity (resp. $(\alpha,\beta)$-local strong convexity) of the $C(x)$ is equivalent to the $\alpha/(h+b)$-global minimal separation condition (resp. $(\alpha/(h+b),\beta)$-local minimal separation condition) on demand distributions.
Thus, to establish a lower bound under $\alpha$-global strong convexity (or $(\alpha,\beta)$-local strong convexity) condition, it suffices to prove the minimax lower bound under the corresponding condition on demand distributions.

In the following theorem, we establish the minimax lower bound for the $\alpha$-global-minimal-separation  demand distribution class $\mathcal{F}_{\alpha}:=\{F_D(\cdot): F_D'(x) \geq \alpha \text{ for } x \in [0,\bar{D}] \}$. This lower bound result matches the existing upper bound result $\mathcal O(\log T/\alpha)$ of SAA and SGD method for newsvendor problems.

\begin{theorem}\label{thm:lbgloabal}
Let $\rho =b/(h+b)$.       For any $\alpha\leq \min\left\{1/2,2\rho,2\left(1-\rho\right)\right\}$, when $T\geq \max\{\alpha^{-3},64\}$, the minimax regret over $\mathcal{F}_{\alpha}$ satisfies\begin{align}
            \nonumber\underset{\pi\in\Pi}{\inf}\underset{F\in\mathcal{F}_{\alpha}}{\sup}R^{\pi}(F,T)\geq K_6\cdot \frac{\ln T}{\alpha}, 
        \end{align}
        where $K_6=\min\{\rho(1-\rho)/4\pi^2,1/(400\pi^2)\}(h+b)/12$.
\end{theorem}

Combining the above theorem with the $\Omega(1/(\alpha\beta))$ lower bound in \cite{lin22datadriven} (see their Theorem 1) for the $(\alpha,\beta)$-local-minimal-separation  demand distribution class
$\mathcal{F}_{\alpha,\beta}:=\{F_D(\cdot): F_D'(x) \geq \alpha \text{ for } x \in [x^*-\beta, x^*+\beta] \}$,  we could get the following lower bound result. This result together with our Theorem~\ref{tm:tm_main_regret_ub} demonstrates the regret rate optimality of SAA under the $(\alpha,\beta)$-local strong convexity condition.

\begin{corollary}\label{cor:lblocal}  Let $\rho:=b/(b+h)$.
        Suppose $\alpha\leq \min\left\{1/2,2\rho,2\left(1-\rho\right)\right\}$, 
        $\alpha\beta \leq \min \{\rho/3, (1-\rho)/2\}$ and $4\beta +1 \leq \E[D]$.
        When $T\geq \max\{\alpha^{-3},64\}$, the minimax regret over $\mathcal{F}_{\alpha,\beta}$ satisfies\begin{align}
            \nonumber\underset{\pi\in\Pi}{\inf}\underset{F\in\mathcal{F}_{\alpha,\beta}}{\sup}R^{\pi}(F,T)\geq K'_6 \cdot \frac{\ln T}{\alpha} + K_7 \cdot \frac{1}{\alpha\beta} , 
        \end{align}
         where $K'_6=\min\{\rho(1-\rho)/4\pi^2,1/(400\pi^2)\}\cdot(h+b)/24$ and $K_7=\exp (-12/ \rho-4 /(1-\rho))(b+h)/64$.
\end{corollary}

\subsection{Main Idea and Proof Sketch of Theorem~\ref{thm:lbgloabal}} 
In this section, we provide intuition for the lower bound construction and outline the main ideas of the proof. The detailed proof, including all technical derivations and information-theoretic arguments, is presented in Section~\ref{subsec:4.2}.

Inspired by the lower bound analysis in \cite{besbes2013implications}, we first reduce the problem of deriving a regret lower bound to that of establishing a lower bound on parameter estimation error, and the reduced step can be achieved by using the van Trees inequality (see Lemma \ref{le:vTI} in the Appendix~\ref{sec:useful-lemma}).

Specifically, we denote a family of parameterized demand distributions as $\{f(\cdot|\theta):\theta\in\Theta \text{ and } \theta\sim q(\theta)\}$, where $f(\cdot|\theta)$ is the demand PDF with  $\theta$, $\Theta$ is the parameter space and  $q(\theta)$ is the prior distribution PDF of $\theta$. We will also use $F_\theta$ to denote the CDF of $f(\cdot|\theta)$.
Let $C_\theta(x)$ be the expected cost function with respect to demand PDF $f(\cdot|\theta)$ and $x^*_\theta=\argmin_{x} C_\theta(x):=h(\theta)$. For any admissible policy $\pi\in\Pi$ and $t\in[T]$, by Taylor expansion and the van Trees inequality we have 
\begin{align}\label{eq:lb-eq-1}
    \mathbb{E}\left[C_\theta\left(x_t^\pi\right)-C_\theta\left(x^*_\theta\right)\right]\geq \left(\underset{x,\theta}{\min} 
 C^{''}_\theta(x)/2\right)\mathbb{E}\left[\left(x_t^\pi-x_\theta^*\right)^2\right] \geq\frac{\left(\underset{x,\theta}{\min} 
 C^{''}_\theta(x)/2\right)\mathbb{E}[h'(\theta)^2]}{\E[I_t(\theta)]+I(q)},
\end{align}
where $I_t(\theta)$ is the expected Fisher information for $\theta$ at time $t$, $I(q)$ is the Fisher information of $q(\theta)$ and the expectation is taken over the joint distribution of the historical demand and $\theta$.

As in many existing lower bound proofs, the most critical part of the proof is the design of the \textit{hard instances},
which are a family of parameterized demand distributions in our problem.
 To establish a lower bound result with respect to both parameter $\alpha$ and $T$ under the $\alpha$-global  minimal separation condition, the hard instances in our problem should satisfy the following basic assumptions:
\begin{enumerate}
    \item[\textbf{A1.}] For any $\theta\in\Theta$, $f(\cdot|\theta)$ should be bounded from below by $\alpha$ within $[0,x_\theta^*]$.\label{ass:A1} 
        \item[\textbf{A2.}] For any $\theta\in\Theta$,   $\sup\{x: F_D(x) <1\}$ should be bounded above by an absolute constant independent of parameter $\alpha$. \label{ass:A2} 
    \item[\textbf{A3.}] For any $x$ in the domain set, the likelihood function $f(x|\cdot)$ should be absolutely continuous with respect to $\theta$.\label{ass:A3} 
        \item[\textbf{A4.}] For random variable $X\sim f(\cdot|\theta)$, we have $\E_\theta\left[\frac{\partial}{\partial\theta}\left(\log f\left(X|\theta\right)\right)\right]=0$,  where $\E_\theta$ denotes the expectation over $X\sim f(\cdot|\theta)$.\label{ass:A4}     
\end{enumerate}

By the $\alpha$-global  minimal separation condition, we have $\min_{x,\theta}C^{''}_\theta(x)=\Omega(\alpha)$. Thus, to attain the desired lower bound $\mathcal O(\log T/\alpha)$, in addition to the above basic requirements, the hard instances should be carefully designed to satisfy $\mathbb{E}[h'(\theta)^2]/(I_t(\theta)+I(q))=\Omega(1/(\alpha^2t))$.

This task involves two primary objectives: 1) ensuring a relatively significant minimum of the variability of the optimal solution (i.e., $\mathbb{E}[h'(\theta)^2]$);  2)
controlling the Fisher information (i.e., $I_t(\theta)$) from being too large. 
The main difficulty lies in reconciling the seemingly conflicting requirements of these two objectives: the first objective requires that the optimal quantities $x_\theta^*$ be sensitive to changes in $\theta$, while, the second objective requires that the distribution $f(\cdot|\theta)$ exhibit a more stable response to parameter shifts.

In this paper, we construct the following novel and carefully designed class of hard instance, which not only meets the basic requirements but also successfully attains the desired lower bound. 

\begin{figure}[t!]
    \centering
    \begin{minipage}{0.49\textwidth}
        \centering
        \includegraphics[width=\linewidth]{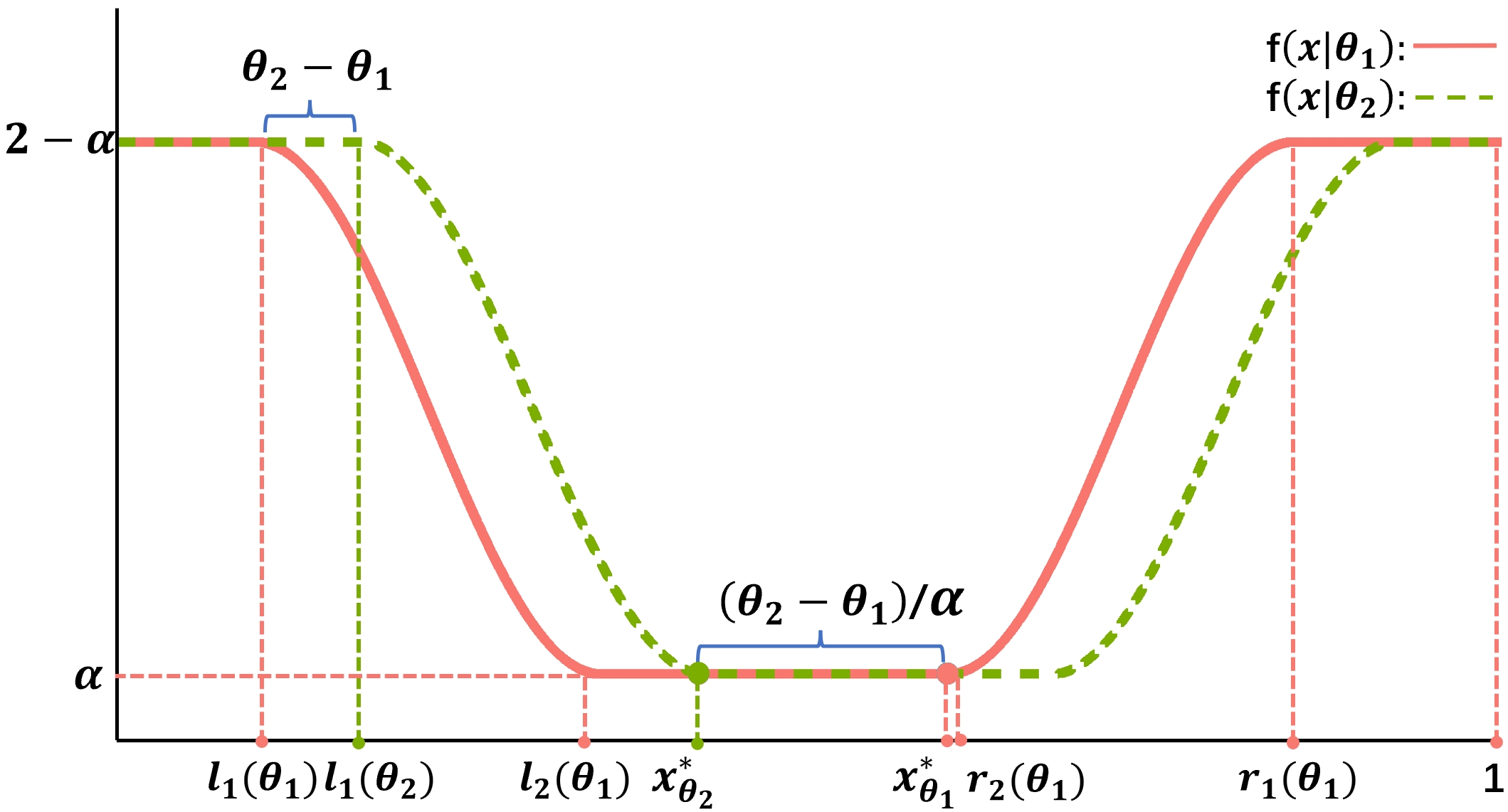}
        \caption{Our inverted-hat-shape  hard instance using \texttt{cosin} functions to connect segments.}
        \label{fig:hard-left}
    \end{minipage}\hfill
    \begin{minipage}{0.49\textwidth}
        \centering
        \includegraphics[width=\linewidth]{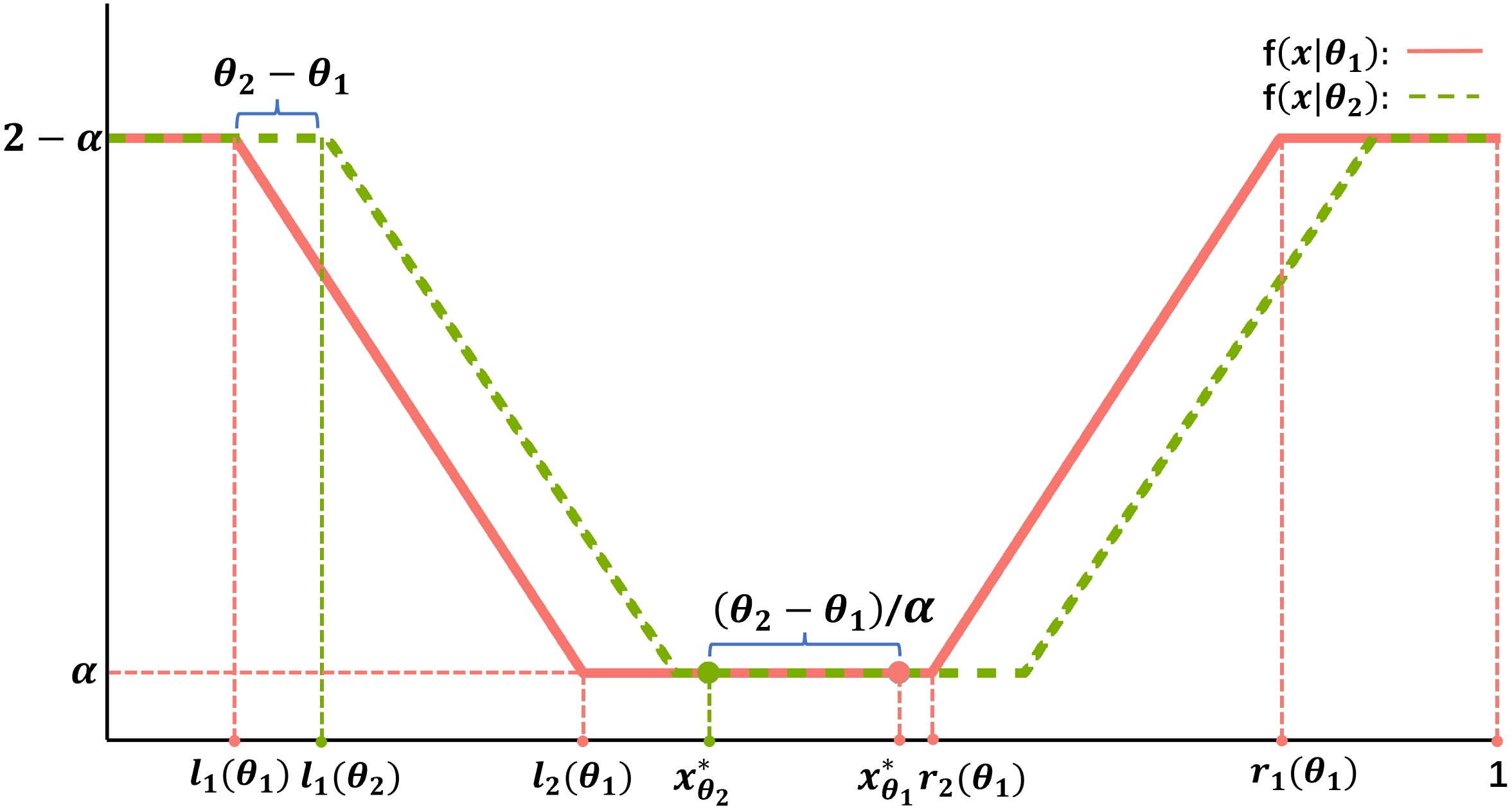}
        \caption{Another choice of hard instance using straight lines to connect segments.}
        \label{fig:hard-right}
    \end{minipage}
\end{figure}

The parameter space is defined as $\Theta=\left[-\alpha/20,\alpha/20\right]$. For $\theta\in\Theta$,$$
        f\left(x|\theta\right)=\left\{\begin{aligned}
		&2-\alpha,&x\in\left[0,l_1(\theta)\right]\cup\left[r_1(\theta),1\right],\\
            &\alpha,&x\in\left(l_2(\theta),r_2(\theta)\right],\\
                &\alpha+(1-\alpha)\left(\cos \left(w_1\left(x-l_1(\theta)\right)\right)+1\right),&x\in\left[l_1(\theta),l_2(\theta)\right],\\
                &\alpha+(1-\alpha)\left(\cos \left(w_2\left(r_1(\theta)-x\right)+1\right)\right),&x\in\left[r_2(\theta),r_1(\theta)\right],
                \end{aligned}
            \right.
    $$
    where $l_1(\theta)=(4\rho-\alpha)/(16-8\alpha)+\theta$, $l_2(\theta)=l_1(\theta)+\rho/2$, $r_2(\theta)=l_2(\theta)+1/4$, $r_1(\theta)=r_2(\theta)+(1-\rho)/2$, $w_1=2\pi/\rho$, and $w_2=2\pi/\left(1-\rho\right)$. 
The PDF of the prior distribution is defined as $$
        q(\theta)=\frac{20}{\alpha}\cos^2\left(\frac{10\pi}{\alpha}\theta\right)\mathbb{I}\left[\theta\in\left[-\frac{\alpha}{20},\frac{\alpha}{20}\right]\right].
    $$
The graph illustration of our hard instance is presented in Figure~\ref{fig:hard-left}.

\subsection{Proof of Theorem~\ref{thm:lbgloabal}}\label{subsec:4.2}
We begin with the following lemma, which shows that our hard instance is well-defined and satisfies the basic assumptions.
\begin{lemma}\label{lem:basic properties} 
Suppose $\alpha\leq \min\left\{1/2,2\rho,2\left(1-\rho\right)\right\}$, 
 $f(\cdot|\theta)$ is a well-defined PDF for any $\theta\in\Theta$, and our hard instance satisfies Assumptions~\hyperref[ass:A1]{A.1} -- \hyperref[ass:A4]{A.4}.
\end{lemma}

Through the careful design of the inverted-hat-shape structure of the PDF plot in Figure~\ref{fig:hard-left}, we can successfully constrain $x_\theta^*$ within the central part of the domain and thus make $x_\theta^*$ sensitive to parameter $\theta$.
 More specifically, as illustrated by Figure~\ref{fig:hard-left} and shown by the following lemma, our design guarantees that $x_\theta^*$ varies $\Omega(1)$ when $\theta$ varies only $\Theta(\alpha)$, which results in the relatively significant variability of $x_\theta^*$.  
\begin{lemma}\label{le:position of optimal solution}
For any $\theta\in\Theta:=[-\alpha/20, \alpha/20]$, we have $x_\theta^*$ lies within interval $[l_2(\theta),r_2(\theta)]$ and
\begin{align}\label{eq:expression_for_xthetastar}
    x_\theta^*:=h(\theta) = \frac{\rho}{2}+\frac{4\rho-\alpha}{16-8\alpha}+\frac{1}{8}-\left(\frac{2}{\alpha}-2\right)\theta.
\end{align}
\end{lemma}
The following lemma demonstrates that the Fisher information of $f(\cdot|\theta)$ is bounded above by $\mathcal O (t)$.
\begin{lemma}\label{le:fisherinformation}
   For any $t$ and $\theta\in\Theta:=[-\alpha/20, \alpha/20]$, the Fisher information 
   \begin{align}
       I_t(\theta):= \E_{\theta}\left[\left(\frac{\partial\log f\left(X_1,\dots,X_{t}|\theta\right)}{\partial\theta}\right)^2\right] \leq \frac{4\pi^2t}{\rho(1-\rho)},
   \end{align}
  where $f\left(X_1,\dots,X_{t}|\theta\right)$ is the joint PDF of independent random variables $X_1,\dots, X_t\sim f(\cdot|\theta)$, and $\E_\theta$ denotes the expectation over $X_1,\dots, X_t$.
\end{lemma}

It is worth noting that in construction, we utilize the \texttt{cosine} functions to connect separated segments rather than employing simpler straight lines (see Figure~\ref{fig:hard-right}). This approach is adopted because the Fisher information for the latter design is $\Theta(t\log1/\alpha)$, as opposed to the desired $\mathcal{O}(t)$ in Lemma~\ref{le:fisherinformation}.

With all the above technical lemmas in hand, we are ready to complete the proof of Theorem~\ref{thm:lbgloabal}.

\proof{Proof of Theorem~\ref{thm:lbgloabal}.}
Since $\Theta=\left[-\alpha/20,\alpha/20\right]$, we have 
    \begin{align}\label{eq:integral_expression_lb}
        \underset{\pi\in\Pi}{\inf}\underset{F\in\mathcal{F}_{\alpha}}{\sup}\{R^{\pi}\left(F,T\right)\}\geq \underset{\pi\in\Pi}{\inf}\int_{-\alpha/20}^{\alpha/20}\mathcal{R}^{\pi}(F_\theta,T)q(\theta)\mathrm{d}\theta.
    \end{align}
    For an arbitrary policy $\pi$, by Eq.~\eqref{eq:lb-eq-1} the regret under the demand distribution $F_\theta$ is  
    \begin{align}\label{eq:regret_one_epoch_def} \int_{-\alpha/20}^{\alpha/20}\mathcal{R}^{\pi}(F_\theta,T)q(\theta)\mathrm{d}\theta\geq\sum_{t=1}^T \mathbb{E}\left[C_\theta\left(x_t^\pi\right)-C_\theta\left(x^*_\theta\right)\right] \geq\sum_{t=1}^T \frac{\left(\underset{x,\theta}{\min} 
 C^{''}_\theta(x)/2\right)\mathbb{E}[h'(\theta)^2]}{\E[I_t(\theta)]+I(q)},
    \end{align}
    Due to $f(x|\theta)\geq \alpha$, we have ${\min_{x,\theta}} 
 C^{''}_\theta(x)\geq (h+b)\alpha$. By Lemma~\ref{le:position of optimal solution}, we have $\mathbb{E}[h'(\theta)^2]\geq (2/\alpha -2)^2$. Lemma~\ref{le:fisherinformation} shows that 
$\E[I_t(\theta)]\leq 4\pi^2t/(\rho(1-\rho)).$  Besides,  the Fisher information of $q(\theta)$ is \begin{align}\label{eq:Fisher_prior}
        I(q)=\int_{-\alpha/20}^{\alpha/20} \frac{20^3\pi^2}{\alpha^3}\frac{\cos^2(10\pi\theta/\alpha)\sin^2(10\pi\theta/\alpha)}{\cos^2(10\pi\theta/\alpha)}\mathrm{d}\theta=\frac{400\pi^2}{\alpha^2}.
    \end{align}
Therefore, 
\begin{align}
    \nonumber\underset{F\in\mathcal{F}_{\alpha}}{\sup}\{R^{\pi}\left(F,T\right)\}&\geq \frac{(h+b)\alpha}{2} \sum_{t=1}^T\frac{(2/\alpha-2)^2}{ 4\pi^2t/(\rho(1-\rho))+400\pi^2/\alpha^2}
    \\&\geq\frac{h+b}{2\alpha} \sum_{t=1}^T\frac{\tilde K_6}{t+\alpha^{-2}}\geq \frac{h+b}{2\alpha}\int_1^{T+1} \frac{\tilde K_6}{t +\alpha^{-2}}\mathrm{d}t\nonumber  \\ 
       &\geq\frac{h+b}{2} \tilde K_6\alpha^{-1}\ln \left(\frac{T+1+\alpha^{-2}}{1+\alpha^{-2}} \right)\geq \frac{h+b}{2}\tilde K_6\alpha^{-1}(\ln T-2\ln \alpha^{-1}-\ln 2),\label{eq:lb_preparation_segmental_result}
\\&\geq  C_0\alpha^{-1}\ln T,\nonumber
    \end{align}
    where  $\tilde K_6=\min\{\rho(1-\rho)/4\pi^2,1/(400\pi^2)\}$, $K_6=\tilde K_6(h+b)/12$, the second inequality is due to $\alpha\leq 1/2$ and the last inequality is due to $T\geq \max\{\alpha^{-3},64\}$.     \Halmos
\endproof

\section{Extension and Discussion}\label{sec:extension}
The simple but generic idea of gradient approximation for analyzing the SAA method enables us to generalize our main results.
In this section, we will extend our method to another general function class motivated by newsvendor problems and discuss the relaxation of some technical assumptions.
\subsection{Extension to $(\alpha,\beta,\zeta)$-clustered Demand Distribution Class}\label{sec:clustered}
 \cite{chen2024survey} considered the linear-cost newsvendor problem and studied a $(\alpha,\beta,\zeta)$-clustered demand distribution class, where the demand CDF $F(\cdot)$ function satisfies
 \begin{align}\label{eq:zeta}
         \alpha|x-x^*|\leq |F(x)-F(x^*)|^{\frac{1}{\zeta+1}},\quad \forall x\in [x^*-\beta,x^*+\beta].
 \end{align}
Note that when $\zeta=0$, the condition reduces to the $(\alpha,\beta)$-local-minimal separation condition on the demand distributions. Therefore, the parameter $\zeta$ bridges convexity and strong convexity.

 For the linear-cost newsvendor problem, the cost function is $C(x) = \E[h\cdot (x-D)^+ + b\cdot (D-x)^+]$. Some calculation gives $C'(x) = (h+b)F(x)-b$ and   $F(x^*) = b/(b+h)$, we can obtain an equivalent formulation of the above expression  Eq.~\eqref{eq:zeta} as follows
 \begin{equation}
     \label{eq:generalized-zeta}
    \alpha|x-x^*|\leq \left|\frac{C'(x)}{h+b}\right|^{\frac{1}{\zeta+1}}.
 \end{equation}

The above inequality describes a function class beyond newsvendor problems. Therefore, inspired by the $(\alpha,\beta,\zeta)$-clustered demand distribution in \cite{chen2024survey} and the above observation, we generalize this condition and propose the following definition for the more general optimization problems.

\begin{definition}[Local gradient growth]\label{def:zeta_local}
   We say $C(x)$ satisfies the $(\alpha,\beta,\zeta)$-local gradient growth condition, if there exist constants $\alpha,\beta>0$, and $\zeta \geq 0$, such that
   \[
   \left\Vert\nabla C(x)\right\Vert_2 \geq \alpha^{\zeta+1}\Vert x-x^*\Vert_2^{\zeta+1},~\forall x\in\mathcal{N}(x^*,\beta) \cap \mathcal{X}.
   \]
\end{definition}

When $\zeta=0$, the above definition becomes $\left\Vert\nabla C(x)\right\Vert_2 \geq \alpha\Vert x-x^*\Vert_2^{1}$, which plays the key role in the regret analysis and can also be derived from Definition~\ref{def:local}. Thus, Definition~\ref{def:zeta_local} can be regarded as a generalization of Definition~\ref{def:local}. In the following, we demonstrate that our analysis technique remains applicable to general one-dimensional problems that satisfy the local gradient growth condition, which further implies the conclusion in the linear inventory cost setting and demand distributions satisfying Eq.~\eqref{eq:zeta}.

Define event that $\mathcal{E}_t=\{x^*-\beta\leq\hat{x}_t\leq x^*+\beta\}$. By Eq.\eqref{eq:g-squre} and Lemma \ref{le:le_uniform_convergence_g}, we can still prove the conclusion that
    $\E[|g(\hat{x}_t)|^2]\leq {K_4}/{t}.$
Thus, it holds that
\begin{align}\label{eq:ub_willma_central}
    \nonumber\E\left[(C(\hat{x}_t)-C(x^*))\cdot\I[\mathcal{E}_t]\right]&\leq \E\left[|g(\hat{x}_t)|\cdot|\hat{x}_t-x^*|\cdot\I[\mathcal{E}_t]\right]\\
    &\leq \alpha^{-1}\E[|g(\hat{x}_t)|^{\frac{\zeta+2}{\zeta+1}}\cdot\I[\mathcal{E}_t]] \nonumber\\
    &\leq \alpha^{-1}\left(\E\left[|g(\hat{x}_t)|^2\right]\right)^{\frac{\zeta+2}{2\zeta+2}} \nonumber\\&\leq \mathcal{O}(\alpha^{-1}t^{-\frac{\zeta+2}{2\zeta+2}}),
\end{align}
where the first inequality is due to the convexity of $C(x)$, the second inequality is due to the local gradient growth condition, and the third inequality is due to the Holder Inequality. 

Recall that $\mathcal{E}_t\supseteq\mathcal{B}_t=\{|g(\hat{x}_t)|\leq\alpha\beta\}$. 
By the similar analysis to the proof of Theorem \ref{tm:tm_main_regret_ub} (Eq.~\eqref{eq:I2-bound}), when  $t$ is sufficiently large ($t \geq 4(4C_0B+C_1)^2/(\alpha\beta)^2$) we could obtain
\begin{align}\label{eq:ub_willma_tail}
  \E\left[(C(\hat{x}_t)-C(x^*))\cdot\I[\mathcal{E}_t^{\complement}]\right]\leq   \E[(C(\hat{x}_t)-C(x^*))\cdot\I[\mathcal{B}_t^{\complement}]]\leq \mathcal{O}(t^{-1/2}\exp(-\alpha^2\beta^2t/16)).
\end{align}

Combining Eq.~\eqref{eq:ub_willma_central} and Eq.~\eqref{eq:ub_willma_tail}, when $t \geq 4(4C_0B+C_1)^2/(\alpha\beta)^2$, we conclude that \[
    \E\left[(C(\hat{x}_t)-C(x^*))\right]\leq \mathcal{O}(\alpha^{-1}t^{-\frac{\zeta+2}{2\zeta+2}}),
\]
which matches the optimal regret in \cite{chen2024survey} in terms of $t$ and parameters $\alpha,\zeta$.

Note that it is possible to establish matching upper and lower bounds in $T$ and all parameters $\alpha,\beta,\zeta$ based on our analysis techniques and hard instance construction. The focus of this paper is to understand the performance of SAA under $(\alpha,\beta)$-local strong convexity, and we leave it as a potential direction.

\subsection{Sample Complexity and Censored Demand Data}
\label{subsec:sample-compexity}
Our analysis can also be adopted to derive results for sample complexity and censored demand data.

\textbf{Sample complexity result.} Since our analysis yields high probability uniform concentration of gradient functions, a byproduct of our regret analysis is the sample complexity result under the local strongly convexity condition (Definition \ref{def:local}). Our results consist of two parts depending on the relationship between the desired accuracy $\epsilon$ and $\alpha\beta^2/2$. 
Noting that $C(x^*\pm\beta) - C(x^*) = \Omega(\alpha\beta^2/2)$, in the case of $\epsilon < \alpha\beta^2/2$,  $C(\hat{x}_t) - C(x^*)\leq \epsilon$ implies that $\hat{x}_t \in [x^* - \beta, x^* + \beta]$. Thus,
we can utilize the local strong convexity of $C(x)$ and our gradient approximation result to obtain a sample complexity of $\mathcal{O}(\epsilon^{-1}\log \delta^{-1})$, which is analogous to the sample complexity in \cite{levi2015data}.
For the case where $\epsilon > \alpha\beta^2/2$, we can combine our gradient approximation result with the global convexity of $C(x)$ to achieve a sample complexity of $\mathcal{O}(\epsilon^{-2}\log \delta^{-1})$, which is consistent with the sample complexity in \cite{levi2007nearoptimal}. Please refer to Section~\ref{subsec:SC_finite_supp} for the detailed theorem and proof.

\textbf{Censored demand data.} We next discuss how the proposed SAA framework can be applied to censored demand observations. Suppose that the historical data are given by $\mathcal{D}:=\{(\hat d_i,y_i)\}_{i=1}^n$, where $y_i$ is the on-hand inventory level before demand is realized and $\hat d_i=\min\{D_i,y_i\}$ is the observed censored demand. One can retain only those observations with $y_i\geq \bar{D}$ and discard the remaining censored observations. Moreover, because we only consider the data collected under $y_i \geq \bar{D}$, the retained observations have the same distribution as uncensored \textit{i.i.d.} demand samples. Therefore, applying SAA to this filtered dataset is statistically equivalent to applying SAA to a fully observed sample, with the effective sample size
\[
N_{\mathcal{D}}:=\sum_{i=1}^n \I[y_i \geq \bar{D}].
\]

Typically, under certain data-collecting policies \citep{fan2022sample}, we know $N_{\mathcal{D}} \approx c n$ for some $c>0$.
The theoretical guarantees developed for the uncensored setting extend directly to the censored setting under this observability condition. This condition can be relaxed for the classical linear-cost newsvendor problem; the optimal solution is characterized by a single critical quantile, so it is sufficient to observe demand up to a level that dominates the newsvendor solution rather than the entire support. 
Moreover, with a more refined, level-wise SAA-type estimator, one can attain the optimal sample complexity for the censored linear-cost newsvendor problem from purely offline censored data. For instance, \cite{fan2022sample} proposed an offline SAA-based algorithm that sweeps the historical inventory levels from low to high, building a level-dependent empirical CDF on the usable subsample at each level and stopping once the critical-quantile threshold is reached. 
 Whether our techniques, in particular the local-strong-convexity arguments, could be combined with such level-wise screening procedures to extend the guarantees to the local minimal separation setting is an interesting direction. However, since our focus is on establishing the optimality of the simplest, vanilla SAA, the precise derivations of these extensions are beyond the scope of this paper, we leave this problem for future research.

Observability conditions of this type are standard in the censored-demand inventory literature; for instance, \cite{fan2022sample} mentioned above; \cite{qin2023sailing} showed that, in multi-period inventory control with inventory carryover, optimal policy learning requires the observation boundary to exceed the upper support of demand. This requirement is particularly natural in our general convex-cost setting: unlike the linear-cost newsvendor problem, whose optimizer is pinned down by a single critical quantile, a general convex cost depends on the entire expected cost function, so distributional perturbations across the support can affect the optimal solution. Hence, insufficient high-demand observations may induce a global identification difficulty rather than only a local quantile-identification issue.
\begin{remark}[Censoring and information loss]
We focus on a general convex-cost newsvendor model with censored demand observations and analyze the performance of an SAA-based policy under suitable observability and coverage conditions. Our results show that, when the relevant region of the demand distribution is sufficiently observed, the SAA solution can learn the optimal decision effectively. A recent paper by \cite{hssaine2024data} provides a different perspective on censored-data newsvendor problems. In the linear-cost setting, they show that censoring may lead to fundamental information loss, so that vanishing regret is not always achievable without additional structural information. This observation highlights an important subtlety: 
censoring can affect what is statistically learnable from the data. Their sharp characterization relies on the critical-quantile structure of the linear-cost newsvendor problem.In contrast, our paper considers a more general class of convex cost functions, for which the optimality condition is not reducible to a single critical quantile. As a result, the mechanisms through which censoring affects learnability are quite different and are not directly covered by the arguments developed for the linear-cost case. A precise characterization for censored newsvendor problems with general convex costs remains an interesting direction for future research.
\end{remark}

\subsection{Relaxation of Boundedness Assumption for Newsvendor Problems}\label{sec:relaxation1}
When applying our general SAA results (Theorem \ref{tm:tm_main_regret_ub}) to the newsvendor problem with linear cost, we impose a technical assumption (see Assumption \ref{ass:demand-nvp}) that demand is bounded by some constant $\bar{D}$ for the verification of Assumption \ref{ass:regular}. The boundedness assumption ensures that SAA solutions are always an interior point of $\mathcal{X}$, which further provides guarantees for the gradients at these solutions. It is remarkable that we can relax the boundedness assumption on $D$ and only assume that the domain $\mathcal{X}$ contains the optimal solution $x^*$ and solve SAA problems within this set. Note by convexity, $C(x)$ is increasing outside $\mathcal{X}$ and the SAA solution $\hat{x}_t$ restricted to $\mathcal{X}$ is better than the unconstrained SAA solution $\hat{x}_t^u$. Therefore, we still have $f(\hat{x}_t) \leq f(\hat{x}_t^u)$ and can bound $f(\hat{x}_t^u)$ similar the to the analysis of Theorem \ref{tm:tm_main_regret_ub}.

Though the boundedness assumption is both mild and practically reasonable, since demand in real-world scenarios is always
bounded, the existing literature (such as \cite{levi2015data} and \cite{lin22datadriven}) on linear-cost newsvendor problems fully leverages the quantile-based structure of the optimal solution for their analysis, and demand is not required to be bounded. To keep aligned with these studies, we also study the linear-cost newsvendor problem without bounded assumption on $\mathcal{X}$ and $D$, and
establish the optimal regret (see Theorem~\ref{tm:tm_linearcost_regret_ub} in Section~\ref{subsec:linear_SC_to_ub} in the appendix).

\section{Numerical Results}\label{sec:numerical}
In this section, we conduct numerical experiments to test the performance of the SAA method under different problem settings and validate the theoretical regret dependence on parameters $\alpha$ and $T$ under $\alpha$-global strong convexity. The numerical study consists of two parts: Section~\ref{subsec:validation} validates the theoretical regret rate of the SAA method, while Section~\ref{subsec:comparison} compares the SAA method with other classical algorithms from the literature.
We use $U(a,b)$ to denote the continuous uniform distribution on the interval $[a,b]$.

\subsection{Validation of the Theoretical Regret Rate}\label{subsec:validation}

In this subsection, we conduct numerical experiments under both the linear-cost and the general-cost newsvendor problems to validate the $\mathcal{O}(\alpha^{-1}\log T)$ regret growth rate of the SAA method (in terms of $\alpha$ and $\log T$) established in Theorem~\ref{tm:tm_main_regret_ub} and Theorem~\ref{tm:tm_linearcost_regret_ub}.

\noindent{\bf Linear-cost newsvendor problems.} In this part, we take the cost function to be $c(x,d)=h\cdot(x-d)^++b\cdot(d-x)^+$. The demand distribution is a uniform distribution with parameters $0$ and $1/\alpha$. Specifically, demand $D$ has CDF $F_D(x)=\alpha x$ with $x\in[0,1/\alpha]$. We select the uniform demand distribution because it ensures the expected cost $C(x) = \mathbb{E}[c(x,D)]$ satisfies the $\alpha$-global strong convexity.

\begin{figure}[ht]
    \begin{center}
        \includegraphics[width =0.48\textwidth]{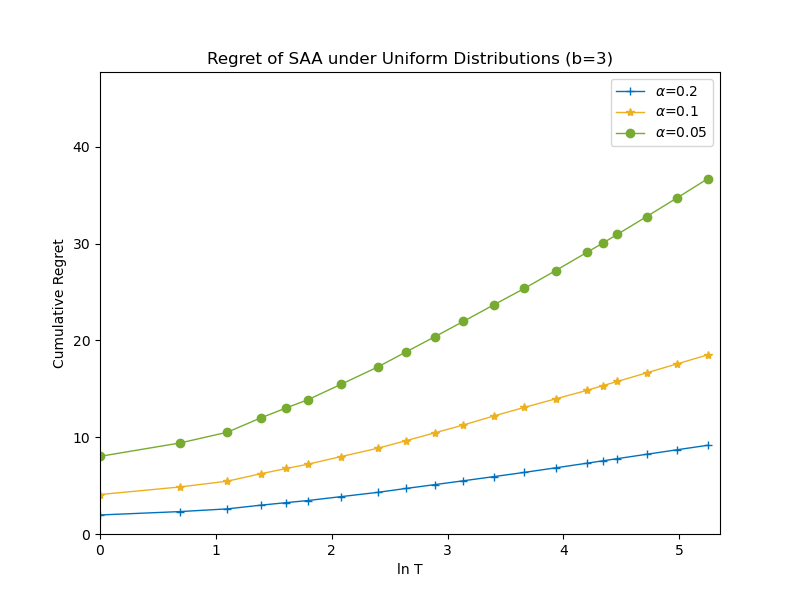}
        \includegraphics[width =0.48\textwidth]{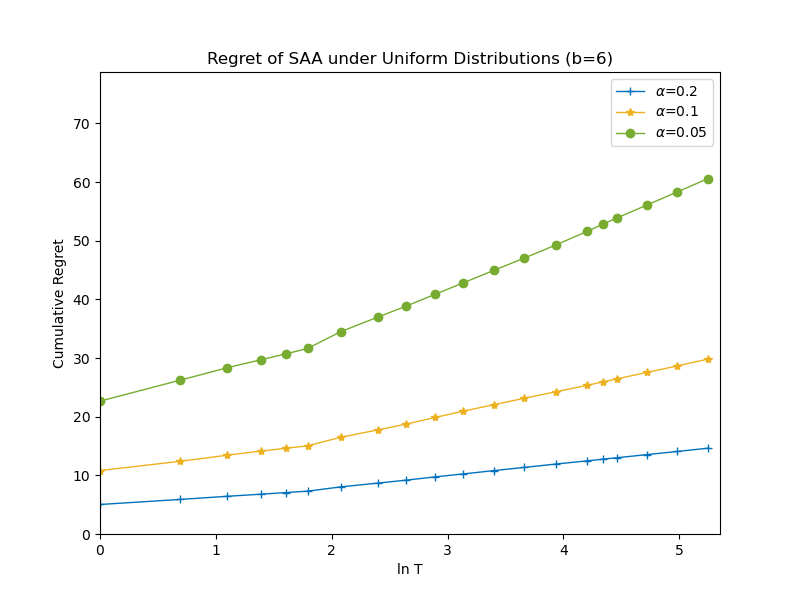}
        \includegraphics[width =0.48\textwidth]{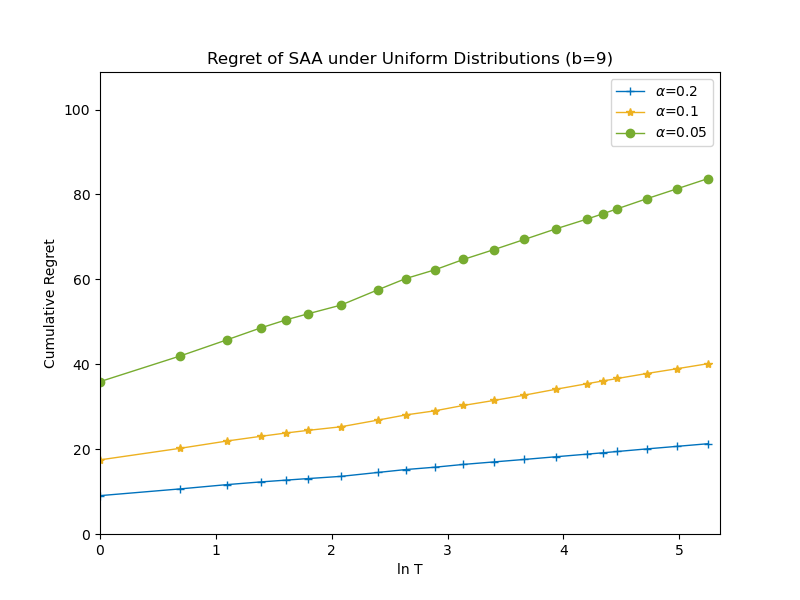}
        \includegraphics[width =0.48\textwidth]{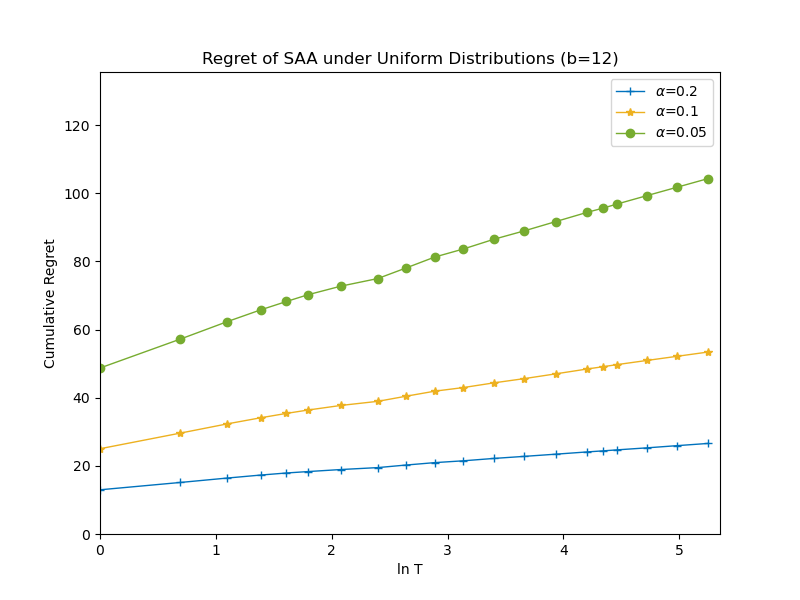}
    \end{center}
    \caption{Regret of SAA under Uniform Distributions and Linear Inventory Costs}\label{fig:NR_uniform_SAA_alpha}
\end{figure}

In Figure~\ref{fig:NR_uniform_SAA_alpha}, we report the cumulative regret of the SAA method for $\alpha \in\{0.2,0.1,0.05\}$, and $b \in \{3,6,9,12 \}$ with \textbf{$T\in\mathbb{N} \cap[1,200]$}. For different strong convexity parameters $\alpha$, we plot the horizontal axis in the logarithmic scale ($\ln T$) to show the logarithmic growth rate of the regret with respect to $T$.

As we can see from Figure~\ref{fig:NR_uniform_SAA_alpha}, the cumulative regret grows linearly with $\ln T$. In Table~\ref{tab2}, we further report the slope of each curve in Figure~\ref{fig:NR_uniform_SAA_alpha}, and we observe that $\alpha$ times the slope of the corresponding curves are approximately equal. Therefore, Figure~\ref{fig:NR_uniform_SAA_alpha} and Table~\ref{tab2} together validate the $\mathcal{O}(\alpha^{-1}\log T)$ regret  growth rate of SAA method established in Theorem~\ref{tm:tm_main_regret_ub} and Theorem~\ref{tm:tm_linearcost_regret_ub}.

\begin{table}[h!]
\caption{Slope of Cumulative Regret vs. $\ln T$ under Different $\alpha$ and $b$.}\label{tab2}
\renewcommand{\arraystretch}{1.5}
\begin{center}
\scalebox{1}{\begin{tabular}{l|l|l|l||l|l|l}
\hline
       & \makecell[l]{$\alpha_1=0.2$} & \makecell[l]{$\alpha_2=0.1$} & \makecell[l]{$\alpha_3=0.05$} & \makecell[l]{$\alpha_1\times \text{slope}$} & \makecell[l]{$\alpha_2\times \text{slope}$} & \makecell[l]{$\alpha_3\times \text{slope}$} \\ \hline
$b=3$  & 1.778                        & 3.524                        & 7.299                         & 0.356                                       & 0.352                                       & 0.365                                       \\ \hline
$b=6$  & 2.070                        & 4.286                        & 8.626                         & 0.414                                       & 0.429                                       & 0.431                                       \\ \hline
$b=9$  & 2.354                        & 4.515                        & 9.135                         & 0.471                                       & 0.452                                       & 0.457                                       \\ \hline
$b=12$ & 2.388                        & 4.773                        & 9.499                         & 0.478                                       & 0.477                                       & 0.475                                       \\ \hline
\end{tabular}

}
\end{center}
\end{table}

\noindent{\bf General-cost newsvendor problems.} In this part, we consider the general-cost newsvendor problem and focus on the common quadratic inventory cost and the quadratic production cost, where the cost function is $c(x,d)=h\cdot((x-d)^+)^2+b\cdot((d-x)^+)^2$ and $c(x,d)=x^2-p\min\{x,d\}$, respectively. The demand distribution has CDF $F_D(x)=x/5$ with $x\in[0,5]$, i.e., uniform distribution with parameters $0$ and $5$.

\begin{figure}[h!]
    \centering
    \includegraphics[width=0.48\textwidth]{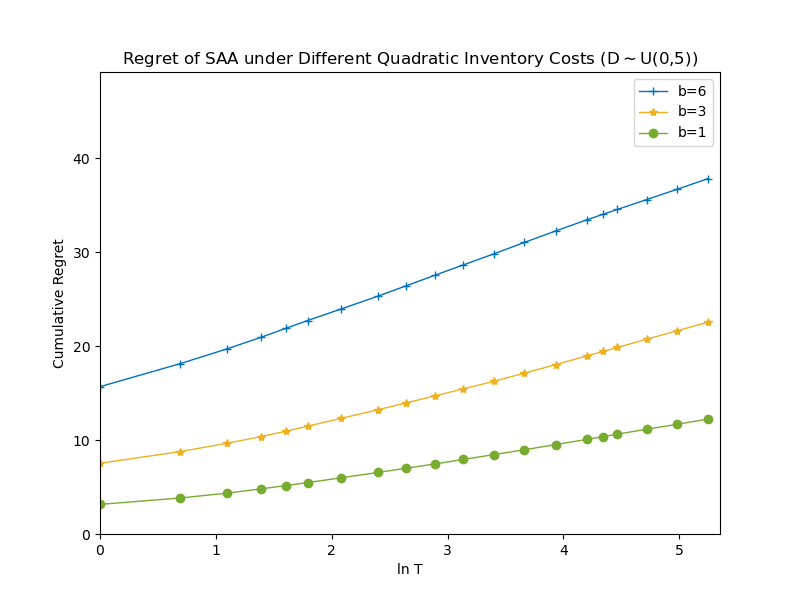}
    \hspace{0cm}
    \includegraphics[width=0.48\textwidth]{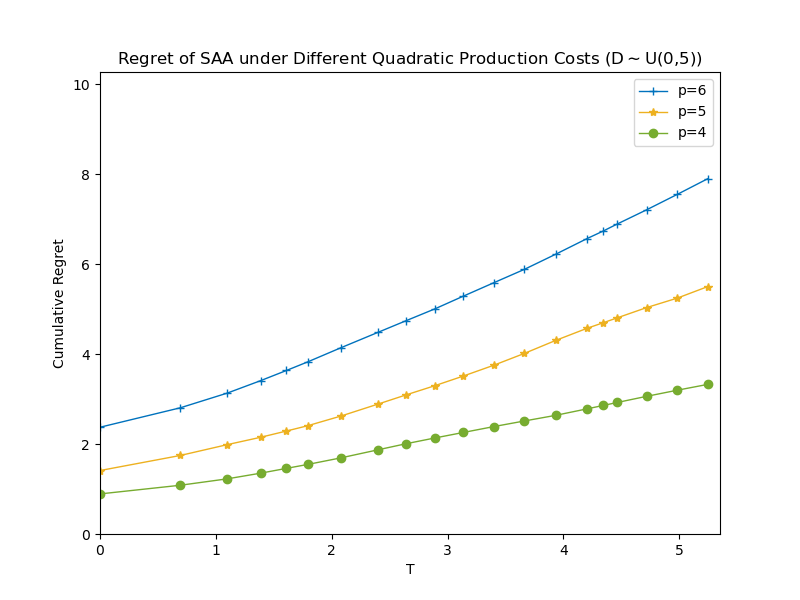}
\caption{Regret Performance for SAA under Quadratic Inventory and Production Costs}\label{fig:NR_general_costs}
\end{figure}

In Figure~\ref{fig:NR_general_costs}, we report the cumulative regret of the SAA method for $b \in\{1,3,6\}$ under the quadratic inventory cost and for $p\in\{4,5,6\}$ under the quadratic production cost. We plot the horizontal axis in the logarithmic scale. From this figure, we observe that the cumulative regret grows linearly with $\ln T$.

\subsection{Comparison with Other Algorithms}\label{subsec:comparison}
In this subsection, we compare the SAA policy with several classical algorithms from the data-driven inventory management literature, and show that the SAA policy remains competitive even in the small-sample regime.
We conduct numerical experiments for the following Huber Loss under four different distributions (Uniform $U(0,1)$, Poisson $\text{Poiss}(5)$, Normal $N(5,1)$, and Geometric $\text{Geom}(0.2)$ distributions) and set $\delta=0.01$, per-unit holding cost $h=1$, per-unit lost-sale cost $b=5$. $$
        c\left(x,d\right)=\left\{\begin{aligned}
		&b\delta \left(d-x-\delta/2\right),&x\in\left(-\infty,d-\delta\right],\\
                &b\left(x-d\right)^2/2,&x\in\left[d-\delta,d\right],\\
                &h\left(x-d\right)^2/2,&x\in\left[d,d+\delta\right],\\
                &h\delta \left(x-d-\delta/2\right),&x\in\left[d+\delta,+\infty\right),
                \end{aligned}
            \right.
    $$

The performance of an algorithm $\pi$ is measured by the \emph{relative average regret} defined as
\begin{equation}
\nonumber
    \frac{C^{\pi}_P(T)-T\cdot C^*_P}{T\cdot C_P^{*}} \times 100 \%,
\end{equation}
where $P$ denotes the problem instance, $C_P^{\pi}(T)$ is the expected cost incurred by the algorithm $\pi$ and $C_P^*$ is the optimal cost.

We test the performance of the following classic algorithms in the literature.
\begin{enumerate}[nolistsep]
    \item \textbf{Sample Average Approximation (SAA)}: This algorithm is our SAA policy applied to the Huber Loss.
    \item \textbf{Stochastic Gradient Descent (SGD)} in \cite{huh2009nonparametric}: This algorithm uses stochastic gradient descent to update the order-up-to level.
    \item \textbf{Minibatch SGD (MS-NV)} in \cite{lyu2023minibatch}: This algorithm (Minibatch SGD for NewsVendor, MS-NV) operates in epochs called minibatches, keeping a constant target inventory level within each minibatch and updating it only at the end using an SGD update.
    \item \textbf{Binary Search (BS)} in \cite{chen2020optimal}: This algorithm constructs confidence intervals of the gradient of cost function combined with binary search to find optimal order-up-to level.
\end{enumerate}

\begin{figure}[ht]
    \begin{center}
        \includegraphics[width =0.48\textwidth]{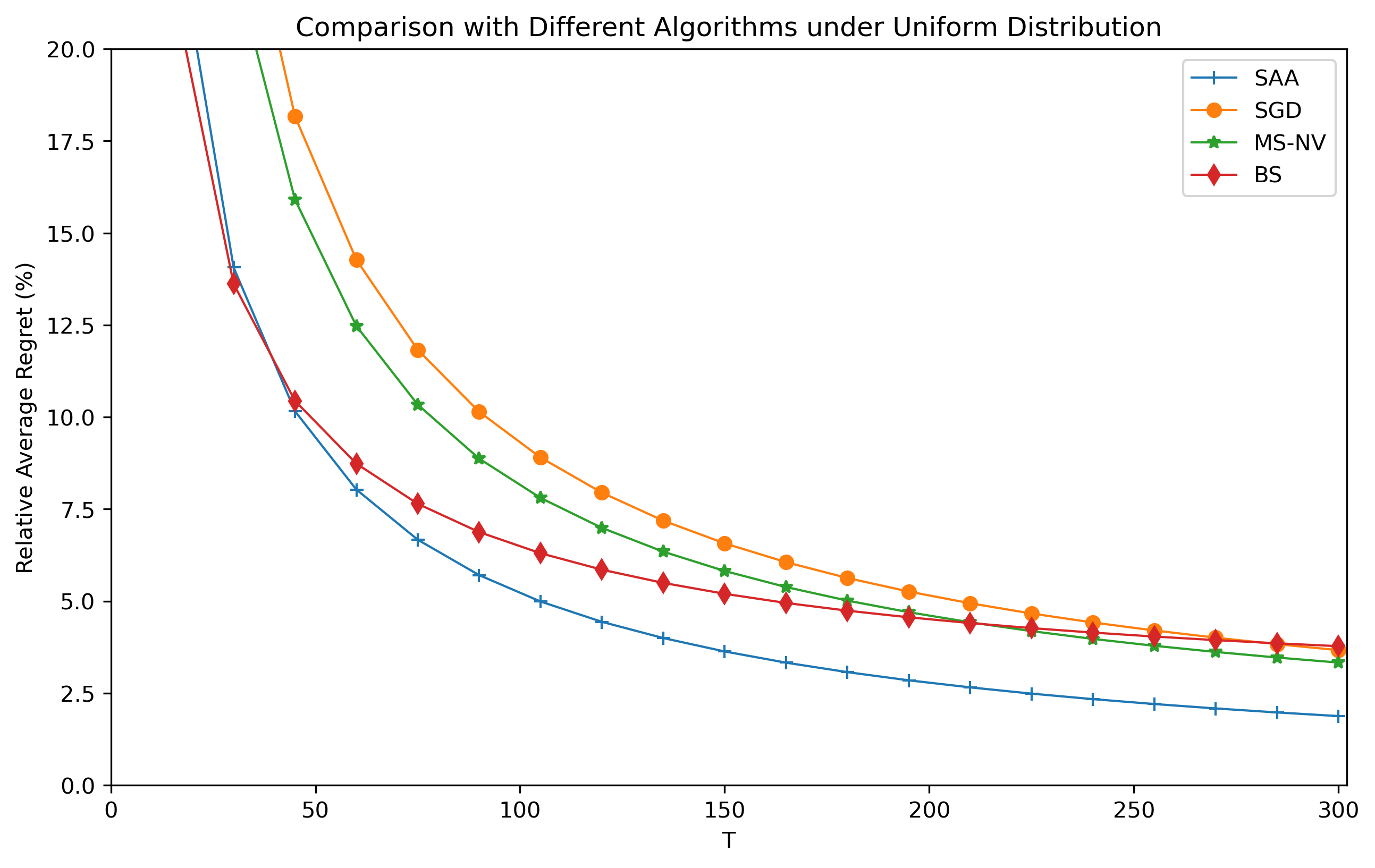}
        \includegraphics[width =0.48\textwidth]{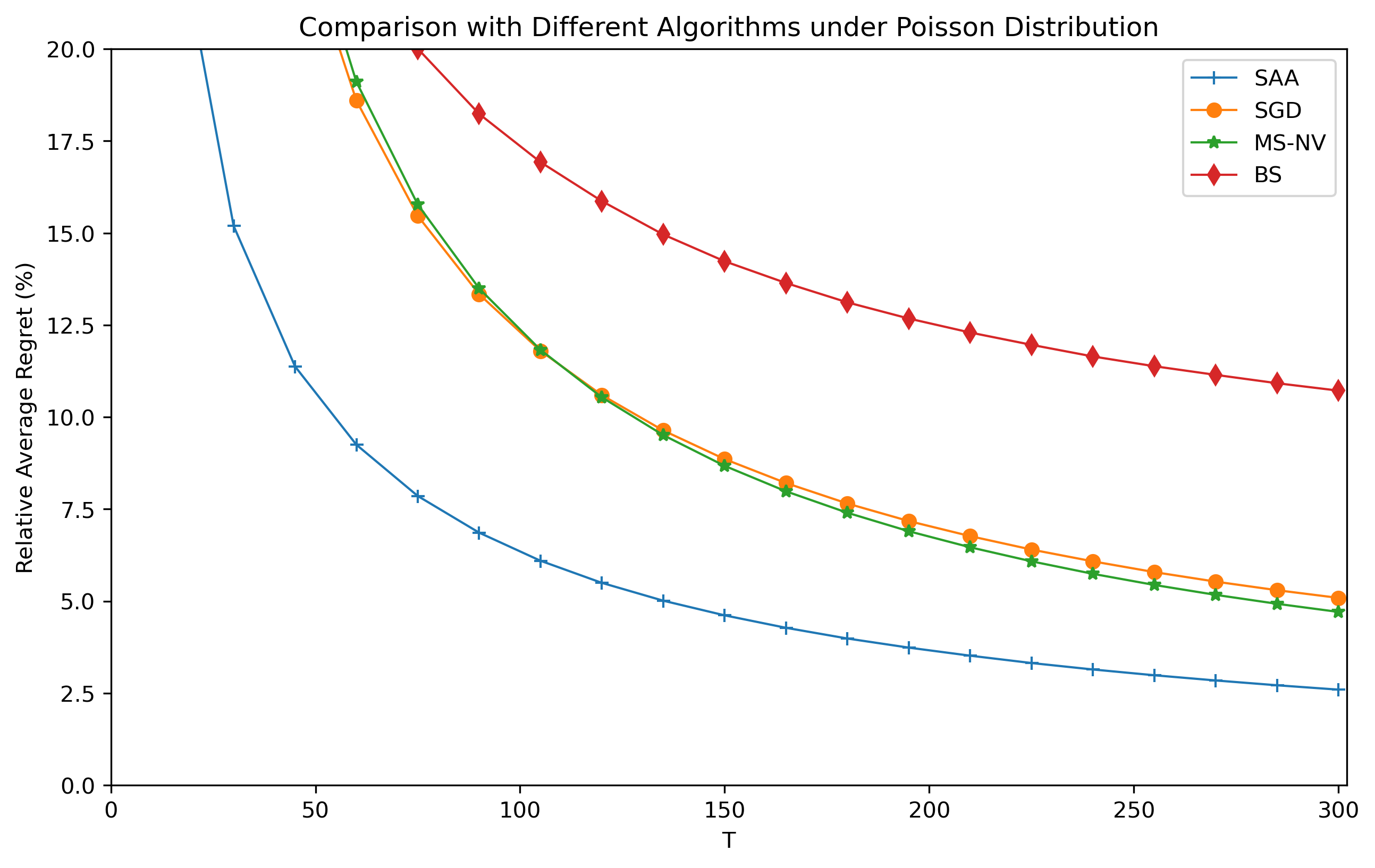}
        \includegraphics[width =0.48\textwidth]{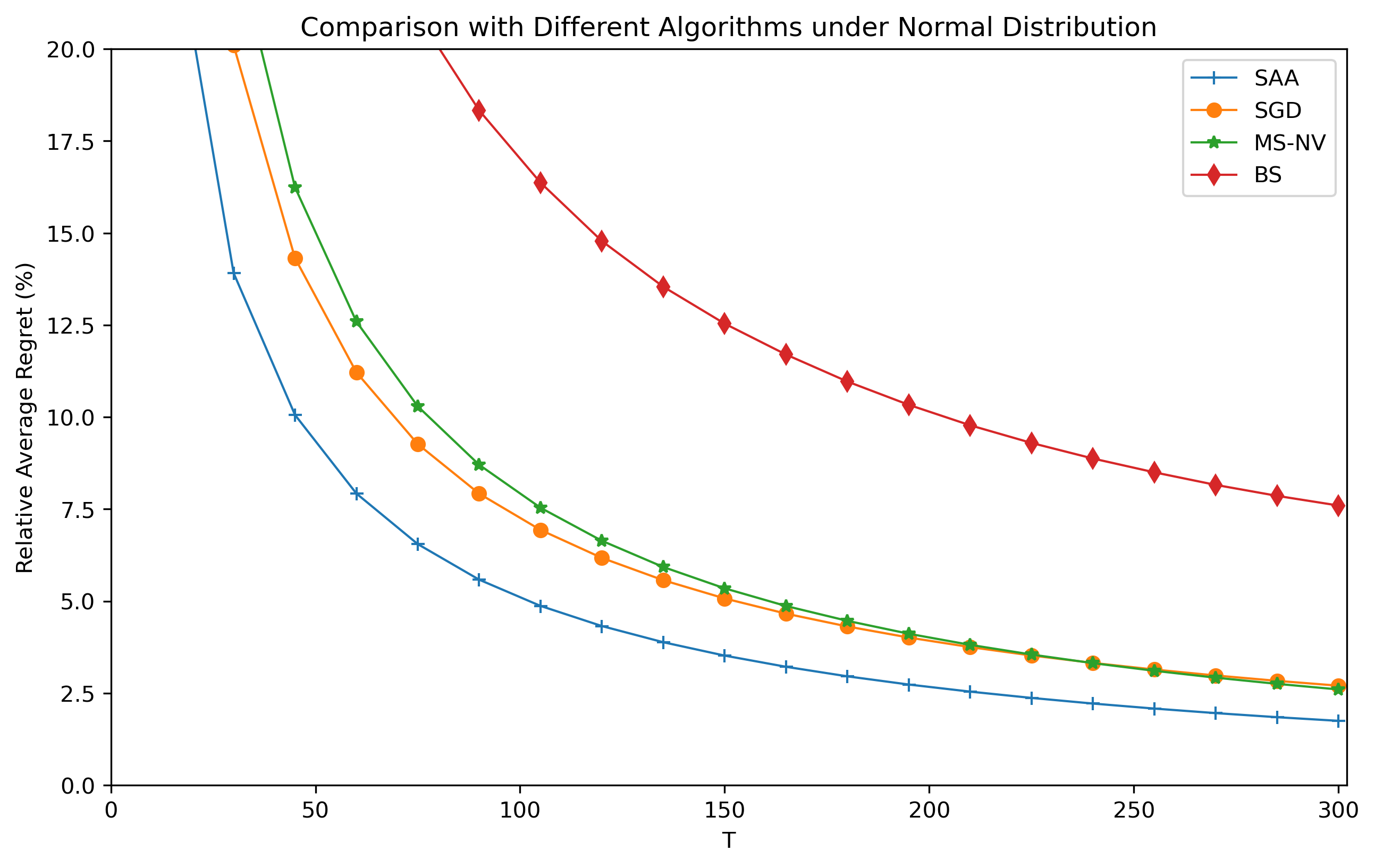}
        \includegraphics[width =0.48\textwidth]{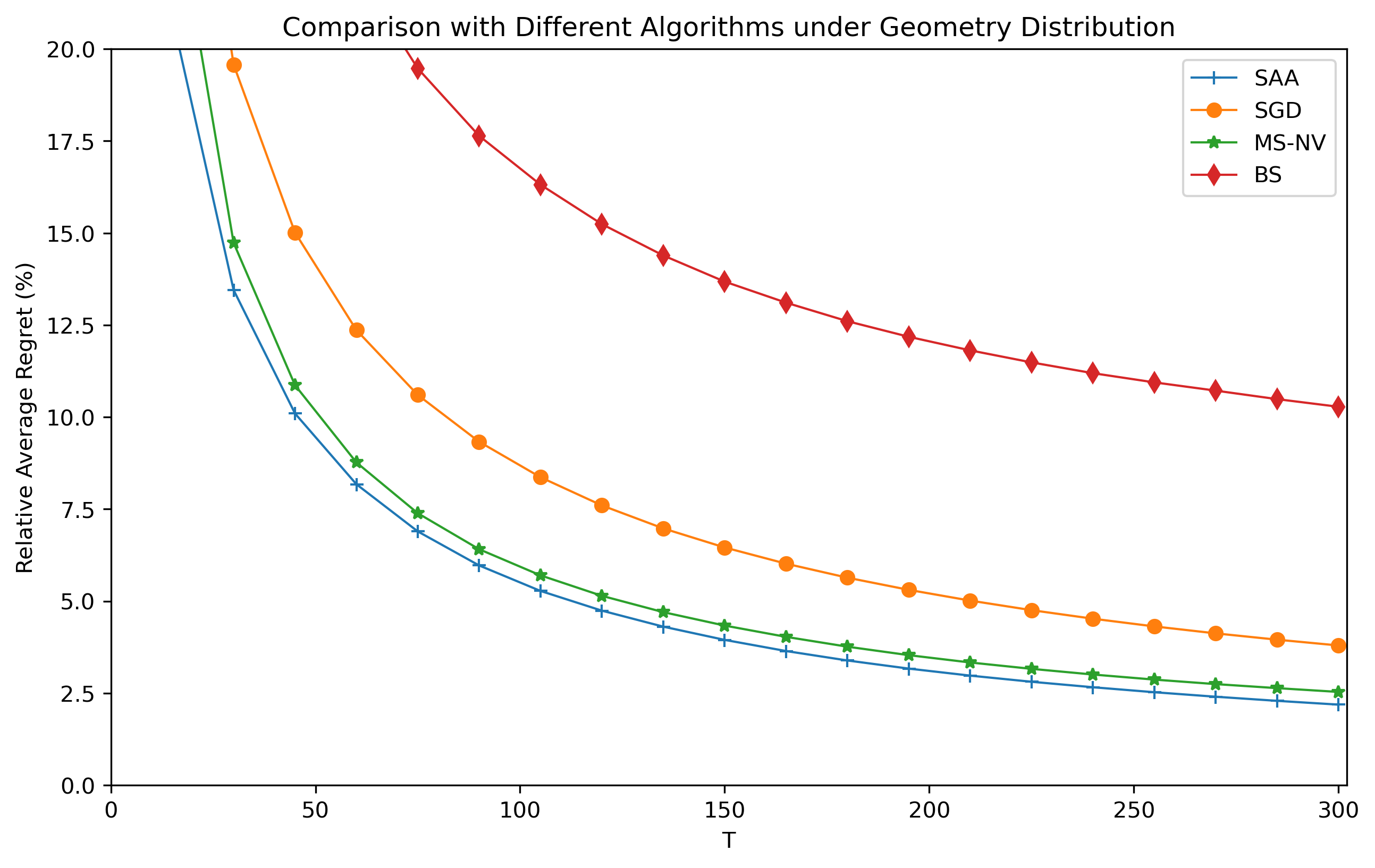}
    \end{center}
    \caption{Comparison with Different Algorithms under Different Distributions and Huber Loss}\label{fig:Com_SAA_sgd_minisgd_bs}
\end{figure}

For each algorithm, we fine-tune their parameters to achieve their peak performance for meaningful comparisons. For example, we choose the stepsize pattern of SGD to be $\eta/t$ and the batch size of MS-NV to be exponential. The relative average regret curves are presented in Figure~\ref{fig:Com_SAA_sgd_minisgd_bs}. 

Through the numerical results presented in Figure~\ref{fig:Com_SAA_sgd_minisgd_bs}, we note that the SAA policy performs very well and is competitive among many classical algorithms even under small samples.

\section{Conclusion}\label{sec:Conclusion}

In this paper, we generalize the structure of the linear-cost newsvendor problem to generic convexity conditions and provide a unified regret analysis of SAA for general sequential stochastic optimization problems.
By isolating the problem's structural properties, our approach deepens our understanding of the intrinsic mechanics that influence the performance of SAA.
Our results establish the regret rate optimality of SAA under both global and local strong convexity conditions for general sequential stochastic optimization problems. Therefore, we close the regret gaps for the linear-cost newsvendor problem.
Our upper bound result demonstrates that the regret performance of SAA is only influenced by $\alpha$ in the long run under \((\alpha,\beta)\)-local strong convexity.   This insight enhances our understanding of how local properties affect the long-term regret performance of decision-making strategies.
Our lower bound result is the first to match the existing upper bound with respect to both the parameter \(\alpha\) and the time horizon \(T\) simultaneously under \(\alpha\)-global strong convexity, which advances the theoretical understanding of the performance limit of data-driven methods in newsvendor problems.
Moreover, the techniques we propose not only enrich the regret analysis techniques (including both upper bounds and lower bounds) in the data-driven inventory management literature but also offer potential applications in broader data-driven decision-making problems. 

We conclude this paper by highlighting several future directions.
First, our approach may provide further insight into the algorithm design and analysis for more complex data-driven inventory problems.
 It would be interesting to study how the analysis techniques in this paper can be generalized to the optimization of more complicated inventory systems. For example, the multi-echelon inventory systems \citep{chen2001optimal}, the perishable inventory systems \citep{zhang2018perishable} and the lost-sales system with positive lead times \citep{zhang2020closing} under base-stock policies. Due to the complex dynamics, these systems cannot be described by the general framework of this paper. Second, it would be interesting to explore the optimal regret dependence of algorithms under the local gradient growth condition (Definition \ref{def:zeta_local}) in terms of parameters $(\alpha,\beta,\zeta)$ and $T$. Third, extending the analysis to non-stationary demand environments is another important direction \citep{keskin2023nonstationary,chen2025learning}. Such an extension would require combining the local-strong-convexity analysis developed here with tools for detecting distributional changes or restarting learning policies, while ensuring that SAA decisions can still enter the locally strongly convex region within each stationary epoch.

% \ACKNOWLEDGMENT{The authors sincerely thank the Department Editor, the Associate Editor, and two referees for their helpful comments and suggestions, which greatly improved the quality of this paper. The authors' names are listed in alphabetical order.}

\bibliographystyle{informs2014}
\bibliography{ref1.bib}

% \bibliographystyle{pomsref}

%  \let\oldbibliography\thebibliography
%  \renewcommand{\thebibliography}[1]{\oldbibliography{#1}\baselineskip14pt \setlength{\itemsep}{10pt}}
% \bibliography{ref1}

\ECSwitch \ECHead{E-Companion for {``Regret Optimality of Sample Average
Approximation for Data-Driven Newsvendor
Problems: A General Optimization Perspective''}}

\section{Some Useful Technical Lemmas}
\label{sec:useful-lemma}

By classic results and standard techniques in learning theory, we could translate pseudo-dimension into uniform convergence guarantees. For simplicity, we omit the definition of pseudo-dimension and the proof of the following lemma, please refer to \cite{pollard2012convergence} for details. 
\begin{lemma}\label{technical lemma concentration}
 Let $\mathcal{F}$ be a $b$-uniformly bounded class, and let $\left\{X_i\right\}_{i=1}^n$ be a collection of \emph{i.i.d.}~samples from random variable $X$.
   For any positive integer $n \geq 1$ and any scalar $\delta \geq 0$, there exists a positive constant $C_0>0$ such that
$$
\sup _{f \in \mathcal{F}}\left|\frac{1}{n} \sum_{i=1}^n f\left(X_i\right)-\mathbb{E}[f(X)]\right| \leq  C_0 b \sqrt{\frac{\mathrm{P}\text{-}\operatorname{dim}(\mathcal{F})}{N}}+\delta
$$
with probability at least $1-\exp \left(-\frac{n \delta^2}{2 b^2}\right)$, where $\mathrm{P}\text{-}\operatorname{dim}(\mathcal{F})$ is the pseudo dimension (or VC-subgraph dimension) of class $\mathcal F$. Following the derivation in \cite{pollard2012convergence}, $C_0$ can be upper bounded as $24\left( \sqrt{1 + 2\log 2} + \frac{\sqrt{\pi}}{2} \right)$.
\end{lemma}

\begin{lemma}[The van Trees inequality \citep{Richard1995ApplyofvanTrees}]\label{le:vTI} 
Let $(\mathcal{X},\mathcal{F},P_\theta:\theta\in\Theta,\mu)$ be a dominated family of distributions, where $\mathcal{X}$ is the sample space, $\mathcal{F}$ is the $\sigma$-algebra, $P_\theta$ is the probability measure with parameter $\theta$,  the parameter space $\Theta$ is a closed interval on the real line, $\mu$ is the dominating measure.  Let $f(x|\theta)$ denote the density of $P_\theta$ with respect to $\mu$, and $\pi$ be some probability distribution on $\Theta$ with a density $\lambda(\theta)$ with respect to the Lebesgue measure.
Suppose the following assumptions hold:\begin{enumerate}
    \item $\lambda(\cdot)$ and $f(x|\cdot)$ are absolutely continuous ($\mu$-almost surely),
    \item for any fixed $\theta\in \Theta$, $\E_\theta\left[\frac{\partial}{\partial\theta}\left(\log f\left(X|\theta\right)\right)\right]=0,$ where $\E_\theta$ denotes the expectation over $X\sim P_\theta$.
    \item $\lambda(\cdot)$ converges to zero at the endpoints of the interval $\Theta$.
\end{enumerate}
Let $\psi(\theta)$ be an absolute function. Then for any estimator $\hat{\theta}(X)$ based on sample $X\sim P_\theta$, it holds that
$$
    \E\left[\left(\hat{\psi}(x)-\psi(\theta)\right)^2\right]\geq \frac{\E[\psi'(\theta)^2]}{\E\left[I(\theta)\right]+I\left(\lambda\right)}, 
$$
where $\E$ denotes the expectation over the  joint distribution of $X$ and $\theta$, $I(\theta) = \E_\theta\left[(\frac{\partial}{\partial\theta}\left(\log f\left(X|\theta\right)\right))^2\right]$ and $I(\lambda) = \E\left[(\frac{\partial}{\partial\theta}\left(\log \lambda(\theta)\right))^2\right]$ is the the Fisher information for $\theta$ and $\lambda$ respectively.
\end{lemma}

\section{Omitted Proofs in Section \ref{sec:Upper bound of SAA}}
\label{sec:omitted-ub}
\subsection{Proof of Lemma \ref{le:le_uniform_convergence_g}}
\proof{Proof of Lemma \ref{le:le_uniform_convergence_g}.}
By Assumption \ref{ass:regular} and the definition of SAA, we know $|\hat{g}_t(\hat{x}_t)| \leq C_1/\sqrt{t}$.
Besides,
\[
|g(\hat{x}_t)| \leq |g(\hat{x}_t) - \hat{g}_t(x_t)| + |\hat{g}_t(\hat{x}_t)| \leq 2\sup_{x \in \mathcal{X}}|\hat{g}_t(x)-g(x)| + |\hat{g}_t(\hat{x}_t)| .
\]

We establish the concentration of $\sup_{x \in \mathcal{X}}|\hat{g}_t(x)-g(x)|$ by deriving the pseudo-dimension of the function class $\mathcal{F}=\{c'(x,\cdot): x \in \mathcal{X}\}$ and applying Lemma~\ref{technical lemma concentration}.

With a slight abuse of notation, we use $c'(x,d)$ to refer the gradient (or subgradient) with respect to $x$. 
Given any demand samples $d_1,d_2,\dots,d_m$ and $\tau_1,\tau_2,\dots,\tau_m \in \R$, consider the set
\[
S = \{ (\I[c'(x,d_1)<\tau_1],\I[c'(x,d_2)<\tau_2],\dots,\I[c'(x,d_t)<\tau_m]) : x \in \mathcal{X} \}.
\]
Since $c(x,d)$ is convex,  $c'(x,d)$ is non-decreasing in $x$ and the set $S$ has at most $m+1$ elements. It follows that the function class $\mathcal{F}=\{c'(x,\cdot): x \in \mathcal{X}\}$ has pseudo-dimension $m\leq 2$ (by $2^m \leq m+1$). Therefore, by Lemma~\ref{technical lemma concentration}, we know that with probability at least $1-\exp(-\frac{t\lambda^2}{2B^2})$,
\[
\sup_{x \in \mathcal{X}}|\hat{g}_t(x)-g(x)| \leq  \frac{\sqrt 2 C_0B}{\sqrt t} + \lambda,
\]
where $B$ is the upper bound on $c'(x,d)$ and $C_0$ is a universal constant in Lemma~\ref{technical lemma concentration}.
Therefore, by the above discussion, we obtain
\begin{align*}
\mathbb{P}\left[\left|g(\hat{x}_t)\right|\geq \lambda+\frac{4C_0B+C_1}{\sqrt t}\right]&\leq \mathbb{P}\left[2\sup_{x \in \mathcal{X}}|\hat{g}_t(x)-g(x)| + |\hat{g}_t(\hat{x}_t)| \geq \lambda+\frac{2\sqrt{2}C_0B+C_1}{\sqrt t}\right] \leq \mathrm{exp} \left(-\frac{t\lambda^2}{2B^2}\right),
\end{align*}
where the second inequality is due to $|\hat{g}_t(\hat{x}_t)| \leq C_1/\sqrt{t}$.
\Halmos
\endproof

\subsection{Regret Upper Bound under General Convexity}\label{sec:osqrtT}
This section presents an upper bound on the regret of the SAA method for general sequential stochastic optimization problems without any strong convexity conditions. Similar to the main context, we consider the one-dimensional $\mathcal{X}$ for simplicity. Note that the $\O(\sqrt{T})$ regret bound is not new for general problems. We present the result only for completeness and show the extensibility of our analysis techniques.
\begin{theorem}\label{tm:tm_general_case_no_local_convex_ass_ub}
    Suppose Assumption  \ref{ass:regular} holds. We have \[
        \sum_{t=1}^T\E\left[C(\hat{x}_t)-C(x^*)\right]\leq \tilde{K}_1\cdot\sqrt{T},
    \]
    where $\tilde{K}_1=2\mathcal{D}_{\mathcal{X}}(\sqrt{\pi}+2(4C_0B+C_1))$, is constant that does not rely on $\alpha,\beta,T$.
\end{theorem}
\proof{Proof of Theorem~\ref{tm:tm_general_case_no_local_convex_ass_ub}.}
By the convexity of $C(x)$, we have
\begin{align*}
    \E\left[C(\hat{x}_t)-C(x^*)\right]\leq \E\left[|g(\hat{x}_t)|\cdot|\hat{x}_t-x^*|\right]\leq\mathcal{D}_{\mathcal{X}}\cdot\E\left[|g(\hat{x}_t)|\right].
\end{align*}

By Lemma \ref{le:le_uniform_convergence_g}, we have \begin{equation}\label{eq:g_tail_without_bias}    
    \P\left[|g(\hat{x}_t)|\geq\lambda\right]\leq\P\left[|g(\hat{x}_t)|\geq \frac{\lambda}{2}+\frac{4C_0B+C_1}{\sqrt{t}}\right]\leq \exp\left(-\frac{t\lambda^2}{8B^2}\right), \quad \forall \lambda>\frac{2(4C_0B+C_1)}{\sqrt{t}}.
\end{equation}
Invoking the above inequality to the definition of $\mathbb{E}\left[|g(\hat{x}_t)|\right]$, we obtain \begin{align}\label{eq:ub_sqrt_result_Eg}
    \nonumber\mathbb{E}\left[|g(\hat{x}_t)|\right]&=\int_{0}^{\frac{2(4C_0B+C_1)}{\sqrt{t}}}\P\left[|g(\hat{x}_t)|\geq\epsilon\right]\mathrm{d}\epsilon+\int_{\frac{2(4C_0B+C_1)}{\sqrt{t}}}^{\infty}\P\left[|g(\hat{x}_t)|\geq\epsilon\right]\mathrm{d}\epsilon \\
    \nonumber&\leq \int_{0}^{\frac{2(4C_0B+C_1)}{\sqrt{t}}}1\mathrm{d}\epsilon+\int_{\frac{2(4C_0B+C_1)}{\sqrt{t}}}^{\infty}\exp\left(-\frac{t\epsilon^2}{8B^2}\right)\mathrm{d}\epsilon \\
    \nonumber&\leq \frac{2(4C_0B+C_1)}{\sqrt{t}}+\int_{0}^{\infty}\exp\left(-\frac{t\epsilon^2}{8B^2}\right)\mathrm{d}\epsilon \\
    &=\frac{2(4C_0B+C_1)+\sqrt{2\pi}B}{\sqrt{t}}.
\end{align}
Thus, the cumulative regret is bounded as \begin{align*}
    &\sum_{t=1}^T\E[C(\hat{x}_t)-C(x^*)]\leq \mathcal{D}_{\mathcal{X}}\sum_{t=1}^T\E\left[|g(\hat{x}_t)|\right]\leq (\sqrt{2\pi}B+2(4C_0B+C_1))\mathcal{D}_{\mathcal{X}}\sum_{t=1}^T\frac{1}{\sqrt{t}}.
\end{align*}

Using $\sum_{t=1}^Tt^{-1/2}\leq2\sqrt{T}$, we conclude that \[
    \sum_{t=1}^T\E[C(\hat{x}_t)-C(x^*)]\leq 2\mathcal{D}_{\mathcal{X}}(\sqrt{2\pi}B+2(4C_0B+C_1))\sqrt{T},
\]
where we complete the proof.
\Halmos\endproof

\section{Verification Illustration of Local Strong Convexity}\label{subsec:huber-lsc}

We verify the example in Section~\ref{sec:problem formulation}: under the Huber-type inventory cost with $D\sim U[a,a+L]$, the expected cost $C(x)=\mathbb{E}[c(x,D)]$ satisfies the $(\alpha,\beta)$-local strong convexity condition (Definition~\ref{def:local}) while failing to be globally strongly convex. The Huber-type inventory cost with parameters $b,h,\delta>0$ is defined as
\[
c(x,d)=\begin{cases}
b\delta\!\left(d-x-\tfrac{\delta}{2}\right), & x\leq d-\delta,\\[4pt]
\tfrac{b}{2}(x-d)^2, & d-\delta\leq x\leq d,\\[4pt]
\tfrac{h}{2}(x-d)^2, & d\leq x\leq d+\delta,\\[4pt]
h\delta\!\left(x-d-\tfrac{\delta}{2}\right), & x\geq d+\delta,
\end{cases}
\]
where $b$ and $h$ are the unit underage and overage costs, and $\delta>0$ controls the width of the quadratic region.

Since $c(x,d)$ is piecewise smooth in $x$, differentiating under the expectation gives
\begin{equation}\label{eq:huber-second-deriv}
    C''(x) = b\bigl[F(x+\delta)-F(x)\bigr] + h\bigl[F(x)-F(x-\delta)\bigr],
\end{equation}
where $F$ is the CDF of $D$. This follows from $\partial^2 c/\partial x^2(x,d) = b\cdot\mathbf{1}[x\leq d\leq x+\delta]+h\cdot\mathbf{1}[x-\delta\leq d\leq x]$ and taking expectations. For $x\in[a+\delta,\,a+L-\delta]$, both $[x,x+\delta]$ and $[x-\delta,x]$ are contained in the support $[a,a+L]$, so
\[
    C''(x) = b\cdot\frac{\delta}{L} + h\cdot\frac{\delta}{L} = \frac{(b+h)\delta}{L} =: \alpha_0 > 0.
\]
Since $C(x)$ is strictly convex and coercive, it has a unique minimizer {\[
x^*=a+\frac{b}{b+h}L+\frac{h-b}{2(h+b)}\delta.
\]} The unimodality of $C$ ensures $x^*\in[a+\delta,\,a+L-\delta]$ whenever {$\delta\leq\min\!\left\{\frac{2b}{3b+h}L,\,\frac{2h}{b+3h}L\right\}$}. Setting {$\beta = \min\bigl\{x^*-(a+\delta),\,(a+L-\delta)-x^*\bigr\}=\min\bigl\{\frac{b}{b+h}L-\frac{h+3b}{2h+2b}\delta,\frac{h}{b+h}L-\frac{3h+b}{2h+2b}\delta\bigr\}>0$}, we have
\[
    C''(x) = \alpha_0 > 0 \quad \forall\, x\in[x^*-\beta,\,x^*+\beta],
\]
which establishes the $(\alpha_0,\beta)$-local strong convexity of $C$ with explicit parameters $\alpha_0=(b+h)\delta/L$. On the other hand, for $x < a-\delta$ or $x > a+L+\delta$, both $[x,x+\delta]$ and $[x-\delta,x]$ are disjoint from $[a,a+L]$, so $C''(x)=0$ and $C$ fails to be globally strongly convex.

Remarkably, here we choose the uniform distribution as an illustrative example. The above argument applies similarly to many other distributions and could yield local strong convexity.

\section{Omitted Proofs in Section~\ref{sec:lower bound}}
\subsection{Proof of Lemma~\ref{lem:basic properties}}
\proof{Proof of Lemma~\ref{lem:basic properties}.}    
Since it is easy to verify the well-definiteness and Assumptions~\hyperref[ass:A1]{A.1} and \hyperref[ass:A2]{A.2} for our hard instance. In the following, we verify Assumptions~\hyperref[ass:A1]{A.3} and \hyperref[ass:A4]{A.4}.

  \noindent{\bf Verification of Assumption~\hyperref[ass:A3]{A.3}.} We will prove that the likelihood function $f(x|\cdot)$ is differentiable continuous, and consequently absolutely continuous.
    Fix $x\in[0,1]$, the likelihood function takes the following form $$
        f(x|\theta)=\left\{\begin{aligned}
		&2-\alpha,&\theta\in((-\infty,\Tilde{l}_1(x)]\cup[\Tilde{r}_1(x),\infty))\cap \Theta,\\
            &\alpha,&\theta\in[\Tilde{l}_2(x),\Tilde{r}_2(x)]\cap \Theta,\\
                &\alpha+(1-\alpha)\left(\cos \left(w_1\left(\tilde{r}_1(x)-\theta\right)\right)+1\right),&\theta\in[\tilde{r}_2(x),\tilde{r}_1(x)]\cap\Theta,\\
                &\alpha+(1-\alpha)(\cos (w_2(\theta-\tilde{l}_1(x)))+1),&\theta\in[\tilde{l}_1(x),\tilde{l}_2(x)]\cap\Theta,
                \end{aligned}
            \right.
    $$
    where $\tilde{l}_1(x)=x-(4\rho-\alpha)/(16-8\alpha)-3/4$, $\tilde{l}_2(x)=x-(4\rho-\alpha)/(16-8\alpha)-1/4-\rho/2$, $\tilde{r}_2(x)=x-(4\rho-\alpha)/(16-8\alpha)-\rho/2$, $\tilde{r}_1(x)=x-(4\rho-\alpha)/(16-8\alpha)$, $w_1=2\pi/\rho$, and $w_2=2\pi/\left(1-\rho\right)$.

   Through the above formulation, the derivative of the likelihood can be derived as  $$
        \frac{\mathrm{d}}{\mathrm{d}\theta}f(x|\theta)=\left\{\begin{aligned}
		&0,&\theta\in((-\infty,\Tilde{l}_1(x)]\cup[\Tilde{l}_2(x),\Tilde{r}_2(x)]\cup[\Tilde{r}_1(x),\infty))\cap \Theta,\\
                &(1-\alpha)w_1\sin \left(w_1\left(\tilde{r}_1(x)-\theta\right)\right),&\theta\in[\tilde{r}_2(x),\tilde{r}_1(x)]\cap\Theta,\\
                &-(1-\alpha)w_2\sin \left(w_2\left(\theta-\tilde{l}_1(x)\right)\right),&\theta\in[\tilde{l}_1(x),\tilde{l}_2(x)]\cap\Theta .
                \end{aligned}
            \right.
    $$
   Thus, it is easy to note that for any $x\in[0,1]$, $f(x|\cdot)$ is absolutely continuous in $\Theta$.

  \noindent{\bf Verification of Assumption~\hyperref[ass:A4]{A.4}.}
 Note that,
    \begin{align}
        \nonumber\E_\theta\left[\frac{\partial\log f\left(X|\theta\right)}{\partial\theta}\right]=&\int_{l_1(\theta)}^{l_2(\theta)}(1-\alpha)w_1\sin\left(w_1\left(x-\frac{4\rho-\alpha}{16-8\alpha}-\theta\right)\right)\mathrm{d}x\\
        \nonumber&-\int_{r_2(\theta)}^{r_1(\theta)}(1-\alpha)w_2\sin\left(w_2\left(\theta-x+\frac{3}{4}\right)\right)\mathrm{d}x.
    \end{align}
    Taking $y=w_1(x-l_1(\theta))$, $z=w_2(r_1(\theta)-x)$, by substituting the integral, one has \begin{align*}
        &\int_{l_1(\theta)}^{l_2(\theta)}(1-\alpha)w_1\sin\left(w_1\left(x-\frac{4\rho-\alpha}{16-8\alpha}-\theta\right)\right)\mathrm{d}x=(1-\alpha)\int_{0}^{\pi}\sin\left(y\right)\mathrm{d}y=2(1-\alpha), \\
        &-\int_{r_2(\theta)}^{r_1(\theta)}(1-\alpha)w_2\sin\left(w_2\left(\theta-x+\frac{3}{4}\right)\right)\mathrm{d}x=-(1-\alpha)\int_{0}^{\pi}\sin\left(z\right)\mathrm{d}z=-2(1-\alpha).
    \end{align*}
    Combining the above equations,  we could get the conclusion.
    \Halmos
    \endproof

\subsection{Proof of Lemma~\ref{le:position of optimal solution}}

\proof{Proof of Lemma~\ref{le:position of optimal solution}.}
 Let $F_\theta$ denote the cumulative distribution of $f(\cdot|\theta)$. For any $\theta\in\Theta$, since $-\alpha/16\leq\theta\leq \alpha/16$, it holds that \begin{align}
        \nonumber F_\theta\left(l_2(\theta)\right)&=\int_{0}^{l_1(\theta)}f\left(x|\theta\right)\mathrm{d}x+\int_{l_1(\theta)}^{l_2(\theta)}f\left(x|\theta\right)\mathrm{d}x \\
        &=\left(2-\alpha\right)\left(\frac{4\rho-\alpha}{16-8\alpha}+\theta\right)+ \frac{\rho}{2}\leq \rho-\frac{\alpha}{8}+\frac{\left(2-\alpha\right)\alpha}{16}< \rho.\label{eq:lb_F_at_l2_less_rho}
    \end{align}
    Similarly, we have
    \begin{align}
        \nonumber F_\theta\left(r_2(\theta)\right)&=\int_{0}^{l_1(\theta)}f\left(x|\theta\right)\mathrm{d}x+\int_{l_1(\theta)}^{l_2(\theta)}f\left(x|\theta\right)\mathrm{d}x+\int_{l_2(\theta)}^{r_2(\theta)}f\left(x|\theta\right)\mathrm{d}x\\
        &=\left(2-\alpha\right)\left(\frac{4\rho-\alpha}{16-8\alpha}+\theta\right)+\rho/2+\frac{\alpha}{4} =\rho+\frac{\alpha}{8}+\left(2-\alpha\right)\theta\geq \rho+\frac{\alpha}{8}-\frac{\left(2-\alpha\right)\alpha}{16}> \rho.\label{eq:lb_F_at_l2_large_rho}
    \end{align}
   Combining Eq.~\eqref{eq:lb_F_at_l2_less_rho} and Eq.~\eqref{eq:lb_F_at_l2_large_rho}, by definition of $x_\theta^*$, we have $\left[l_2(\theta),r_2(\theta)\right]$.
        Thus, solving equation $$
\rho=\int_{0}^{l_1(\theta)}f\left(x|\theta\right)\mathrm{d}x+\int_{l_1(\theta)}^{l_2(\theta)}f\left(x|\theta\right)\mathrm{d}x+\int_{l_2(\theta)}^{x_\theta^*}f\left(x|\theta\right)\mathrm{d}x=\left(2-\alpha\right)l_1(\theta)+\left(l_2(\theta)-l_1(\theta)\right)+\alpha\left(x_\theta^*-l_2(\theta)\right),
    $$
    we obtain \begin{align*}
        x_\theta^*=\frac{\rho}{2}+\frac{4\rho-\alpha}{16-8\alpha}+\frac{1}{8}-\left(\frac{2}{\alpha}-2\right)\theta,
    \end{align*}
    which completes the proof of this lemma.
\Halmos

\subsection{Proof of Lemma~\ref{le:fisherinformation}}
\proof{Proof of Lemma~\ref{le:fisherinformation}.}
    For any $t\geq 1$, let $I_t$ denote the expected Fisher information for $\theta$ at time $t$. It holds that, \begin{align}\label{eq:lb_decompose_of_Fisher_info}
        \nonumber I_t(\theta)&=\E_{\theta}\left[\left(\frac{\partial\log f\left(X_1,\dots,X_{t}|\theta\right)}{\partial\theta}\right)^2\right]\\
        &=\sum_{i=1}^t\E_{\theta}\left[\left(\frac{\partial\log f\left(X_i|\theta\right)}{\partial\theta}\right)^2\right]+2\sum_{1\leq i<j\leq t}\E_{\theta}\left[\frac{\partial\log f\left(X_i|\theta\right)}{\partial\theta}\right]\E_{\theta}\left[\frac{\partial\log f\left(X_j|\theta\right)}{\partial\theta}\right]=tI_1(\theta).
    \end{align}
    where the first equality is due to the independence of samples, and the second equality is due to Assumption~\hyperref[ass:A4]{A.4}.
    Noting that $f(\cdot|\theta)$ is supported on $\left[0,1\right]$, we have \begin{align*}
        I_1(\theta)&=\E_{\theta}\left[\left(\frac{\mathrm{d}}{\mathrm{d}\theta}\log f(x|\theta)\right)^2\right]=\int_0^1 \frac{\left(\frac{\mathrm{d}}{\mathrm{d}\theta}f(x|\theta)\right)^2}{f(x|\theta)}\mathrm{d}x.\nonumber
 \\&=\int_{l_1(\theta)}^{l_2(\theta)}\frac{(1-\alpha)^2w_1^2\sin^2\left(w_1\left(x-l_1(\theta)\right)\right)}{1+(1-\alpha)\cos\left(w_1\left(x-l_1(\theta)\right)\right)} \mathrm{d}x
        +\int_{r_2(\theta)}^{r_1(\theta)}\frac{(1-\alpha)^2w_2^2\sin^2\left(w_2\left(r_1(\theta)-x\right)\right)}{1+(1-\alpha)\cos\left(w_2\left(r_1(\theta)-x\right)\right)} \mathrm{d}x\\
        \nonumber&=(1-\alpha)^2w_1\int_0^{\pi}\frac{1-\cos^2\left(y\right)}{1+(1-\alpha)\cos\left(y\right)}\mathrm{d}y+(1-\alpha)^2w_2\int_0^{\pi}\frac{1-\cos^2\left(z\right)}{1+(1-\alpha)\cos\left(z\right)}\mathrm{d}z,
    \end{align*}
    where the last equality is due to $y=w_1(x-l_1(\theta))$, $z=w_2(r_1(\theta)-x)$ and by substitution of the integral. Since $1+(1-\alpha)\cos x>0$, and $1-\cos x\leq 2$ holds for any $x$, it is easy to calculate that \begin{align}
        \frac{1-\cos^2(x)}{1+(1-\alpha)\cos(x)}\leq \frac{2\left(1+\cos(x)\right)}{1+(1-\alpha)\cos (x)}=\frac{2}{1-\alpha}\frac{(1-\alpha)\left(1+\cos(x)\right)}{\alpha+(1-\alpha)\left(1+\cos (x)\right)}\leq\frac{2}{1-\alpha}, \nonumber
    \end{align}
    and it follows that \begin{align*}
        \nonumber I_1(\theta)&=(1-\alpha)^2(w_1+w_2)\int_0^{\pi}\frac{1-\cos^2(x)}{1+(1-\alpha)\cos(x)}\mathrm{d}x \\
        \nonumber&\leq (1-\alpha)^2(w_1+w_2)\int_0^{\pi}2(1-\alpha)^{-1}\mathrm{d}x=2\pi(1-\alpha)(w_1+w_2).
    \end{align*}
        Combine this with Eq.~\eqref{eq:lb_decompose_of_Fisher_info}, we get the conclusion that for any $\theta\in\Theta$, one has 
       $I_t(\theta)=tI_1(\theta)\leq 2\pi(w_1+w_2)t,$
    where $w_1 = 2\pi/\rho$ and $w_2=2\pi/(1-\rho)$.
\Halmos

\section{Constant Regret under Uniform Distributions}
\label{sec:ub-uniform}

For the linear-cost newsvendor problems (i.e., $c(x,d)=h\cdot(x-d)^++ b\cdot(d-x)^+$),
\cite{besbes2013implications} proved the minimax lower bound under the global minimal separation condition (see their Theorem 1) by constructing a family of uniform distributions with CDF $F_{\theta},~\theta \in [0, 1/2]$, where
\begin{align}\label{hard instance in bes}
    F_\theta(x)=\left(x-\theta\right)\left(1-\theta\right)^{-1}, \quad x\in [\theta,1].
\end{align}

Let $U(a,{{p}})$ denote the uniform distribution on $[a, {{p}}]$, and let $F_{a,{{p}}}$ be its CDF. 
We denote the uniform distribution family used in the proof of Theorem 1 in \cite{besbes2013implications} as
$$\mathcal{F}_{u} = \{F_{\theta,1} \mid \theta \in [0,1/2]\}.$$

Next, we show that there exists a policy that can achieve constant regret under uniform distributions. Specifically, we define critical ratio $\rho = b/(h+b)$ and policy $\pi = (\pi_t)_{t=1}^T$ as
\begin{equation}
\label{eq:mle}
\pi_t\left(D_1,\dots,D_t\right)=\left(1-\rho\right)\min\left\{D_1,\dots,D_t\right\}+\rho\max\left\{D_1,\dots,D_t\right\},
\end{equation}
where $D_1,\dots,D_t$ are \emph{i.i.d.}~samples from $U(a,{{p}})$. Note that the policy $\pi$ is inspired by the \textit{Maximum Likelihood Estimation (MLE)} of parameters $a,{{p}}$ for uniform distributions.

We have the following theorem for the performance of policy $\pi$.
\begin{theorem}\label{thm:minimax}
For any fixed $a,{{p}}\in\mathbb{R}$, we have
\[
\mathcal{R}^{\pi}\left(F_{a,{{p}}},T\right)\leq \left(\rho^2+\left(1-\rho\right)^2\right)(h+b)\left({{p}}-a\right).
\]
\end{theorem}

\proof{Proof of Theorem~\ref{thm:minimax}.}
Define $Y_t=\min\{D_1,\dots,D_t\}$, $Z_t=\max\{D_1,\dots,D_t\}$, then we know 
\[
\hat{x}_t :=\pi_t\left(D_1,\dots,D_t\right)=\left(1-\rho\right)Y_t+\rho Z_t.
\]

When the demand distribution is $U(a,{{p}})$, we know the newsvendor solution is
$x_{a,{{p}}}^*=\left(1-\rho\right)a+\rho {{p}}.$

Some calculation gives that $C^{\prime\prime}(x)=(h+b)/({{p}}-a)$. Therefore, from the smoothness of $C(x)$, we have 
\begin{align}\nonumber
C(\hat{x}_t)-C\left(x_{a,{{p}}}^*\right)\leq \frac{h+b}{2\left({{p}}-a\right)}\left(\hat{x}_t-x_{a,{{p}}}^*\right)^2.        
\end{align}  
It follows that
\begin{align}
\nonumber
\mathcal{R}^{\pi}\left(F_{a,{{p}}},T\right)&\leq
\frac{h+b}{2({{p}}-a)}\sum_{t=1}^T \mathbb{E}\left[\left(\hat{x}_t-x_{a,{{p}}}^*\right)^2\right]\\
&\leq \frac{h+b}{{{p}}-a}\sum_{t=1}^T\left(\left(1-\rho\right)^2\mathbb{E}\left[\left(Y_t-a\right)^2\right]+\rho^2\mathbb{E}\left[\left(Z_t-{{p}}\right)^2\right]\right).\label{eq:appen_D_expression_regret}
\end{align}
Then we estimate the right-hand side terms of the above inequality. Some simple calculation gives
\begin{align*}
        &\mathbb{E}\left[Y_t\right]\nonumber=\int_a^{{{p}}}\frac{t}{{{p}}-a}\left(\frac{{{p}}-x}{{{p}}-a}\right)^{t-1}x\mathrm{d}x
    =a+\frac{{{p}}-a}{t+1},
        \\&\nonumber\E \left[Y_t^2\right]=\frac{t}{\left({{p}}-a\right)^t}\int_a^{{{p}}}\left({{p}}-x\right)^{t-1}x^2\mathrm{d}x=a^2+\frac{2\left({{p}}-a\right)}{t+1}\left(a+\frac{{{p}}-a}{t+2}\right).
    \end{align*}
Therefore, we get that
    \[
        \E \left[\left(Y_t-a\right)^2\right]=\E \left[Y_t^2\right]+a^2-2a\E \left[Y_t\right]=\frac{2\left({{p}}-a\right)^2}{\left(t+1\right)\left(t+2\right)}.
    \]
Similarly, we can show that 
    \[
        \E \left[\left(Z_t-{{p}}\right)^2\right]=\frac{2\left({{p}}-a\right)^2}{\left(t+1\right)\left(t+2\right)}.
    \]
Combine the above equations with Eq.~\eqref{eq:appen_D_expression_regret}, we obtain
\begin{align*}
\mathcal{R}^{\pi}\left(F_{a,{{p}}},T\right)&\leq \frac{h+b}{{{p}}-a}\sum_{t=1}^T\left(\left(1-\rho\right)^2\mathbb{E}\left[\left(Y_t-a\right)^2\right]+\rho^2\mathbb{E}\left[\left(Z_t-{{p}}\right)^2\right]\right) \\
&\leq \frac{h+b}{{{p}}-a}\sum_{t=1}^T\left(\left(1-\rho\right)^2\frac{2\left({{p}}-a\right)^2}{\left(t+1\right)\left(t+2\right)}+\rho^2\frac{2\left({{p}}-a\right)^2}{\left(t+1\right)\left(t+2\right)}\right) \\
&\leq \left(\rho^2+\left(1-\rho\right)^2\right)(h+b)\left({{p}}-a\right),
\end{align*}
where we complete the proof.
\Halmos
\endproof

By the above conclusion, we obtain the following theorem regarding minimax regret over the uniform distribution family $\mathcal{F}_{u} = \{F_{\theta,1} \mid \theta \in [0,1/2]\}$, where we have ${{p}} -a \leq 1$.

\begin{theorem}
\label{thm:constant-regret}
The minimax regret over $\mathcal{F}_u$ satisfies\begin{align}
\nonumber\underset{\pi\in\Pi}{\inf}\underset{F\in\mathcal{F}_u}{\sup}\mathcal{R}^{\pi}(F,T) \leq \left(\rho^2+\left(1-\rho\right)^2\right)(h+b).
        \end{align}
\end{theorem}

The above theorem contradicts the proof of Theorem 1 in \cite{besbes2013implications}. We also find a subtle technical oversight in their lower bound proof, which might be the cause of such a contradiction.

We now elaborate on this technical oversight to make the argument more transparent. \cite{besbes2013implications} considered the linear-cost newsvendor problem $c(x,d)=h\cdot(x-d)^++b\cdot(d-x)^+$ and constructed a parametric family of uniform demand distributions with CDFs $F_\theta$, $\theta\in[0,1/2]$, given in Eq.~\eqref{hard instance in bes}. They then chose a prior density
\[
q(\theta)=2\cos^2\bigl(2\pi(\theta-1/4)\bigr)\mathbb{I}\bigl[\theta\in[0,1/2]\bigr]
\]
on $\theta$ and invoked the \emph{van Trees inequality} to claim that, for any estimator $\hat{\theta}_t$ based on $\mathbf{D}_{t-1}=(D_1,\dots,D_{t-1})$,
\[
\int_0^{1/2}(\hat{\theta}_t-\theta)^2 q(\theta)\,d\theta
\;\ge\; \bigl(I_t+I(q)\bigr)^{-1},
\]
where $I_t$ and $I(q)$ denote the Fisher information of the observations and the prior, respectively.

The regularity conditions required for the van Trees inequality, however, are violated in this construction. In particular, applying the van Trees inequality requires the likelihood $f(x\mid\theta)$ to be absolutely continuous as a function of $\theta$; see, e.g., \cite{Richard1995ApplyofvanTrees}. Under the above family, for any fixed $x\in(0,1/2)$,
\[
f(x\mid\theta)=\frac{1}{1-\theta}\,\mathbb{I}[0\le \theta\le x].
\]
As a function of $\theta$, $f(x\mid\theta)$ has a jump discontinuity at the breakpoint $\theta=x$: when $\theta<x$, we have $f(x\mid\theta)=(1-\theta)^{-1}$, whereas when $\theta>x$, we have $f(x\mid\theta)=0$. Hence, $f(x\mid\cdot)$ is not even continuous, let alone absolutely continuous, in $\theta$.

This discontinuity undermines the validity of the use of the van Trees inequality, because the derivative of the likelihood with respect to $\theta$ is not well-defined at the breakpoint $\theta=x$, and the Fisher-information-based derivation no longer applies. Therefore, the lower-bound argument in \cite{besbes2013implications} cannot be justified under their own hard instance. This is consistent with our Theorem~\ref{thm:constant-regret}, which shows that the MLE-type policy in Eq.~\eqref{eq:mle} achieves $\mathcal{O}(1)$ regret on this very family, and with the numerical validation reported in Section~\ref{subsec:numerical-constant-regret}.

\subsection{Numerical Validation for the Constant Regret}\label{subsec:numerical-constant-regret}
\begin{figure}[h!]
    \centering
    \includegraphics[width=0.48\textwidth]{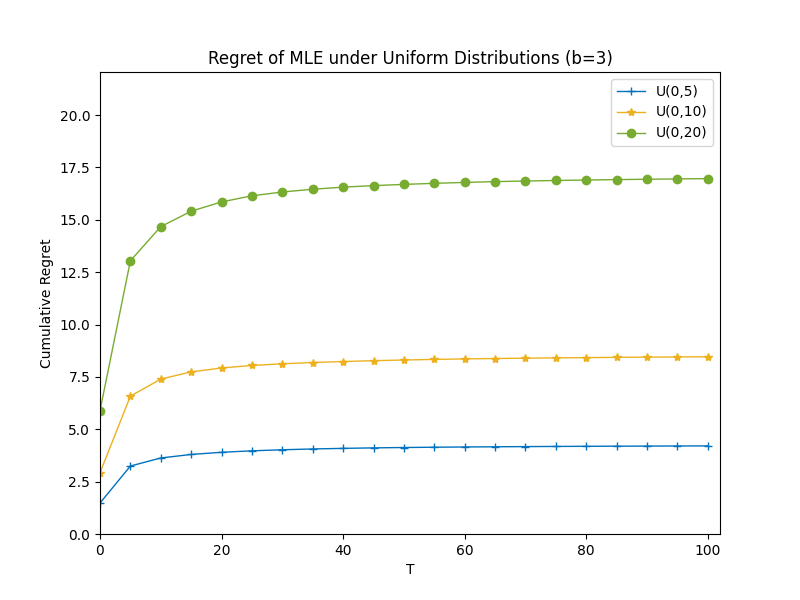}
    \hspace{0cm} 
    \includegraphics[width=0.48\textwidth]{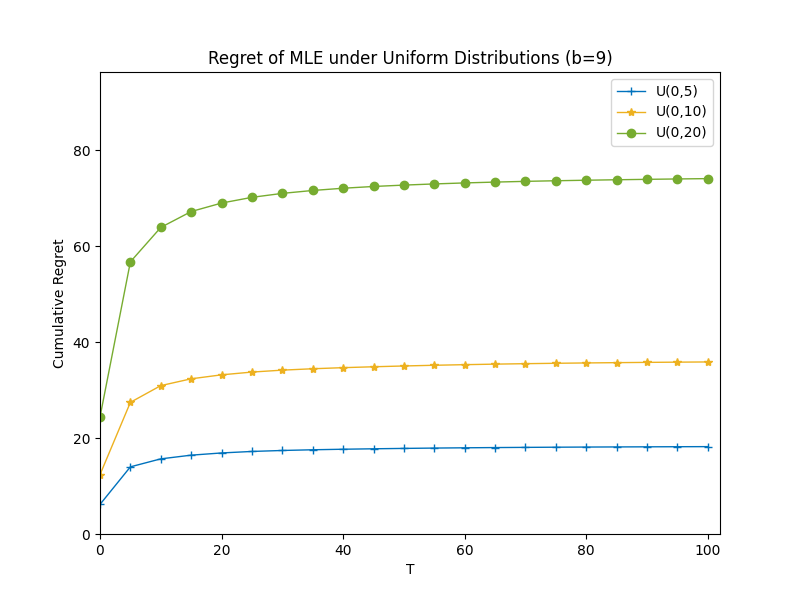}
\caption{Regret Performance for MLE policy under Different Distributions and Underage Costs}\label{fig:NR_unifrom_MLE_alpha}
\end{figure}

We also complement the conclusion of Theorem \ref{thm:constant-regret} with numerical experiments, which show that the MLE policy defined by Eq.~\eqref{eq:mle} can achieve constant regret. In the simulation, we consider the linear inventory cost function to be $c(x,d)=h\cdot(x-d)^++p\cdot(d-x)^+$. To test the robustness of our conclusion, we conduct experiments under different underage cost parameters $b = 3,~9$ and uniform distributions $U(0,{{p}})$, where ${{p}} = 5,~10,~20$. The cumulative regret curves of the MLE policy are presented in Figure~\ref{fig:NR_unifrom_MLE_alpha}.

As we can see from Figure~\ref{fig:NR_unifrom_MLE_alpha},
the MLE policy enjoys constant order, i.e., $\mathcal{O}(1)$ regret as $T$ grows from $1$ to $100$ across all problem instances, which validates our finding from a numerical perspective.

\section{Details Omitted in Section~\ref{sec:extension}}

\subsection{Sample Complexity Results} \label{subsec:SC_finite_supp}
\begin{proposition}\label{Prop:sample_complexity}
    Suppose Assumption \ref{ass:regular} and  \ref{ass:demand-nvp} hold, and $C(x)$ satisfies the $(\alpha,\beta)$-local strong convexity condition. Let $N_{\epsilon,\delta}$ represent the number of samples needed for the SAA method to achieve accuracy $\epsilon$ with probability at least $1-\delta$. For all $0<\delta<1/e$, without loss of generality, we have \begin{align*}
        N_{\epsilon,\delta}=\left\{\begin{array}{ll}
            \mathcal{O}\left(\frac{1}{\alpha}\frac{1}{\epsilon}\log \frac{1}{\delta}\right)&\text{if}\quad \epsilon<\alpha\beta^2/2,\\~\\
            \mathcal{O}\left(\bar{D}^2\frac{1}{\epsilon^2}\log \frac{1}{\delta}\right)&\text{if}\quad \epsilon\geq\alpha\beta^2/2.
        \end{array}\right.
    \end{align*}
\end{proposition}
This result consists of two parts depending on the relationship between $\epsilon$ and $C(x^*\pm\beta)-C(x^*)$.  
At a high level, we use the relationship to infer whether $\hat{x}_t$ should enter $[x^*-\beta,x^*+\beta]$ and apply the corresponding convexity to obtain the complexity results, i.e.,
\begin{itemize}
    \item When $\epsilon < C(x^*\pm\beta) - C(x^*)$, we obtain that $\hat x_t \in [x^*-\beta,x^*+\beta]$ and leverage the local strong convexity of $C(x)$ to obtain a sample complexity of $\mathcal{O}(\alpha^{-1}\epsilon^{-1}\log \delta^{-1})$.
    \item When $\epsilon > C(x^*\pm\beta) - C(x^*)$, we cannot conclude that $\hat x_t \in [x^*-\beta,x^*+\beta]$ and instead rely on the global convexity of $C(x)$ to derive a sample complexity of $\mathcal{O}(\epsilon^{-2}\log \delta^{-1})$.
\end{itemize}

When $C(\hat{x}_t) - C(x^*) < \epsilon < \alpha\beta^2/2$, we have $\hat{x}_t \in [x^* - \beta, x^* + \beta]$. By the $(\alpha, \beta)$-local strong convexity, this implies
$C(\hat{x}_t) - C(x^*) \leq (2\alpha)^{-1} |g(\hat{x}_t)|^2$.
Furthermore, by Eq.~\eqref{eq:g-squre} in Theorem~\ref{tm:tm_main_regret_ub}, we have $\mathbb{E}[|g(\hat{x}_t)|^2] \leq \mathcal{O}(t^{-1})$. This allows us to exploit the local strong convexity near $x^*$, yielding a sample complexity of $\mathcal{O}(\alpha^{-1}\epsilon^{-1}\log \delta^{-1})$ in this regime.
For the case where $\epsilon > \alpha\beta^2/2$, we directly apply the global convexity of $C(x)$ to achieve a sample complexity of $\mathcal{O}(\epsilon^{-2}\log \delta^{-1})$.

Meanwhile, Proposition EC.1 explicitly characterizes how statistical accuracy depends on the structural parameters $\alpha$ and $\beta$. Specifically, the dependence on $\alpha$ appears through a factor of $1/\alpha$: a larger $\alpha$, corresponding to stronger curvature, leads to faster concentration of the empirical minimizer around the true optimal solution and hence reduces the required sample size proportionally.

The parameter $\beta$ enters the bound through a term of order $1/(\alpha\beta)$, reflecting the cost incurred before the SAA solution enters the locally strongly convex neighborhood of the optimum. Thus, $\beta$ primarily affects the duration of the transient learning phase rather than the asymptotic convergence rate. Once the sample size exceeds the transient threshold $1/(\alpha\beta)^2$, the convergence behavior is effectively governed by $\alpha$ alone, driven mainly by the global strong convexity condition.

This structural decomposition is consistent with the regret bound $O!\left(\frac{1}{\alpha}\log T + \frac{1}{\alpha\beta}\right)$ established in the main text. It also provides a transparent interpretation of how the sample complexity depends on $\alpha$ and $\beta$.

\proof{Proof of Proposition~\ref{Prop:sample_complexity}.} By that $C(x)$ is $\alpha$-strongly convex on $[x^*-\beta,x^*+\beta]\cap\mathcal{X}$, we obtain\[
    C(x^*\pm\beta)-C(x^*)=\int_{x^*}^{x^*\pm\beta}g(y)\mathrm{d}y\geq \int_{x^*}^{x^*\pm\beta}\alpha(y-x^*)\mathrm{d}y= \frac{1}{2}\alpha\beta^2.
\]
Combining the above equation with the convexity of $C(x)$, we know that $C(\hat{x}_t)-C(x^*)<\alpha\beta^2/2$ implies $\hat{x}_t\in[x^*-\beta,x^*+\beta]$. Thus, we can combine the conclusion with (local) convexity to obtain complexity.

\noindent{\bf Case 1: $0<\epsilon<\alpha\beta^2/2$}. In this case, when $t\geq(4C_0B+C_1)^2/(2\alpha\epsilon)$, by simple calculation, we have
\begin{align}\label{eq:eq_sample_complexity_epsilon_small_first_step}
    \P[C(\hat{x}_t)-C(x^*)\leq \epsilon]\geq \P[|g(\hat{x}_t)|\leq \sqrt{2\alpha\epsilon}]\geq 1-\exp\left(-\frac{t}{2B^2}\cdot\left(\sqrt{2\alpha\epsilon}-\frac{4C_0B+C_1}{\sqrt{t}}\right)^2\right),
\end{align}
where the first inequality is due to $C(x)-C(x^*)\leq (2\alpha)^{-1}|g(x)|^2, \forall x\in[x^*-\beta,x^*+\beta]$, and the second inequality is due to the uniform concentration of gradient (Lemma \ref{le:le_uniform_convergence_g}).

Let $t\geq (3+2\sqrt{2})\max\{1,(4C_0B+C_1)^2\}\max\{1,2B^2\}(\alpha\epsilon)^{-1}\log \delta^{-1}$, Eq.~\eqref{eq:eq_sample_complexity_epsilon_small_first_step} implies that \begin{align*}
    \P[C(\hat{x}_t)-C(x^*)\leq \epsilon]&\geq 1-\exp\left(-\frac{t}{2B^2}\cdot\left(\sqrt{2\alpha\epsilon}-\frac{4C_0B+C_1}{\sqrt{t}}\right)^2\right)\\
    &\geq 1-\exp\left(-\frac{t\alpha\epsilon}{2B^2}\right)\geq 1-\exp\left(\left(3+2\sqrt{2}\right)\log \delta\right)\geq 1-\delta,
\end{align*}
which completes the proof of the case when $\epsilon<\alpha\beta^2/2$.

\noindent{\bf Case 2: $\epsilon>\alpha\beta^2/2$}. In this case, when $t\geq (4C_0B+C_1)^2\bar{D}^2/\epsilon^2$, by simple calculation, we have
\begin{align}\label{eq:eq_sample_complexity_epsilon_large_first_step}
    \P[C(\hat{x}_t)-C(x^*)\leq \epsilon]\geq \P[|g(\hat{x}_t)|\leq \epsilon/\bar{D}]\geq 1-\exp\left(-\frac{t}{2B^2}\left(\frac{\epsilon}{\bar{D}}-\frac{4C_0B+C_1}{\sqrt{t}}\right)^2\right),
\end{align}
where the first inequality is due to $C(x)-C(x^*)\leq |g(x)|\cdot|x-x^*|\leq \bar{D}|g(x)|$, and the second inequality is due to Lemma~\ref{le:le_uniform_convergence_g}.

Let $t\geq 4\bar{D}^2\cdot\max\{1,2B^2\}\cdot\max\{1,(4C_0B+C_1)^2\}\cdot\epsilon^{-2}\log \delta^{-1}$, Eq.~\eqref{eq:eq_sample_complexity_epsilon_large_first_step} implies that \begin{align*}
    \P[C(\hat{x}_t)-C(x^*)\leq \epsilon]\geq 1-\exp\left(-\frac{t}{2B^2}\left(\frac{\epsilon}{\bar{D}}-\frac{4C_0B+C_1}{\sqrt{t}}\right)^2\right)\geq 1-\exp\left(-\frac{t}{2B^2}\cdot\frac{\epsilon^2}{4\bar{D}^2}\right)\geq 1-\delta,
\end{align*}
where we complete the proof of the case when $\epsilon>\alpha\beta^2/2$.

\subsection{Optimal Regret for Linear Inventory Cost Setting Without Assumption~\ref{ass:demand-nvp}}\label{subsec:linear_SC_to_ub}

In this subsection, we consider the case where \[
    C(x)=h\cdot\E[x-D]^++b\cdot\E[D-x]^+.
\] Moreover, as discuss earlier, the demand $D$ and domain $\mathcal{X}$ are unbounded. We relax Assumption \ref{ass:demand-nvp} and rigorously formulate the assumptions imposed on demand distribution as follows.
\begin{assumption}
    \label{ass:relax-demand} 
$\E[D]\leq\mu<\infty$ and $C(x)$ satisfies the $(\alpha,\beta)$-local strong convexity assumption. 
\end{assumption}

With the above assumption, we will also improve the upper bound of cumulative regret in previous works.

\begin{theorem}\label{tm:tm_linearcost_regret_ub}
Suppose Assumption \ref{ass:relax-demand} holds, $0<\alpha<1$, and $\alpha\beta<1/10$. We have
\begin{align}
\sum_{t=1}^T\mathbb{E}\left[C(\hat{x}_t)-C\left(x^*\right)\right]\leq \tilde{K}_1 +\tilde{K}_2\cdot \frac{1}{\alpha\beta}\log\left(\left(\alpha\beta\right)^{-1}\right)+\tilde{K}_3 \cdot \frac{\ln T}{\alpha},\nonumber
\end{align}
where $\tilde{K}_1$, $\tilde{K}_2$, $\tilde{K}_3$ are absolute constants independent of $\alpha$, $\beta$, $T$.
\end{theorem}

Different from the existing methods that utilize quantile properties of newsvendor problems (\cite{lin22datadriven}), our proof is based on the following two carefully designed types of partitions: partition of the time periods and partition of the integral.

\noindent{\bf{Partition of the time periods.}} Recall that combining the results of Theorem \ref{tm:tm_main_regret_ub} and Theorem \ref{tm:tm_general_case_no_local_convex_ass_ub}, the cumulative regret can be upper bounded by $\min\{\sqrt{T},\alpha^{-1}\log T+(\alpha\beta)^{-1}\}$. This joint regret indicates that, when $t$ is small, our analysis may not lead to a better single period result than $\mathcal{O}(t^{-1/2})$. This observation encourages us to partition the time period into two parts: for large $t$, we utilize strongly convexity property to derive a $\mathcal{O}(\alpha^{-1}t^{-1})$ bound; for small $t$, we take the $\mathcal{O}(t^{-1/2})$ bound derived in the proof of Theorem \ref{tm:tm_general_case_no_local_convex_ass_ub}. Specifically, we define the thereshold to be $T_{\alpha,\beta}=\mathcal{O}\left((\alpha\beta)^{-2}(\log(\alpha\beta))^2\right)$, and then consider the cumulative regret form $1$ to $T_{\alpha,\beta}$ and the cumulative regret from $T_{\alpha,\beta}+1$ to $T$ separately:
\[
    \mathcal{R}^{SAA}(T)=\sum_{t=1}^{T_{\alpha,\beta}}\E[C(\hat{x}_t)-C(x^*)]+\sum_{t=T_{\alpha,\beta}+1}^T\E[C(\hat{x}_t)-C(x^*)].
\]

\noindent{\bf{Partition of the integral.}} The first part can be directly upper bounded by $\mathcal{O}(\sqrt{T_{\alpha,\beta}})$. For the second part,  we decompose the regret incurred in period $t$ as follows (assuming $\alpha\beta<C(x^*)<t$):\begin{align*}
    &\hat{I}_{1,t}=\int_{0}^{\alpha\beta^2/2}\P\left[C(\hat{x}_t)-C(x^*)>\epsilon\right]\mathrm{d}\epsilon,\quad \hat{I}_{2,t}=\int_{\alpha\beta^2/2}^{C(x^*)}\P\left[C(\hat{x}_t)-C(x^*)>\epsilon\right]\mathrm{d}\epsilon,\\
    &\hat{I}_{3,t}=\int_{C(x^*)}^{t}\P\left[C(\hat{x}_t)-C(x^*)>\epsilon\right]\mathrm{d}\epsilon,\quad\hat{I}_{4,t}=\int_{t}^{\infty}\P\left[C(\hat{x}_t)-C(x^*)>\epsilon\right]\mathrm{d}\epsilon.
\end{align*}Then, we establish the upper bound of each term individually.
At a high level:\begin{itemize}
    \item When $0\leq\epsilon\leq\alpha\beta^2/2$, we leverage the local strong convexity of $C(x)$ to derive the following lemma, and then derive an upper bound of $\mathcal{O}(\alpha^{-1}\log T)$ for $\hat{I}_{1,t}$.
    \begin{lemma}\label{le:ub_SC_I1t}
    When $0<\epsilon<\alpha\beta^2/2$, $t>(4C_0(h+b)+C_1)^2/(2\alpha\epsilon)$, we have \[
        \P\left[C(\hat{x}_t)-C(x^*)\geq \epsilon\right]\leq \exp\left(-\frac{t}{2(h+b)^2}\left(\sqrt{2\alpha\epsilon}-\frac{4C_0(h+b)+C_1}{\sqrt{t}}\right)^2\right).
    \] 
\end{lemma}
    \item When $\alpha\beta^2/2\leq\epsilon\leq C(x^*)$, we rely on the global convexity of $C(x)$ to obtain the following lemma, and then derive an upper bound of $\mathcal{O}((\alpha\beta)^{-1})$ for $\hat{I}_{2,t}$.
    \begin{lemma}\label{le:ub_SC_I2t}
    When $\alpha\beta^2/2<\epsilon<C(x^*)$, we have \[
        \P\left[C(\hat{x}_t)-C(x^*)\geq \epsilon\right]\leq \exp\left(-\frac{\min\{b,h\}^2\cdot t\epsilon^2}{9(h+b)^2C(x^*)^2}\right)\leq\exp\left(-\frac{\min\{b,h\}^2\cdot t\epsilon^2}{9\mu^2(h+b)^4}\right).
    \]
\end{lemma}
    \item When $C(x^*)\leq\epsilon\leq t$, we use $\P[C(\hat{x}_t)-C(x^*)\geq C(x^*)]$ to give a uniform upper bound of $\P[C(\hat{x}_t)-C(x^*)\geq \epsilon]$, and subsequently give an upper bound of $\mathcal{O}(1)$ for $\hat{I}_{3,t}$.
    \item When $\epsilon\geq t$, we instead consider the tail bound of $\hat{x}_t$, using $\P[|\hat{x}_t-x^*|\geq\epsilon/(h+b)]$ to bound $\P[C(\hat{x}_t)-C(x^*)\geq \epsilon]$. Noting that by Markov inequality and simple calculation, we have $\P[\hat{x}_t\geq x]\leq \mathcal{O}(t^{-1}\P[D\geq x])$ when $x$ is large. Therefore, an upper bound of $\mathcal{O}(\log T)$ for $\hat{I}_{4,t}$ can be obtained.
\end{itemize}

\subsection{Proof of Theorem \ref{tm:tm_linearcost_regret_ub}}
\proof{Proof of Theorem \ref{tm:tm_linearcost_regret_ub}.}
We define the threshold \begin{align}\label{eq:hat_T_def}\nonumber    
    T_{\alpha,\beta}=&400\max\{1,(4C_0(h+b)+C_1)^2\}\cdot\max\{(h+b)\mu/(1-\rho),1\}\\
    &\cdot\max\{1,(3\mu(h+b)^2\min\{h,b\}^{-1})^4\}\cdot(\alpha\beta)^{-2}(\log(\alpha\beta))^2,
\end{align}
As discussed earlier, we have\[
    \mathcal{R}^{SAA}(T)=\sum_{t=1}^{T_{\alpha,\beta}}\E[C(\hat{x}_t)-C(x^*)]+\sum_{t=T_{\alpha,\beta}+1}^T\hat{I}_{1,t}+\sum_{t=T_{\alpha,\beta}+1}^T\hat{I}_{2,t}+\sum_{t=T_{\alpha,\beta}+1}^T\hat{I}_{3,t}+\sum_{t=T_{\alpha,\beta}+1}^T\hat{I}_{4,t}.
\]
\noindent{\bf Upper bound of $\sum_{t=1}^{T_{\alpha,\beta}}\E[C(\hat{x}_t)-C(x^*)]$}. By Proposition~1 in \cite{lin22datadriven}, we have \begin{align}\label{eq:regret_small_T_finite_mean_ub}
    \sum_{t=1}^{T_{\alpha,\beta}}\E[C(\hat{x}_t)-C(x^*)]\leq K_G\sqrt{T_{\alpha,\beta}}\leq \mathcal{O}\left(\left(\alpha\beta\right)^{-1}\log\left(\alpha\beta\right)^{-1}\right),
\end{align}
where $K_G=((b+h)/(1-\rho))\cdot2(1+\mu)^2$. Next, we give the upper bound of the summations of $\hat{I}_{1,t},\hat{I}_{2,t},\hat{I}_{3,t},\hat{I}_{4,t}$ separately.

\noindent{\bf Upper bound of $\sum_{t=T_{\alpha,\beta}+1}^T \hat{I}_{1,t}$}. Let $\bar{r}_{1,t}=\min\{(3+2\sqrt{2})(4C_0(h+b)+C_1)^2/(\alpha t),\alpha\beta^2/2\}$. Recall that in Lemma~\ref{le:ub_SC_I1t}, we obtain that $\P[C(\hat{x}_t)-C(x^*)\geq \epsilon]\leq\exp\left(-\frac{t}{2(h+b)^2}\left(\sqrt{2\alpha\epsilon}-\frac{4C_0(h+b)+C_1}{\sqrt{t}}\right)^2\right)$ when $\frac{(4C_0(h+b)+C_1)^2}{2\alpha t}\leq\epsilon\leq\frac{\alpha\beta^2}{2}$. Thus, for each $t> T_{\alpha,\beta}$, we have \begin{align*}
    \hat{I}_{1,t}&=\int_{0}^{\frac{\alpha\beta^2}{2}}\P\left[C(\hat{x}_t)-C(x^*)\geq\epsilon\right]\mathrm{d}\epsilon\\
    &\leq\int_{0}^{\bar{r}_{1,t}}1\mathrm{d}\epsilon+\int_{\bar{r}_{1,t}}^{\frac{\alpha\beta^2}{2}}\P\left[C(\hat{x}_t)-C(x^*)\geq\epsilon\right]\mathrm{d}\epsilon\\
    &\leq \bar{r}_{1,t}+\int_{\bar{r}_{1,t}}^{\frac{\alpha\beta^2}{2}}\exp\left(-\frac{t}{2(h+b)^2}\left(\sqrt{2\alpha\epsilon}-\frac{4C_0(h+b)+C_1}{\sqrt{t}}\right)^2\right)\mathrm{d}\epsilon\\
    &\leq \bar{r}_{1,t}+\int_{\bar{r}_{1,t}}^{\frac{\alpha\beta^2}{2}}\exp\left(- \frac{t\alpha\epsilon}{2(h+b)^2}\right)\mathrm{d}\epsilon\\&\leq \bar{r}_{1,t}+\int_0^{\infty}\exp\left(- \frac{t\alpha\epsilon}{2(h+b)^2}\right)\mathrm{d}\epsilon\\
    &\leq \frac{(3+2\sqrt{2})(4C_0(h+b)+C_1)^2+2(h+b)^2}{\alpha t},
\end{align*}
where the second inequality and the third inequality are due to the fact that $\sqrt{2\alpha\epsilon}-(4C_0(h+b)+C_1)t^{-1/2}\geq\sqrt{\alpha\epsilon}$ for $\epsilon\geq\bar{r}_{1,t}$.

Cumulating $\hat{I}_{1,t}$ from $T_{\alpha,\beta}+1$ to $T$, we obtain that for $T\geq 2$, \begin{align}
    \nonumber\sum_{t=T_{\alpha,\beta}+1}^T\hat{I}_{1,t}&\leq\sum_{t=T_{\alpha,\beta}+1}^T\frac{(3+2\sqrt{2})(4C_0(h+b)+C_1)^2+2(h+b)^2}{\alpha t}\\
    &\leq  2\left((3+2\sqrt{2})(4C_0(h+b)+C_1)^2+2(h+b)^2\right)\cdot\frac{\log T}{\alpha}. \label{eq:hat_I1_bound}
\end{align}

\noindent{\bf Upper bound of $\sum_{t=T_{\alpha,\beta}+1}^T \hat{I}_{2,t}$}. Note that \[
    \hat{I}_{2,t}=\int_{\alpha\beta^2/2}^{\alpha\beta}\P[C(\hat{x}_t)-C(x^*)\geq \epsilon]\mathrm{d}\epsilon+\int_{\alpha\beta}^{C(x^*)}\P[C(\hat{x}_t)-C(x^*)\geq \epsilon]\mathrm{d}\epsilon.
\]
For the first term, suppose $\alpha\beta<1/e$, when $t>T_{\alpha,\beta}\geq 16(4C_0(h+b)+C_1)^{2}\cdot(\alpha\beta)^{-2}$, by Lemma~\ref{le:ub_SC_I1t} and simple calculation, we obtain that \begin{align*}
    \int_{\alpha\beta^2/2}^{\alpha\beta}\P[C(\hat{x}_t)-C(x^*)\geq \epsilon]\mathrm{d}\epsilon&\leq\int_{\alpha\beta^2/2}^{\alpha\beta}\P[C(\hat{x}_t)-C(x^*)\geq \alpha\beta^2/2]\mathrm{d}\epsilon\\
    &\leq \int_{\alpha\beta^2/2}^{\alpha\beta}\exp\left(-\frac{t}{2(h+b)^2}\left(\alpha\beta-\frac{4C_0(h+b)+C_1}{\sqrt{t}}\right)^2\right)\mathrm{d}\epsilon\\
    &\leq \alpha\beta\cdot \exp\left(-\frac{t}{2(h+b)^2}\left(\alpha\beta-\frac{4C_0(h+b)+C_1}{\sqrt{t}}\right)^2\right)\\
    &\leq \alpha\beta\cdot\exp\left(-\frac{t\alpha^2\beta^2}{4(h+b)^2}\right).
\end{align*}
For the second term, by Lemma~\ref{le:ub_SC_I2t}, we obtain that \begin{align}
    \nonumber\int_{\alpha\beta}^{C(x^*)}\P[C(\hat{x}_t)-C(x^*)\geq \epsilon]\mathrm{d}\epsilon&\leq \int_{\alpha\beta}^{C(x^*)}\exp\left(\frac{-\min\{b,h\}^2\cdot t\epsilon^2}{9\mu^2(h+b)^4}\right)\mathrm{d}\epsilon\\
    \nonumber&\leq\int_{\alpha\beta}^{\infty}\exp\left(\frac{-\min\{b,h\}^2\cdot t\epsilon^2}{9\mu^2(h+b)^4}\right)\mathrm{d}\epsilon\\
    \nonumber&=\frac{3\mu(h+b)^2}{\sqrt{2t}\min\{b,h\}}\int_{\frac{\alpha\beta\sqrt{2t}\min\{b,h\}}{3\mu(h+b)^2}}^{\infty}\exp\left(\frac{-\epsilon^2}{2}\right)\mathrm{d}\epsilon\\
    \nonumber&\leq \frac{9\mu^2(h+b)^4}{2t\alpha\beta\min\{b,h\}^2}\exp\left(\frac{-\min\{b,h\}^2\alpha^2\beta^2t}{9\mu^2(h+b)^4}\right),
\end{align}
where the last inequality is due to the fact that $\int_{z}^{\infty}\exp(-\epsilon^2/2)\mathrm{d}\epsilon\leq z^{-1}\exp(-z^2/2)$.

Noting that when $t>T_{\alpha,\beta}$, we have $\sqrt{t}\geq 3\mu(h+b)^2\min\{h,b\}^{-1}(\alpha\beta)^{-1}\log t$. By this fact, we obtain the following bound for the cumulative bound of $\hat{I}_{2,t}$:\begin{align}\label{eq:hat_I3_bound}
    \nonumber\sum_{t=T_{\alpha,\beta}+1}^T\hat{I}_{2,t}&=\sum_{t=T_{\alpha,\beta}+1}^T\left(\int_{\alpha\beta^2/2}^{\alpha\beta}\P[C(\hat{x}_t)-C(x^*)\geq \epsilon]\mathrm{d}\epsilon+\int_{\alpha\beta}^{C(x^*)}\P[C(\hat{x}_t)-C(x^*)\geq \epsilon]\mathrm{d}\epsilon\right)\\
    \nonumber&\leq\sum_{t=T_{\alpha,\beta}+1}^T\left(\alpha\beta\exp\left(-\frac{t\alpha^2\beta^2}{4(h+b)^2}\right)+\frac{9\mu^2(h+b)^4}{2t\alpha\beta(\min\{b,h\})^2}\cdot\exp\left(\frac{-(\min\{b,h\})^2\alpha^2\beta^2t}{9\mu^2(h+b)^4}\right)\right)\\
    \nonumber&\leq\sum_{t=T_{\alpha,\beta}+1}^T\left(\alpha\beta\exp\left(-\frac{t\alpha^2\beta^2}{4(h+b)^2}\right)+\frac{9\mu^2(h+b)^4}{2t\alpha\beta(\min\{b,h\})^2}\cdot\exp\left(-\left(\log t\right)^2\right)\right) \\
    &\leq\frac{4(h+b)^2}{\alpha\beta}+\frac{9\mu^2(h+b)^4}{2(\min\{b,h\})^2\alpha\beta}.
\end{align}

\noindent{\bf Upper bound of $\sum_{t=T_{\alpha,\beta}+1}^T\hat{I}_{3,t}$}. Recall that Lemma~\ref{le:ub_SC_I2t} implies that, when $\epsilon=C(x^*)$, the sample complexity is  $N_{C(x^*),\delta}=9(\min\{b,h\}/(b+h))^{-2}\log \delta^{-1}$. Thus, by calculation, we have \begin{align*}
    &\hat{I}_{3,t}=\int_{C(x^*)}^{t}\P[C(\hat{x}_t)-C(x^*)\geq \epsilon]\mathrm{d}\epsilon\leq t\cdot\P[C(\hat{x}_t)-C(x^*)\geq C(x^*)]\leq t\cdot\exp\left(-\left(\frac{\min\{b,h\}}{3(b+h)}\right)^2t\right).
\end{align*}
Cumulating $\hat{I}_{3,t}$ form $T_{\alpha,\beta}+1$ to $T$, we obtain that \begin{align}\label{eq:hat_I4_bound}
    \sum_{t=T_{\alpha,\beta}+1}^T\hat{I}_{3,t}\leq\sum_{t=T_{\alpha,\beta}+1}^Tt\cdot\exp\left(-\left(\frac{\min\{b,h\}}{3(b+h)}\right)^2t\right)\leq \left(\frac{3(b+h)}{\min\{b,h\}}\right)^{4}.
\end{align}

\noindent{\bf Upper bound of $\sum_{t=T_{\alpha,\beta}+1}^T\hat{I}_{4,t}$}. By the fact that $|g(x)|=|(h+b)\P[D\leq x]-b|\leq h+b$, we have \begin{equation}\label{eq:eq_hat_I4_C_ub}
    C(\hat{x}_t)-C(x^*)\leq |g(\hat{x}_t)|\cdot|\hat{x}_t-x^*|\leq (h+b)\cdot |\hat{x}_t-x^*|,
\end{equation}
where the first inequality is due to the convexity of $C(x)$. When $\epsilon\geq\mu(h+b)/(1-\rho)\geq (h+b)x^*$, Eq.~\eqref{eq:eq_hat_I4_C_ub} implies that \begin{align}\nonumber
    \P[C(\hat{x}_t)-C(x^*)\geq\epsilon]&\leq \P\left[\left|\hat{x}_t-x^*\right|\geq\epsilon/(h+b)\right]=\P[\hat{x}_t-x^*\geq \epsilon/(h+b)]\\
    &\leq \frac{1}{(1-\rho)^2} \cdot\frac{(h+b)\mu+1}{t}\cdot\P\left[D\geq\epsilon/(h+b)+x^*\right], \label{eq:eq_tail_xt}
\end{align}
where the second inequality is due to the fact that \begin{align*}
    \P\left[\hat{x}_t\geq x\right]&=\P\left[\hat{F}_t\left(x\right)\leq \rho\right]=\P\left[\frac{1}{t}\sum_{i=1}^t\mathbb{I}\left[d_i\geq x\right]\geq 1-\rho\right]\\
    &\leq \frac{1}{(1-\rho)^2t^2}\E\left[\left(\sum_{i=1}^t\mathbb{I}\left[d_i\geq x\right]\right)^2\right] \\
    &=\frac{1}{(1-\rho)^2}\left(\frac{t-1}{t}\cdot\P\left[D\geq x\right]^2+\frac{1}{t}\P\left[D\geq x\right]\right)\\
    &\leq \frac{1}{(1-\rho)^2}\left(\frac{\E[D]}{x}\cdot\P\left[D\geq x\right]+\frac{1}{t}\P\left[D\geq x\right]\right)\\
    &\leq\frac{1}{(1-\rho)^2} \cdot\frac{(h+b)\mu+1}{t}\cdot\P\left[D\geq x\right], \forall x\geq t/(h+b),
\end{align*}
where the second inequality is due to Markov inequality. 

By the analysis above, we conclude that \begin{align}\label{eq:hat_I5_bound}
    \nonumber\sum_{t=T_{\alpha,\beta}+1}^T\hat{I}_{4,t}&=\sum_{t=T_{\alpha,\beta}+1}^T\int_{t}^{\infty}\P[C(\hat{x}_t)-C(x^*)\geq \epsilon]\mathrm{d}\epsilon\\
    \nonumber&\leq\sum_{t=T_{\alpha,\beta}+1}^T\int_t^{\infty}\frac{1}{(1-\rho)^2} \frac{(h+b)\mu+1}{t}\P\left[D\geq\frac{\epsilon}{h+b}+x^*\right]\mathrm{d}\epsilon \\
    \nonumber&\leq\sum_{t=T_{\alpha,\beta}+1}^T\frac{(h+b)\mu+1}{(1-\rho)^2} \frac{1}{t}\int_0^{\infty}\P\left[D\geq\frac{\epsilon}{h+b}\right]\mathrm{d}\epsilon\\
    \nonumber&=\sum_{t=T_{\alpha,\beta}+1}^T\frac{(h+b)((h+b)\mu+1)}{(1-\rho)^2} \frac{1}{t}\int_0^{\infty}\P\left[D\geq\epsilon\right]\mathrm{d}\epsilon\\
    &\leq \sum_{t=T_{\alpha,\beta}+1}^T\frac{(h+b)((h+b)\mu+1)}{(1-\rho)^2} \frac{1}{t}\mu\nonumber\\&\leq\frac{(h+b)\mu((h+b)\mu+1)}{(1-\rho)^2} \log T,
\end{align}
where the first inequality is due to Eq.~\eqref{eq:eq_tail_xt}.

Combining Eqs. (\ref{eq:regret_small_T_finite_mean_ub},\ref{eq:hat_I1_bound},\ref{eq:hat_I1_bound},\ref{eq:hat_I3_bound},\ref{eq:hat_I4_bound},\ref{eq:hat_I5_bound}), we conclude that \begin{align*}
    \sum_{t=1}^T\E[C(\hat{x}_t)-C(x^*)]&
    =\sum_{t=1}^{T_{\alpha,\beta}}\E[C(\hat{x}_t)-C(x^*)]+\sum_{t=T_{\alpha,\beta}+1}^T\hat{I}_{1,t}+\sum_{t=T_{\alpha,\beta}+1}^T\hat{I}_{2,t}+\sum_{t=T_{\alpha,\beta}+1}^T\hat{I}_{3,t}+\sum_{t=T_{\alpha,\beta}+1}^T\hat{I}_{4,t}\\
    &\leq \mathcal{O}\left(\frac{1}{\alpha}\log T+\frac{1}{\alpha\beta}\log \frac{1}{\alpha\beta}\right).
\end{align*}

\subsection{Proof of Lemma~\ref{le:ub_SC_I1t} and Lemma~\ref{le:ub_SC_I2t}}
\proof{Proof of Lemma~\ref{le:ub_SC_I1t}.}
Note that in the proof of Proposition \ref{Prop:sample_complexity}, when considering the case when $0<\epsilon<\alpha\beta^2/2$, we do not the boundedness assumption on demand. Also note that the linear inventory cost naturally satisfies Assumption \ref{ass:regular}, where in the assumption ``$c^{\prime}(\cdot,d)\leq B$'', $B$ is replaced by $h+b$ due to the fact that $|C^{\prime}(x)|=|g(x)|=|(h+b)\P[D\leq x]-b|\leq h+b$. Thus, following almost step by step as the proof of Proposition \ref{Prop:sample_complexity}, for $0<\epsilon<(h+b)\alpha\beta^2/2$, $t\geq(3+2\sqrt{2})\max\{1,4C_0(h+b)+C_1\}^2\max\{1,2(h+b)^2\}\alpha^{-1}\epsilon^{-1}\log \delta^{-1}$, one has \[
    \P[C(\hat{x}_t)-C(x^*)\leq\epsilon]\geq 1-\exp\left((3+\sqrt{2})\log\delta\right)\geq 1-\delta.
\]
In other words, when $0<\epsilon<\alpha\beta^2/2$, $t>(4C_0(h+b)+C_1)/(2\alpha\epsilon)$, we have \[
        \P\left[C(\hat{x}_t)-C(x^*)\geq \epsilon\right]\leq \exp\left(-\frac{t}{2(h+b)^2}\left(\sqrt{2\alpha\epsilon}-\frac{4C_0(h+b)+C_1}{\sqrt{t}}\right)^2\right),
    \] 
which completes the proof.

\proof{Proof of Lemma~\ref{le:ub_SC_I2t}.}
Note that for all $x>0$, we have \begin{align*}C(x)=h\cdot\E[x-D]^++b\cdot\E[D-x]^+\leq h\cdot x\cdot\P[D\leq x]+b\cdot\E[D]\leq h\cdot\E[D]+b\cdot\E[D]\leq (b+h)\mu,
\end{align*}
where the second inequality is due to Markov inequality. Combining this result and Theorem~2.2 in \cite{levi2007nearoptimal}, we can directly obtain that \[
    \P\left[C(\hat{x}_t)-C(x^*)\geq \epsilon\right]\leq \exp\left(-\frac{\min\{b,h\}^2\cdot t\epsilon^2}{9\mu^2(h+b)^4}\right).
\]
which completes the proof.

\end{document}